\documentclass[10pt,twocolumn,letterpaper]{article}
\usepackage[pagenumbers]{cvpr}
\usepackage{tikz}
\usepackage{fix-cm}
\usepackage{colortbl}
\definecolor{cvprblue}{rgb}{0.21,0.49,0.74}
\usepackage[pagebackref,breaklinks,colorlinks,allcolors=cvprblue]{hyperref}
\def\confName{CVPR}
\def\confYear{2027}
\newcommand{\method}{AesDistill\xspace}
\definecolor{titlebluedark}{HTML}{123E73}
\definecolor{titlebluelight}{HTML}{78BCE6}
\definecolor{titleorangedark}{HTML}{9D3E0C}
\definecolor{titleorangelight}{HTML}{EFA663}
\newsavebox{\titlegradientbox}
\newcommand{\titlegradient}[3]{%
  \begingroup
  \sbox{\titlegradientbox}{#3}%
  \leavevmode\hbox{%
    \pdfliteral{q}%
    \rlap{\pdfliteral{7 Tr}\usebox{\titlegradientbox}\pdfliteral{0 Tr}}%
    \rlap{\raisebox{-\dp\titlegradientbox}{%
      \begin{tikzpicture}
        \shade[left color=#1,right color=#2] (0,0) rectangle
          (\wd\titlegradientbox,\dimexpr\ht\titlegradientbox+\dp\titlegradientbox\relax);
      \end{tikzpicture}%
    }}%
    \pdfliteral{Q}%
    \phantom{\usebox{\titlegradientbox}}%
  }%
  \endgroup
}
\newcommand{\titlemethod}{%
  \titlegradient{titlebluedark}{titlebluelight}{Aes}%
  \titlegradient{titleorangedark}{titleorangelight}{Distill}%
}
\newcommand{\studentname}{AesDistill Student\xspace}
\newcommand{\datasetname}{AesDistill Dataset\xspace}
\newcommand{\fiveksubset}{AesDistill FiveK\xspace}
\newcommand{\pprsubset}{AesDistill PPR10K\xspace}
\newcommand{\correspondencemodule}{Confidence-Aware Correspondence\xspace}
\newcommand{\transformmodule}{Teacher-Conditioned Color Transforms\xspace}
\newcommand{\supervisionmodule}{Confidence-Weighted Color Supervision\xspace}
\newcommand{\textbfnoindent}[1]{\par\noindent\textbf{#1}\enspace}
\title{\texorpdfstring{\titlemethod}{AesDistill}: Distilling Generative Aesthetics for Image Color Enhancement}
\author{%
Chuanzhi Xu\textsuperscript{1,$\dagger$}\quad
Chengkun Yue\textsuperscript{2}\quad
Huiqi Liang\textsuperscript{1}\quad
Haoyang Li\textsuperscript{1}\\
Zihan Deng\textsuperscript{3}\quad
Qingwen Zeng\textsuperscript{1}\quad
Pengfei Ye\textsuperscript{4}\quad
Weidong Cai\textsuperscript{1}\\[4pt]
{\small\textsuperscript{1}The University of Sydney\quad
\textsuperscript{2}Indiana University\quad
\textsuperscript{3}The University of Hong Kong\\[2pt]
\textsuperscript{4}Massachusetts Institute of Technology\quad\href{https://github.com/LouckXu/AesDistill}{\raisebox{-0.12em}{\includegraphics[height=0.9em]{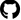}}\hspace{0.3em}Project Page}}}
\hypersetup{pdftitle={AesDistill: Distilling Generative Aesthetics for Image Color Enhancement},pdfauthor={Chuanzhi Xu, Chengkun Yue, Huiqi Liang, Haoyang Li, Zihan Deng, Qingwen Zeng, Pengfei YE, Weidong Cai}}
\makeatletter
\renewcommand{\@maketitle}{\begin{center}{\Large\bfseries\@title\par}\vspace{14pt}{\large\@author\par}\end{center}\vspace{10pt}}
\makeatother
\begin{document}
\raggedbottom
\twocolumn[{%
\maketitle
\begin{minipage}{\textwidth}
\centering
\includegraphics[width=\textwidth]{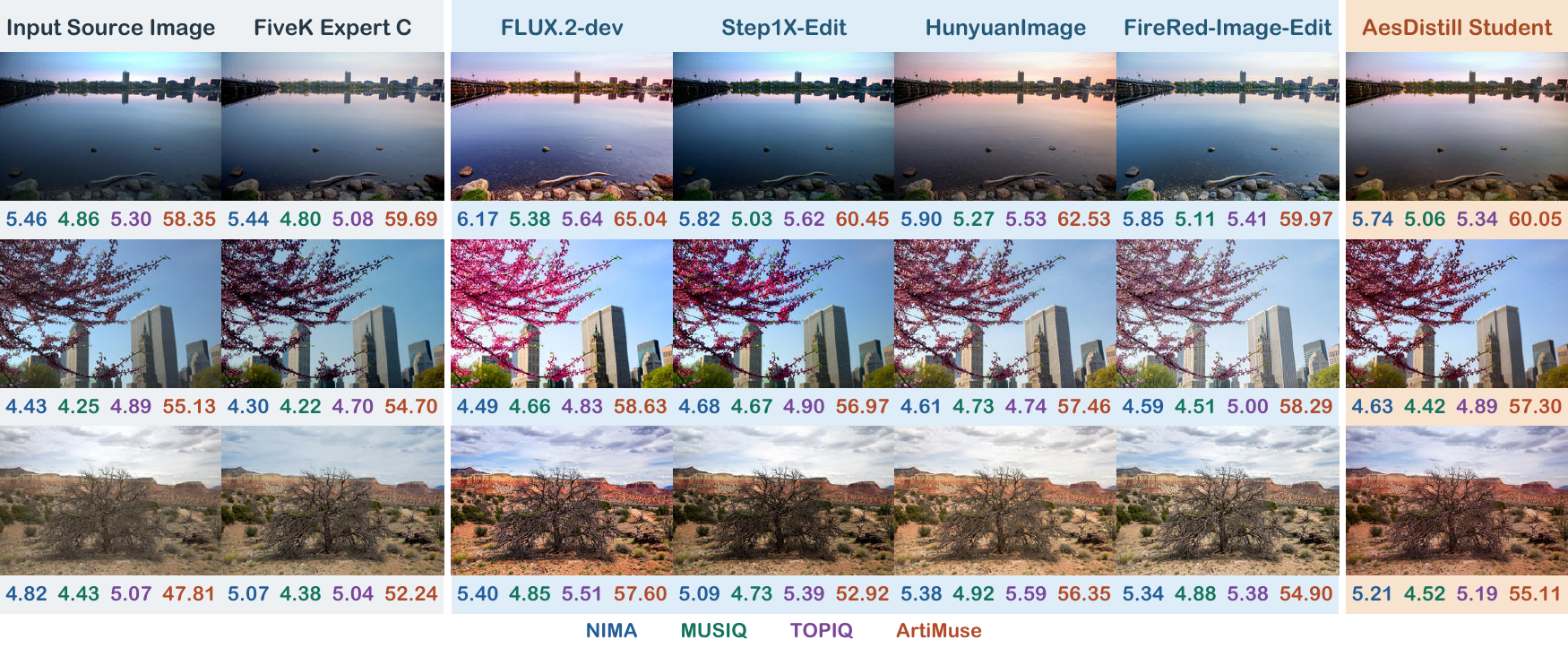}
\captionsetup{hypcap=false}
\captionof{figure}{\method distills aesthetic knowledge from large generative model teachers trained on diverse images into a compact student. Our teachers and \studentname show higher aesthetic quality than expert targets on the FiveK examples shown above.}
\label{fig:teaser}
\end{minipage}\par
\vspace{12pt}
}]
\begingroup
\renewcommand{\thefootnote}{\ensuremath{\dagger}}
\footnotetext[2]{Corresponding author: \href{mailto:chuanzhi.xu@sydney.edu.au}{\texttt{chuanzhi.xu@sydney.edu.au}}.}
\endgroup

\begin{abstract}
Image color enhancement aims to improve visual appeal, but previous methods often learn to reproduce ground-truth expert targets. These targets reflect the preferences of a few experts and cover a limited range of aesthetic choices. Matching them does not ensure stronger aesthetic quality across different tastes and scenes. By contrast, large generative models learn rich visual knowledge from large image collections and often produce vivid, appealing image edits. However, direct generation can change the source image's semantic structure along with its colors. In this paper, we introduce \method, a framework for distilling generative aesthetics into image color enhancement. We use 4 leading open-source large generative models to produce 64,644 edited images as teacher targets at a cost of about 3,360 H100 GPU-hours, forming the \datasetname. \method also includes \studentname, a 1.13M-parameter model that learns teacher color changes through confidence-weighted supervision while preserving the source image's semantic structure. We evaluate \method with 12 aesthetic and 4 image-quality assessment models. Every teacher's targets rank above the FiveK and PPR10K expert target means in the joint assessment. Across both datasets, \studentname outperforms all 20 baseline methods and ranks directly behind the 4 teachers. These results demonstrate that distilling aesthetics from generative models enables a compact student to outperform methods trained on expert targets.
\end{abstract}

\begin{figure}[t]
\centering
\includegraphics[width=\columnwidth]{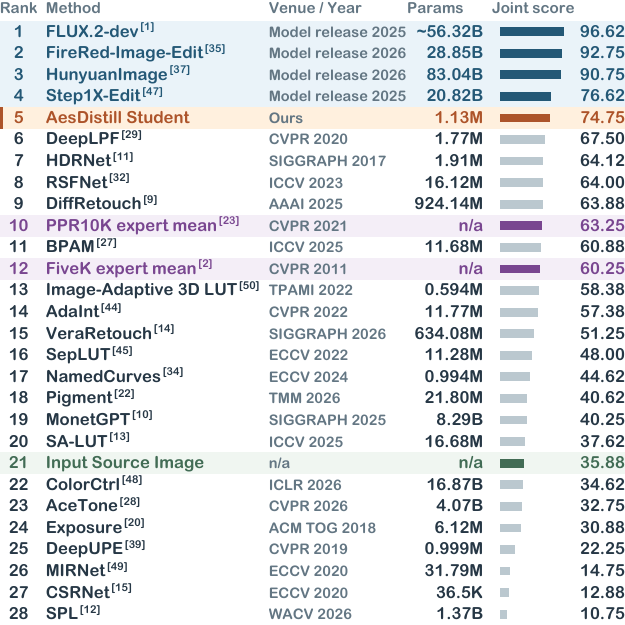}
\caption{Joint Score averages percentile ranks over 12 aesthetic and 4 image-quality assessments. \studentname leads all 20 baselines, directly behind the four large generative teachers.}
\label{fig:joint-rank-teaser}
\end{figure}

\section{Introduction}
\label{sec:introduction}

Image color enhancement improves visual appeal through changes in tone, color, and local contrast. Bilateral affine grids~\cite{hdrnet}, image-adaptive lookup tables (LUTs)~\cite{adaptive3dlut}, and tone curves~\cite{namedcurves} make these adjustments efficient at high resolution. Paired datasets such as MIT-Adobe FiveK~\cite{fivek} and PPR10K~\cite{ppr10k} provide expert-retouched images as ground-truth targets. These datasets have made agreement with expert references a common basis for learning and evaluating color enhancement.

However, matching an expert reference does not establish broader aesthetic quality. FiveK provides ground-truth edits from five experts and PPR10K from three experts~\cite{fivek,ppr10k}. Their targets reflect a limited set of aesthetic preferences for tone, contrast, and color balance. Training against these targets teaches a model to reproduce those fixed aesthetic choices, while other appealing image edits may be penalized for departing from the reference. Therefore, improving visual appeal requires broader aesthetic supervision beyond these fixed references.

Large generative models learn visual knowledge from large image collections, linking image understanding with visually appealing generation~\cite{flux2,hunyuan,januspro,showo,bagel}. Color edits from these models often show vivid colors and pleasing tones (Figure~\ref{fig:teaser}). Therefore, their visual knowledge offers a source of aesthetic supervision beyond the preferences recorded in expert targets. However, using these generative models to directly produce enhanced images can alter the source image's content and semantic structure along with its colors~\cite{gencolor,instantretouch} (Figure~\ref{fig:structure-preservation}).

In this paper, we introduce the \datasetname and a lightweight image color enhancement model, \studentname, to turn generative aesthetic knowledge into reusable supervision and efficient image color enhancement. The dataset pairs each source image with edits from four leading open-source large generative models: FLUX.2-dev~\cite{flux2} ($\approx$56.3B), Step1X-Edit~\cite{step1x_base,reasonedit} (20.8B), HunyuanImage~\cite{hunyuan} (83.0B), and FireRed-Image-Edit~\cite{firered} (28.8B). \studentname learns their color choices through reliable correspondences and regional color distributions, then applies teacher-conditioned global and local transforms at the original pixel positions to preserve source image structure. Evaluation on FiveK and PPR10K covers 12 aesthetic and four image-quality assessments, with comparisons against 20 baseline methods (Figure~\ref{fig:joint-rank-teaser}).

Our main contributions can be summarized as follows:
\begin{itemize}
\item We construct the \datasetname using about 3,360 H100 GPU-hours, providing 64,644 aesthetic annotations for FiveK and PPR10K. On both FiveK and PPR10K, the four teachers' average score exceeds the best expert's mean score under all 12 aesthetic and 4 image-quality assessment methods.
\item We propose \studentname, an image color enhancement model with only 1.13M parameters. It distills four large generative teachers' aesthetic knowledge into color transforms that preserve source image structure and reproduce the teachers' distinct editing choices.
\item Across both datasets, \studentname leads all 20 image aesthetic enhancement baselines in the joint aesthetic and image-quality ranking, placing directly behind the four teachers. Our results demonstrate that generative teachers surpass expert targets in aesthetic quality and that this advantage transfers to a compact student.
\end{itemize}

\section{Related Work}
\label{sec:related-work}

\textbfnoindent{Image Color Enhancement.}
Image-adaptive LUTs provide efficient RGB mappings~\cite{adaptive3dlut,adaint,seplut}. NamedCurves uses color-specific tone curves~\cite{namedcurves}, while Pigment blends learned pigment representations~\cite{pigment}. HDRNet and BPAM predict bilateral grids of affine transforms or pixel-adaptive MLPs~\cite{hdrnet,bpam}, and DeepLPF and RSFNet learn local filters~\cite{deeplpf,rsfnet}. CSRNet modulates pointwise features~\cite{csrnet}, MIRNet fuses multi-resolution features~\cite{mirnet}, and DeepUPE predicts illumination~\cite{deepupe}. Exposure learns editing sequences from unpaired images~\cite{exposure}. However, only improving color representations does not establish the aesthetic value of training targets. We compare generative teacher targets with expert references and use these annotations to train a compact model for aesthetic color editing.

\textbfnoindent{Generative Aesthetic Transfer.}
DiffRetouch uses Stable Diffusion 2.1 with bilateral transforms for controllable editing~\cite{diffretouch}. GenColor combines Stable Diffusion 2.1 and ControlNet with texture-preserving color transfer~\cite{gencolor}. Their multi-step generation requires repeated model evaluations for each input. Instead, VeraRetouch uses a 0.5B vision-language model with aesthetic rewards~\cite{veraretouch}. InstantRetouch distills a Stable Diffusion teacher using variational score distillation and prompt alignment~\cite{instantretouch}. Its inference combines a one-step diffusion branch with bilateral rendering. By contrast, our four large teachers provide reusable image targets with higher aesthetic quality than expert edits. Confidence-weighted color supervision trains a 1.13M student to predict color transforms without diffusion.

\textbfnoindent{Image Aesthetic Assessment.}
NIMA predicts rating distributions, while MLSP pools features from multiple layers~\cite{nima,mlsp}. MUSIQ preserves multi-scale inputs, and TOPIQ uses semantic guidance across feature scales~\cite{musiq,topiq}. ArtQuant learns artistic image scoring through structured aesthetic descriptions~\cite{artquant}. Q-Align learns text-defined rating levels, while ArtiMuse combines aesthetic analysis with fine-grained scoring~\cite{qalign,artimuse}. Many models share AVA crowd ratings~\cite{ava}. However, one score cannot establish consistency across aesthetic perspectives. We evaluate with 12 aesthetic assessment models spanning recent advances and influential methods.

\section{The \datasetname}
\label{sec:dataset}

\begin{figure}[t]
\centering
\includegraphics[width=\linewidth]{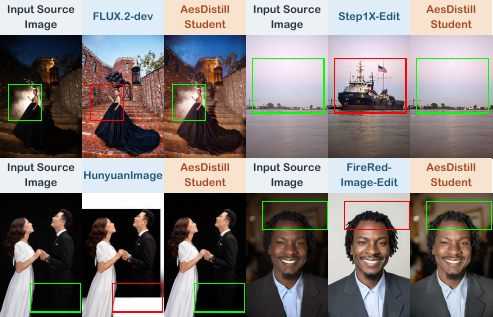}
\caption{Inputs, teacher edits, and student outputs with matching teacher conditions. Red boxes highlight teacher content changes, and green boxes mark corresponding input and student regions.}
\label{fig:structure-preservation}
\end{figure}

\begin{figure}[t]
\centering
\includegraphics[width=\columnwidth]{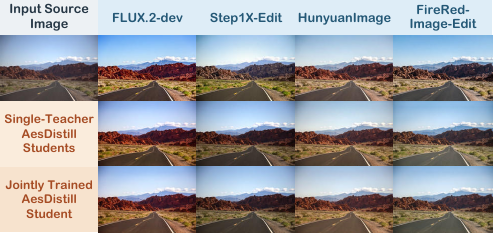}
\caption{One FiveK input with teacher edits (top), single-teacher students with matched training steps (middle), and one jointly trained shared student under all four teacher conditions (bottom).}
\label{fig:teacher-student-styles}
\end{figure}

\textbfnoindent{Dataset Construction.}
FiveK~\cite{fivek} and PPR10K~\cite{ppr10k} are widely used benchmarks for image color enhancement, covering general photography and portraits. We use all 5,000 FiveK and 11,161 PPR10K images to construct \fiveksubset and \pprsubset, respectively. We use four generative teachers: FLUX.2-dev~\cite{flux2}, Step1X-Edit~\cite{reasonedit}, HunyuanImage~\cite{hunyuan}, and FireRed-Image-Edit~\cite{firered}. They receive the same source image and an instruction requesting natural, visually pleasing color and tone changes while preserving content (Figure~\ref{fig:teaser}). However, these models can alter scene content in some cases (Figure~\ref{fig:structure-preservation}). Each teacher produces one edit per source image, giving a dataset of 16,161 source images and 64,644 teacher edits at a cost of about 3,360 H100 GPU-hours. Edits from different teachers reflect different choices of tone, contrast, and color balance (Figure~\ref{fig:teacher-student-styles}) and provide reusable supervision. Source splits and generation settings are provided in Appendix~\ref{app:data-details}.

\textbfnoindent{Multi-Model Aesthetic Assessment.}
We evaluate teacher and expert edits on 500 FiveK and 2,286 PPR10K test images using 12 aesthetic and 4 technical-quality assessment methods (Table~\ref{tab:assessment-joint}). Evaluating one output variant takes 1.03 H100 GPU-hours on average (Appendix~\ref{app:evaluation-cost}). Assessment models use different score scales, so directly averaging their raw scores is unsuitable. Therefore, we define Joint Score (JS) using percentile ranks. For method $f$, JS is:
\begin{equation}
\mathrm{JS}(f)=\frac{100}{2\cdot16}
\sum_{d=1}^{2}\sum_{m=1}^{16}
\frac{L_{d,m,f}+\tfrac12 E_{d,m,f}}{25},
\label{eq:joint-score}
\end{equation}
where $L_{d,m,f}$ and $E_{d,m,f}$ count lower and tied reference scores for method $f$ under assessment $m$ on dataset $d$. Each fixed reference pool contains scores from 20 baselines, 4 teachers, and the input, with one entry each (Appendix~\ref{app:ranking}). JS weights assessments and datasets equally. Higher JS indicates better overall performance. All four teachers rank above the expert target means (Figure~\ref{fig:joint-rank-teaser}). On each dataset, FLUX.2-dev, HunyuanImage, and FireRed-Image-Edit exceed the best expert on all 12 aesthetic assessments, while Step1X-Edit does so on 10. See Appendices~\ref{app:assessors} and~\ref{app:scores}.

\begin{figure*}[t]
\centering
\includegraphics[width=\textwidth]{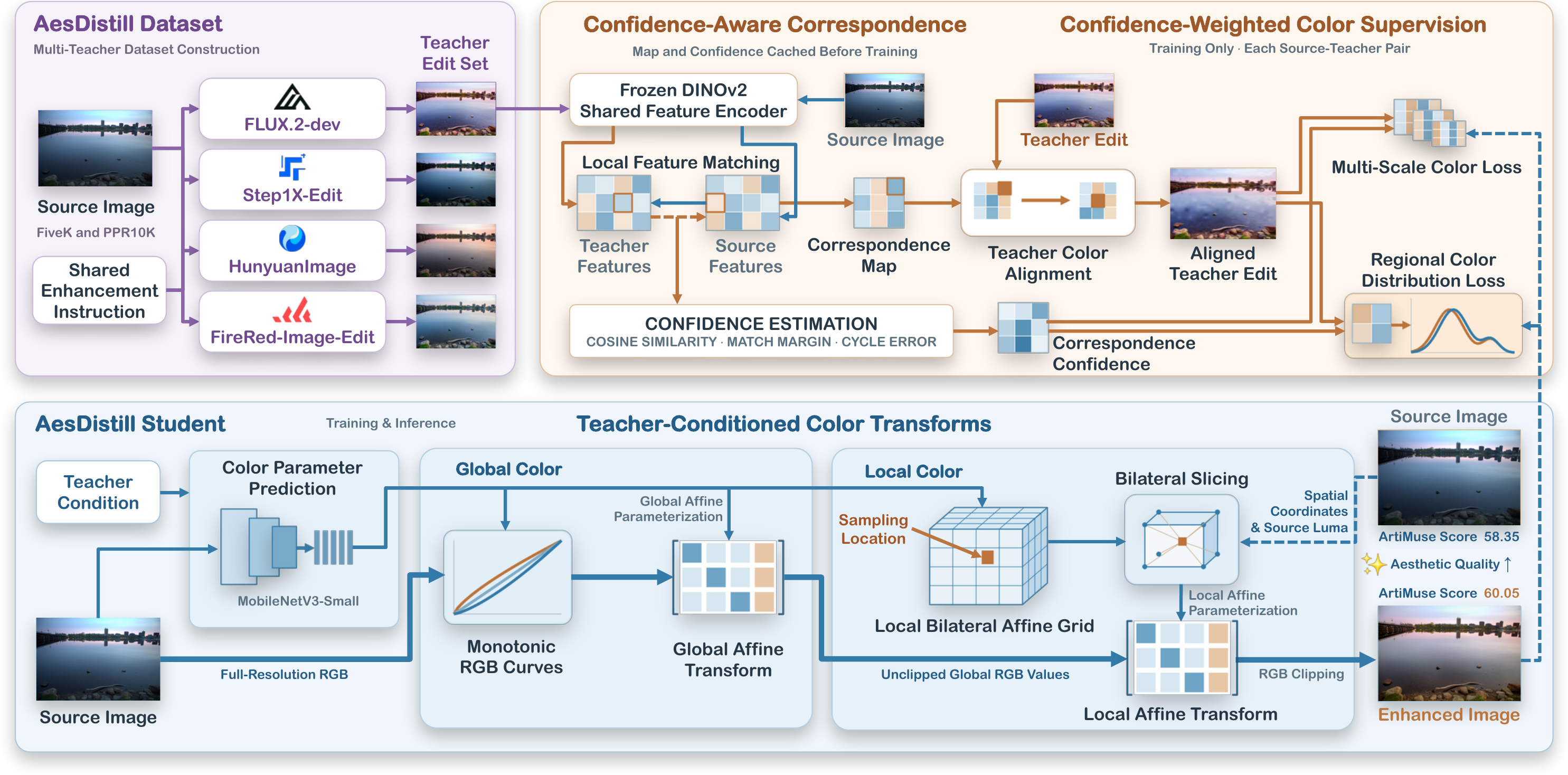}
\caption{\method uses confidence-weighted teacher supervision to learn color transforms that preserve source image structure.}
\label{fig:pipeline}
\end{figure*}

\section{Methodology}
\label{sec:methodology}

Figure~\ref{fig:pipeline} outlines \studentname. Given a source RGB image $I$ in $[0,1]$ and a teacher condition $t$ selecting one of four teachers, the student predicts the enhanced image $\hat I_t$. During training, \correspondencemodule matches patches in $I$ and the teacher-edited image $J_t$ and assigns confidence weights. The \transformmodule module predicts global and local color transforms from $I$ and $t$. \supervisionmodule trains these transforms using aligned teacher colors and regional color distributions. All four conditions share one student, which requires only $I$ and $t$ at inference.

\subsection{\correspondencemodule}
\label{sec:correspondence}

Generative edits can shift content while changing colors. Therefore, local feature matching pairs source patches with patches in the teacher-edited image $J_t$. Frozen DINOv2~\cite{dinov2} extracts unit-normalized patch features $f_I$ and $f_{J_t}$ from $I$ and $J_t$, respectively. For each source feature-grid position $p$, cosine similarity selects the best match within a neighborhood $\mathcal N(p)$ around $p$ in the teacher grid:
\begin{equation}
\phi_t(p)=\mathop{\arg\max}_{q\in\mathcal N(p)}
f_I(p)^\top f_{J_t}(q),
\label{eq:matching}
\end{equation}
where $\phi_t(p)$ is the matched teacher-grid position. Reverse matching maps that position back to the source grid. The distance from the returned position to $p$ defines the cycle error. Confidence $w_t(p)$ combines the best cosine similarity, the gap to the second-best similarity, and the cycle error to downweight weak, ambiguous, or inconsistent matches. The map $\phi_t$ and weights $w_t$ are cached before training.

For multi-scale supervision, the difference between matched teacher and source positions gives a displacement field in normalized coordinates. At each supervision scale, this field is resized bilinearly and added to the normalized source pixel coordinates to sample colors from the resized teacher image. Confidence weights use nearest-neighbor resizing. Thus, teacher colors are aligned to the student's pixel positions. Appendix~\ref{app:correspondence} gives the confidence function and coordinate conversion.

\subsection{\transformmodule}
\label{sec:color-transforms}

\begin{figure*}[t]
\centering
\includegraphics[width=\textwidth]{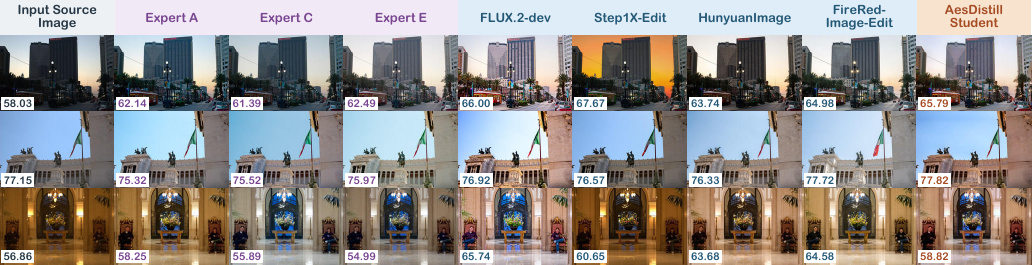}
\caption{Generative model teachers and \studentname surpass expert edits in aesthetic quality on FiveK (ArtiMuse scores $\uparrow$).}
\label{fig:expert-teacher-qualitative}
\end{figure*}

A learned teacher embedding $e_t$ encodes condition $t$ for global and local parameter prediction. A MobileNetV3-Small encoder~\cite{mobilenetv3} extracts features from a low-resolution source image preview. Pooled encoder features, RGB statistics from the preview, and $e_t$ form a context vector that feeds the global prediction head and conditions a multi-scale decoder for local parameters. The global head predicts Monotonic RGB Curves $\mathcal C$ and a Global Affine Transform $A_g$. The decoder's local head outputs a Local Bilateral Affine Grid $\mathcal B$ of raw coefficients.

\textbfnoindent{Global Color.}
The curves express nonlinear tone changes. For channel $c$, logits $z_{c,k}$ define $K$ positive increments:
\begin{equation}
\Delta_{c,k}=(1-\kappa)\operatorname{softmax}(z_c)_k+\kappa/K,
\label{eq:curves}
\end{equation}
where $0<\kappa<1$ prevents flat curve segments. Cumulative sums give knots from zero to one. Linear interpolation between knots gives curves that preserve intensity order within each channel. $A_g$ then applies positive diagonal gains, bounded channel mixing, and offsets. Zero-initialized heads make the curves and affine transforms identity mappings, so the initial output equals $I$. Appendix~\ref{app:color-transforms} details the renderer.

\textbfnoindent{Local Color.}
The grid $\mathcal B$ adds regional corrections to the global color transform. At each full-resolution source pixel $p=(x,y)$, Bilateral Slicing~\cite{hdrnet} interpolates raw grid coefficients along the spatial coordinates $(x,y)$ and source luma $\ell_I(p)$, a weighted RGB brightness guide. The sampled coefficients are then converted to a bounded Local Affine Transform $A_\ell(p)$. Writing $\bar v=[v^\top,1]^\top$ to append a constant to an RGB vector, Full-Resolution Rendering is:
\begin{equation}
\begin{aligned}
u_t(p)&=A_g\,\overline{\mathcal C(I(p))},\\
v_t(p)&=A_\ell(p)\,\overline{u_t(p)},\qquad
\hat I_t(p)=\operatorname{clip}_{[0,1]}(v_t(p)),
\end{aligned}
\label{eq:render}
\end{equation}
where $u_t(p)$ and $v_t(p)$ are RGB values after the global and local transforms, respectively. All transforms act at the source pixel $p$, preserving source sampling positions. Neural encoding uses the low-resolution preview, so its cost is independent of output size. Only coefficient lookup and color arithmetic run at full resolution.

\subsection{\supervisionmodule}
\label{sec:color-supervision}

Two losses transfer teacher colors by comparing corresponding pixels and regional color distributions. At scale $s$, $\hat I_{t,s}$ is the resized, clipped student output, and $\widetilde J_{t,s}$ is the resized teacher image aligned using Section~\ref{sec:correspondence}. Confidence weights $w_{t,s}$ exclude low-confidence and out-of-bounds matches. Both losses use the robust penalty $\rho(a)=\sqrt{a^2+\epsilon^2}-\epsilon$, where $\epsilon>0$ controls smoothing near zero.

\textbfnoindent{Multi-Scale Color Loss.}
Multi-scale color supervision compares $\hat I_{t,s}$ and $\widetilde J_{t,s}$ at each resolution $s\in\mathcal S$:
\begin{equation}
\mathcal L_{\mathrm{col}}=
\frac{1}{|\mathcal S|}\sum_{s\in\mathcal S}
\frac{\sum_{p,c}w_{t,s}(p)\rho(\hat I_{t,s,c}(p)-\widetilde J_{t,s,c}(p))}
{3\max\!\left(\sum_p w_{t,s}(p),\eta|\Omega_s|\right)},
\label{eq:color-loss}
\end{equation}
where $\Omega_s$ is the pixel grid at resolution $s$ and $c$ indexes RGB channels. Coarser views emphasize overall tone, while finer views constrain local color differences. The coverage floor $\eta\in(0,1]$ prevents a few reliable matches from receiving excessive weight.

\textbfnoindent{Regional Color Distribution Loss.}
This loss complements pixelwise color supervision by matching regional color balance without requiring individual color pairs to agree. Regional partitioning divides the resized student output and aligned teacher image into the same fixed regions $\mathcal R$. Following sliced distribution matching~\cite{swd}, RGB colors in each region are projected onto fixed unit directions $d$. Confidence-weighted quantiles $Q^{\hat I}_{r,d}(\tau)$ and $Q^{\widetilde J}_{r,d}(\tau)$ summarize these projections at level $\tau$ in region $r$. Both images share sampling positions and confidence weights. Their projected colors are sorted independently and compared at matching quantile levels. Averaging over directions and quantile levels gives:
\begin{equation}
\mathcal L_{\mathrm{dist}}=
\frac{1}{|\mathcal R|}\sum_{r\in\mathcal R}\gamma_r
\underset{d,\tau}{\operatorname{mean}}
\rho\!\left(Q^{\hat I}_{r,d}(\tau)-Q^{\widetilde J}_{r,d}(\tau)\right),
\label{eq:distribution-loss}
\end{equation}
where $P_r$ contains the sampled positions in region $r$ and $w_{t,\mathrm{dist}}$ denotes their confidence weights at the distribution resolution. The factor $\gamma_r=\min(1,\sum_{p\in P_r}w_{t,\mathrm{dist}}(p)/(\eta|P_r|))$ downweights regions with low confidence, giving zero contribution when all weights are zero. Shared spatial partitions preserve coarse color placement. Appendix~\ref{app:color-supervision} details projection and quantile computation.

\begin{table*}[t]
\centering
\caption{Scores ($\uparrow$) and ranks on FiveK and PPR10K. Model and input scores average both datasets equally, while expert rows use the named dataset. Ranks are in parentheses. Bold marks the best enhancement method per column. Evaluation settings follow Appendix~\ref{app:ranking}.}
\label{tab:assessment-joint}
\begingroup
\definecolor{assessmentblue}{HTML}{245876}
\definecolor{assessmentpurple}{HTML}{794690}
\definecolor{assessmentorange}{HTML}{A64C24}
\definecolor{assessmentgreen}{HTML}{426C55}
\definecolor{assessmentgreendark}{HTML}{D3E4D8}
\definecolor{assessmentgreenlight}{HTML}{E1EDE3}
\definecolor{assessmentbluedark}{HTML}{D8E9F5}
\definecolor{assessmentbluelight}{HTML}{F0F7FC}
\definecolor{assessmentpurpledark}{HTML}{D5BFE5}
\definecolor{assessmentpurplelight}{HTML}{E4D6EF}
\definecolor{assessmentorangedark}{HTML}{F9D9BC}
\definecolor{assessmentorangelight}{HTML}{FDF0E4}
\definecolor{assessmentgraydark}{HTML}{E5E8EB}
\definecolor{assessmentgraylight}{HTML}{F5F6F8}
\fontsize{7}{9}\selectfont
\setlength{\tabcolsep}{0.05pt}
\renewcommand{\arraystretch}{1.12}
\begin{tabular*}{\textwidth}{@{\extracolsep{\fill}}l|>{\hspace{1pt}}r<{\hspace{1pt}}|*{16}{r}@{}}
\toprule
\multicolumn{2}{c}{} & \multicolumn{12}{c}{Image Aesthetics (12)} & \multicolumn{4}{c}{Image Quality (4)} \\
\cmidrule(lr){3-14}\cmidrule(lr){15-18}
\multicolumn{1}{@{}l}{Method} & \multicolumn{1}{c}{Joint (\%)} & \rotatebox{75}{NIMA~\cite{nima}} & \rotatebox{75}{MUSIQ-AVA~\cite{musiq}} & \rotatebox{75}{TOPIQ-IAA~\cite{topiq}} & \rotatebox{75}{Q-Align~\cite{qalign}} & \rotatebox{75}{ArtiMuse~\cite{artimuse}} & \rotatebox{75}{ArtQuant~\cite{artquant}} & \rotatebox{75}{NIMA-VGG16~\cite{nima}} & \rotatebox{75}{MLSP-AVA~\cite{mlsp}} & \rotatebox{75}{TANet-AVA~\cite{tanet}} & \rotatebox{75}{TOPIQ-Res50~\cite{topiq}} & \rotatebox{75}{LAION-AES~\cite{laion_aes}} & \rotatebox{75}{ArtiMuse-AVA~\cite{artimuse}} & \rotatebox{75}{MANIQA~\cite{maniqa}} & \rotatebox{75}{LIQE~\cite{liqe}} & \rotatebox{75}{TOPIQ-NR~\cite{topiq}} & \rotatebox{75}{CLIP-IQA~\cite{clipiqa}} \\
\midrule
\cellcolor{assessmentbluelight}\textcolor{assessmentblue}{FLUX.2-dev~\cite{flux2}} & \cellcolor{assessmentbluelight}96.62{\fontsize{6}{7}\selectfont(1)} & \cellcolor{assessmentbluedark}5.49{\fontsize{6}{7}\selectfont(1)} & \cellcolor{assessmentbluelight}5.36{\fontsize{6}{7}\selectfont(1)} & \cellcolor{assessmentbluedark}5.48{\fontsize{6}{7}\selectfont(1)} & \cellcolor{assessmentbluelight}3.34{\fontsize{6}{7}\selectfont(1)} & \cellcolor{assessmentbluedark}61.20{\fontsize{6}{7}\selectfont(1)} & \cellcolor{assessmentbluelight}3.93{\fontsize{6}{7}\selectfont(1)} & \cellcolor{assessmentbluedark}5.53{\fontsize{6}{7}\selectfont(1)} & \cellcolor{assessmentbluelight}5.47{\fontsize{6}{7}\selectfont(1)} & \cellcolor{assessmentbluedark}5.40{\fontsize{6}{7}\selectfont(1)} & \cellcolor{assessmentbluelight}5.60{\fontsize{6}{7}\selectfont(1)} & \cellcolor{assessmentbluedark}5.54{\fontsize{6}{7}\selectfont(2)} & \cellcolor{assessmentbluelight}54.68{\fontsize{6}{7}\selectfont(1)} & \cellcolor{assessmentbluedark}0.436{\fontsize{6}{7}\selectfont(1)} & \cellcolor{assessmentbluelight}4.606{\fontsize{6}{7}\selectfont(1)} & \cellcolor{assessmentbluedark}0.663{\fontsize{6}{7}\selectfont(1)} & \cellcolor{assessmentbluelight}0.682{\fontsize{6}{7}\selectfont(1)} \\
\cellcolor{assessmentbluelight}\textcolor{assessmentblue}{FireRed-Image-Edit~\cite{firered}} & \cellcolor{assessmentbluelight}92.75{\fontsize{6}{7}\selectfont(2)} & \cellcolor{assessmentbluedark}5.44{\fontsize{6}{7}\selectfont(2)} & \cellcolor{assessmentbluelight}5.15{\fontsize{6}{7}\selectfont(3)} & \cellcolor{assessmentbluedark}5.42{\fontsize{6}{7}\selectfont(2)} & \cellcolor{assessmentbluelight}3.34{\fontsize{6}{7}\selectfont(2)} & \cellcolor{assessmentbluedark}59.84{\fontsize{6}{7}\selectfont(5)} & \cellcolor{assessmentbluelight}3.91{\fontsize{6}{7}\selectfont(2)} & \cellcolor{assessmentbluedark}5.50{\fontsize{6}{7}\selectfont(2)} & \cellcolor{assessmentbluelight}5.27{\fontsize{6}{7}\selectfont(3)} & \cellcolor{assessmentbluedark}5.37{\fontsize{6}{7}\selectfont(2)} & \cellcolor{assessmentbluelight}5.47{\fontsize{6}{7}\selectfont(3)} & \cellcolor{assessmentbluedark}5.55{\fontsize{6}{7}\selectfont(1)} & \cellcolor{assessmentbluelight}54.28{\fontsize{6}{7}\selectfont(2)} & \cellcolor{assessmentbluedark}0.397{\fontsize{6}{7}\selectfont(2)} & \cellcolor{assessmentbluelight}4.510{\fontsize{6}{7}\selectfont(2)} & \cellcolor{assessmentbluedark}0.591{\fontsize{6}{7}\selectfont(2)} & \cellcolor{assessmentbluelight}0.530{\fontsize{6}{7}\selectfont(3)} \\
\cellcolor{assessmentbluelight}\textcolor{assessmentblue}{HunyuanImage~\cite{hunyuan}} & \cellcolor{assessmentbluelight}90.75{\fontsize{6}{7}\selectfont(3)} & \cellcolor{assessmentbluedark}5.32{\fontsize{6}{7}\selectfont(3)} & \cellcolor{assessmentbluelight}5.17{\fontsize{6}{7}\selectfont(2)} & \cellcolor{assessmentbluedark}5.39{\fontsize{6}{7}\selectfont(3)} & \cellcolor{assessmentbluelight}3.28{\fontsize{6}{7}\selectfont(3)} & \cellcolor{assessmentbluedark}60.07{\fontsize{6}{7}\selectfont(4)} & \cellcolor{assessmentbluelight}3.90{\fontsize{6}{7}\selectfont(3)} & \cellcolor{assessmentbluedark}5.40{\fontsize{6}{7}\selectfont(3)} & \cellcolor{assessmentbluelight}5.32{\fontsize{6}{7}\selectfont(2)} & \cellcolor{assessmentbluedark}5.31{\fontsize{6}{7}\selectfont(3)} & \cellcolor{assessmentbluelight}5.48{\fontsize{6}{7}\selectfont(2)} & \cellcolor{assessmentbluedark}5.46{\fontsize{6}{7}\selectfont(3)} & \cellcolor{assessmentbluelight}53.97{\fontsize{6}{7}\selectfont(3)} & \cellcolor{assessmentbluedark}0.386{\fontsize{6}{7}\selectfont(3)} & \cellcolor{assessmentbluelight}4.473{\fontsize{6}{7}\selectfont(3)} & \cellcolor{assessmentbluedark}0.577{\fontsize{6}{7}\selectfont(3)} & \cellcolor{assessmentbluelight}0.614{\fontsize{6}{7}\selectfont(2)} \\
\cellcolor{assessmentbluelight}\textcolor{assessmentblue}{Step1X-Edit~\cite{reasonedit}} & \cellcolor{assessmentbluelight}76.62{\fontsize{6}{7}\selectfont(4)} & \cellcolor{assessmentbluedark}5.20{\fontsize{6}{7}\selectfont(4)} & \cellcolor{assessmentbluelight}4.69{\fontsize{6}{7}\selectfont(5)} & \cellcolor{assessmentbluedark}5.19{\fontsize{6}{7}\selectfont(4)} & \cellcolor{assessmentbluelight}3.10{\fontsize{6}{7}\selectfont(4)} & \cellcolor{assessmentbluedark}58.17{\fontsize{6}{7}\selectfont(15)} & \cellcolor{assessmentbluelight}3.86{\fontsize{6}{7}\selectfont(8)} & \cellcolor{assessmentbluedark}5.35{\fontsize{6}{7}\selectfont(4)} & \cellcolor{assessmentbluelight}4.72{\fontsize{6}{7}\selectfont(4)} & \cellcolor{assessmentbluedark}5.28{\fontsize{6}{7}\selectfont(5)} & \cellcolor{assessmentbluelight}5.20{\fontsize{6}{7}\selectfont(4)} & \cellcolor{assessmentbluedark}5.32{\fontsize{6}{7}\selectfont(7)} & \cellcolor{assessmentbluelight}51.75{\fontsize{6}{7}\selectfont(4)} & \cellcolor{assessmentbluedark}0.313{\fontsize{6}{7}\selectfont(4)} & \cellcolor{assessmentbluelight}3.105{\fontsize{6}{7}\selectfont(4)} & \cellcolor{assessmentbluedark}0.408{\fontsize{6}{7}\selectfont(4)} & \cellcolor{assessmentbluelight}0.442{\fontsize{6}{7}\selectfont(24)} \\
\specialrule{0.4pt}{1pt}{1pt}
\cellcolor{assessmentorangelight}\textcolor{assessmentorange}{\textbf{AesDistill Student}} & \cellcolor{assessmentorangelight}\textcolor{assessmentorange}{\textbf{74.75}{\fontsize{6}{7}\selectfont(5)}} & \cellcolor{assessmentorangedark}\textcolor{assessmentorange}{5.16{\fontsize{6}{7}\selectfont(6)}} & \cellcolor{assessmentorangelight}\textcolor{assessmentorange}{4.64{\fontsize{6}{7}\selectfont(8)}} & \cellcolor{assessmentorangedark}\textcolor{assessmentorange}{\textbf{5.17}{\fontsize{6}{7}\selectfont(5)}} & \cellcolor{assessmentorangelight}\textcolor{assessmentorange}{\textbf{3.04}{\fontsize{6}{7}\selectfont(5)}} & \cellcolor{assessmentorangedark}\textcolor{assessmentorange}{58.74{\fontsize{6}{7}\selectfont(10)}} & \cellcolor{assessmentorangelight}\textcolor{assessmentorange}{3.86{\fontsize{6}{7}\selectfont(6)}} & \cellcolor{assessmentorangedark}\textcolor{assessmentorange}{\textbf{5.31}{\fontsize{6}{7}\selectfont(5)}} & \cellcolor{assessmentorangelight}\textcolor{assessmentorange}{4.57{\fontsize{6}{7}\selectfont(10)}} & \cellcolor{assessmentorangedark}\textcolor{assessmentorange}{\textbf{5.29}{\fontsize{6}{7}\selectfont(4)}} & \cellcolor{assessmentorangelight}\textcolor{assessmentorange}{5.14{\fontsize{6}{7}\selectfont(7)}} & \cellcolor{assessmentorangedark}\textcolor{assessmentorange}{\textbf{5.37}{\fontsize{6}{7}\selectfont(4)}} & \cellcolor{assessmentorangelight}\textcolor{assessmentorange}{\textbf{51.35}{\fontsize{6}{7}\selectfont(5)}} & \cellcolor{assessmentorangedark}\textcolor{assessmentorange}{0.298{\fontsize{6}{7}\selectfont(8)}} & \cellcolor{assessmentorangelight}\textcolor{assessmentorange}{\textbf{2.877}{\fontsize{6}{7}\selectfont(5)}} & \cellcolor{assessmentorangedark}\textcolor{assessmentorange}{0.379{\fontsize{6}{7}\selectfont(6)}} & \cellcolor{assessmentorangelight}\textcolor{assessmentorange}{0.474{\fontsize{6}{7}\selectfont(15)}} \\
\specialrule{0.4pt}{1pt}{1pt}
\cellcolor{assessmentgraylight}DeepLPF~\cite{deeplpf} & \cellcolor{assessmentgraylight}67.50{\fontsize{6}{7}\selectfont(6)} & \cellcolor{assessmentgraydark}5.16{\fontsize{6}{7}\selectfont(7)} & \cellcolor{assessmentgraylight}4.64{\fontsize{6}{7}\selectfont(7)} & \cellcolor{assessmentgraydark}5.13{\fontsize{6}{7}\selectfont(6)} & \cellcolor{assessmentgraylight}3.01{\fontsize{6}{7}\selectfont(6)} & \cellcolor{assessmentgraydark}58.48{\fontsize{6}{7}\selectfont(11)} & \cellcolor{assessmentgraylight}3.84{\fontsize{6}{7}\selectfont(17)} & \cellcolor{assessmentgraydark}5.29{\fontsize{6}{7}\selectfont(7)} & \cellcolor{assessmentgraylight}4.57{\fontsize{6}{7}\selectfont(9)} & \cellcolor{assessmentgraydark}5.25{\fontsize{6}{7}\selectfont(8)} & \cellcolor{assessmentgraylight}5.15{\fontsize{6}{7}\selectfont(6)} & \cellcolor{assessmentgraydark}5.29{\fontsize{6}{7}\selectfont(15)} & \cellcolor{assessmentgraylight}51.10{\fontsize{6}{7}\selectfont(6)} & \cellcolor{assessmentgraydark}\textbf{0.301}{\fontsize{6}{7}\selectfont(5)} & \cellcolor{assessmentgraylight}2.775{\fontsize{6}{7}\selectfont(10)} & \cellcolor{assessmentgraydark}\textbf{0.381}{\fontsize{6}{7}\selectfont(5)} & \cellcolor{assessmentgraylight}0.484{\fontsize{6}{7}\selectfont(8)} \\
\cellcolor{assessmentgraylight}HDRNet~\cite{hdrnet} & \cellcolor{assessmentgraylight}64.12{\fontsize{6}{7}\selectfont(7)} & \cellcolor{assessmentgraydark}\textbf{5.17}{\fontsize{6}{7}\selectfont(5)} & \cellcolor{assessmentgraylight}\textbf{4.69}{\fontsize{6}{7}\selectfont(4)} & \cellcolor{assessmentgraydark}5.12{\fontsize{6}{7}\selectfont(8)} & \cellcolor{assessmentgraylight}2.95{\fontsize{6}{7}\selectfont(18)} & \cellcolor{assessmentgraydark}60.51{\fontsize{6}{7}\selectfont(3)} & \cellcolor{assessmentgraylight}3.89{\fontsize{6}{7}\selectfont(5)} & \cellcolor{assessmentgraydark}5.31{\fontsize{6}{7}\selectfont(6)} & \cellcolor{assessmentgraylight}\textbf{4.62}{\fontsize{6}{7}\selectfont(5)} & \cellcolor{assessmentgraydark}5.25{\fontsize{6}{7}\selectfont(9)} & \cellcolor{assessmentgraylight}\textbf{5.15}{\fontsize{6}{7}\selectfont(5)} & \cellcolor{assessmentgraydark}5.27{\fontsize{6}{7}\selectfont(22)} & \cellcolor{assessmentgraylight}50.79{\fontsize{6}{7}\selectfont(12)} & \cellcolor{assessmentgraydark}0.292{\fontsize{6}{7}\selectfont(18)} & \cellcolor{assessmentgraylight}2.594{\fontsize{6}{7}\selectfont(18)} & \cellcolor{assessmentgraydark}0.374{\fontsize{6}{7}\selectfont(16)} & \cellcolor{assessmentgraylight}\textbf{0.492}{\fontsize{6}{7}\selectfont(5)} \\
\cellcolor{assessmentgraylight}RSFNet~\cite{rsfnet} & \cellcolor{assessmentgraylight}64.00{\fontsize{6}{7}\selectfont(8)} & \cellcolor{assessmentgraydark}5.13{\fontsize{6}{7}\selectfont(9)} & \cellcolor{assessmentgraylight}4.63{\fontsize{6}{7}\selectfont(10)} & \cellcolor{assessmentgraydark}5.12{\fontsize{6}{7}\selectfont(9)} & \cellcolor{assessmentgraylight}3.00{\fontsize{6}{7}\selectfont(9)} & \cellcolor{assessmentgraydark}58.08{\fontsize{6}{7}\selectfont(17)} & \cellcolor{assessmentgraylight}3.84{\fontsize{6}{7}\selectfont(18)} & \cellcolor{assessmentgraydark}5.28{\fontsize{6}{7}\selectfont(8)} & \cellcolor{assessmentgraylight}4.58{\fontsize{6}{7}\selectfont(6)} & \cellcolor{assessmentgraydark}5.25{\fontsize{6}{7}\selectfont(7)} & \cellcolor{assessmentgraylight}5.12{\fontsize{6}{7}\selectfont(9)} & \cellcolor{assessmentgraydark}5.31{\fontsize{6}{7}\selectfont(9)} & \cellcolor{assessmentgraylight}50.94{\fontsize{6}{7}\selectfont(10)} & \cellcolor{assessmentgraydark}0.298{\fontsize{6}{7}\selectfont(9)} & \cellcolor{assessmentgraylight}2.762{\fontsize{6}{7}\selectfont(11)} & \cellcolor{assessmentgraydark}0.377{\fontsize{6}{7}\selectfont(10)} & \cellcolor{assessmentgraylight}0.482{\fontsize{6}{7}\selectfont(10)} \\
\cellcolor{assessmentgraylight}DiffRetouch~\cite{diffretouch} & \cellcolor{assessmentgraylight}63.88{\fontsize{6}{7}\selectfont(9)} & \cellcolor{assessmentgraydark}5.12{\fontsize{6}{7}\selectfont(11)} & \cellcolor{assessmentgraylight}4.61{\fontsize{6}{7}\selectfont(12)} & \cellcolor{assessmentgraydark}5.13{\fontsize{6}{7}\selectfont(7)} & \cellcolor{assessmentgraylight}3.01{\fontsize{6}{7}\selectfont(7)} & \cellcolor{assessmentgraydark}57.98{\fontsize{6}{7}\selectfont(18)} & \cellcolor{assessmentgraylight}3.84{\fontsize{6}{7}\selectfont(14)} & \cellcolor{assessmentgraydark}5.27{\fontsize{6}{7}\selectfont(10)} & \cellcolor{assessmentgraylight}4.56{\fontsize{6}{7}\selectfont(11)} & \cellcolor{assessmentgraydark}5.25{\fontsize{6}{7}\selectfont(6)} & \cellcolor{assessmentgraylight}5.11{\fontsize{6}{7}\selectfont(10)} & \cellcolor{assessmentgraydark}5.33{\fontsize{6}{7}\selectfont(5)} & \cellcolor{assessmentgraylight}50.99{\fontsize{6}{7}\selectfont(7)} & \cellcolor{assessmentgraydark}0.297{\fontsize{6}{7}\selectfont(12)} & \cellcolor{assessmentgraylight}2.841{\fontsize{6}{7}\selectfont(7)} & \cellcolor{assessmentgraydark}0.377{\fontsize{6}{7}\selectfont(8)} & \cellcolor{assessmentgraylight}0.479{\fontsize{6}{7}\selectfont(12)} \\
\cellcolor{assessmentpurplelight}\textcolor{assessmentpurple}{PPR10K expert mean~\cite{ppr10k}} & \cellcolor{assessmentpurplelight}63.25{\fontsize{6}{7}\selectfont(10)} & \cellcolor{assessmentpurpledark}5.35{\fontsize{6}{7}\selectfont(14)} & \cellcolor{assessmentpurplelight}4.75{\fontsize{6}{7}\selectfont(17)} & \cellcolor{assessmentpurpledark}5.38{\fontsize{6}{7}\selectfont(10)} & \cellcolor{assessmentpurplelight}3.42{\fontsize{6}{7}\selectfont(8)} & \cellcolor{assessmentpurpledark}57.80{\fontsize{6}{7}\selectfont(22)} & \cellcolor{assessmentpurplelight}3.97{\fontsize{6}{7}\selectfont(10)} & \cellcolor{assessmentpurpledark}5.52{\fontsize{6}{7}\selectfont(10)} & \cellcolor{assessmentpurplelight}4.72{\fontsize{6}{7}\selectfont(18)} & \cellcolor{assessmentpurpledark}5.39{\fontsize{6}{7}\selectfont(9)} & \cellcolor{assessmentpurplelight}5.29{\fontsize{6}{7}\selectfont(16)} & \cellcolor{assessmentpurpledark}5.69{\fontsize{6}{7}\selectfont(4)} & \cellcolor{assessmentpurplelight}54.42{\fontsize{6}{7}\selectfont(12)} & \cellcolor{assessmentpurpledark}0.303{\fontsize{6}{7}\selectfont(5)} & \cellcolor{assessmentpurplelight}2.934{\fontsize{6}{7}\selectfont(6)} & \cellcolor{assessmentpurpledark}0.368{\fontsize{6}{7}\selectfont(11)} & \cellcolor{assessmentpurplelight}0.500{\fontsize{6}{7}\selectfont(4)} \\
\cellcolor{assessmentgraylight}BPAM~\cite{bpam} & \cellcolor{assessmentgraylight}60.88{\fontsize{6}{7}\selectfont(11)} & \cellcolor{assessmentgraydark}5.13{\fontsize{6}{7}\selectfont(10)} & \cellcolor{assessmentgraylight}4.61{\fontsize{6}{7}\selectfont(14)} & \cellcolor{assessmentgraydark}5.11{\fontsize{6}{7}\selectfont(11)} & \cellcolor{assessmentgraylight}3.00{\fontsize{6}{7}\selectfont(8)} & \cellcolor{assessmentgraydark}58.08{\fontsize{6}{7}\selectfont(16)} & \cellcolor{assessmentgraylight}3.84{\fontsize{6}{7}\selectfont(16)} & \cellcolor{assessmentgraydark}5.27{\fontsize{6}{7}\selectfont(9)} & \cellcolor{assessmentgraylight}4.55{\fontsize{6}{7}\selectfont(13)} & \cellcolor{assessmentgraydark}5.25{\fontsize{6}{7}\selectfont(10)} & \cellcolor{assessmentgraylight}5.10{\fontsize{6}{7}\selectfont(12)} & \cellcolor{assessmentgraydark}5.32{\fontsize{6}{7}\selectfont(6)} & \cellcolor{assessmentgraylight}50.95{\fontsize{6}{7}\selectfont(9)} & \cellcolor{assessmentgraydark}0.300{\fontsize{6}{7}\selectfont(6)} & \cellcolor{assessmentgraylight}2.857{\fontsize{6}{7}\selectfont(6)} & \cellcolor{assessmentgraydark}0.376{\fontsize{6}{7}\selectfont(12)} & \cellcolor{assessmentgraylight}0.485{\fontsize{6}{7}\selectfont(7)} \\
\cellcolor{assessmentpurplelight}\textcolor{assessmentpurple}{FiveK expert mean~\cite{fivek}} & \cellcolor{assessmentpurplelight}60.25{\fontsize{6}{7}\selectfont(12)} & \cellcolor{assessmentpurpledark}4.85{\fontsize{6}{7}\selectfont(17)} & \cellcolor{assessmentpurplelight}4.46{\fontsize{6}{7}\selectfont(15)} & \cellcolor{assessmentpurpledark}4.84{\fontsize{6}{7}\selectfont(12)} & \cellcolor{assessmentpurplelight}2.58{\fontsize{6}{7}\selectfont(10)} & \cellcolor{assessmentpurpledark}58.34{\fontsize{6}{7}\selectfont(14)} & \cellcolor{assessmentpurplelight}3.73{\fontsize{6}{7}\selectfont(12)} & \cellcolor{assessmentpurpledark}5.00{\fontsize{6}{7}\selectfont(17)} & \cellcolor{assessmentpurplelight}4.38{\fontsize{6}{7}\selectfont(10)} & \cellcolor{assessmentpurpledark}5.09{\fontsize{6}{7}\selectfont(17)} & \cellcolor{assessmentpurplelight}4.87{\fontsize{6}{7}\selectfont(17)} & \cellcolor{assessmentpurpledark}5.00{\fontsize{6}{7}\selectfont(12)} & \cellcolor{assessmentpurplelight}47.35{\fontsize{6}{7}\selectfont(11)} & \cellcolor{assessmentpurpledark}0.301{\fontsize{6}{7}\selectfont(6)} & \cellcolor{assessmentpurplelight}2.864{\fontsize{6}{7}\selectfont(5)} & \cellcolor{assessmentpurpledark}0.388{\fontsize{6}{7}\selectfont(7)} & \cellcolor{assessmentpurplelight}0.518{\fontsize{6}{7}\selectfont(4)} \\
\cellcolor{assessmentgraylight}Image-Adaptive 3D LUT~\cite{adaptive3dlut} & \cellcolor{assessmentgraylight}58.38{\fontsize{6}{7}\selectfont(13)} & \cellcolor{assessmentgraydark}5.14{\fontsize{6}{7}\selectfont(8)} & \cellcolor{assessmentgraylight}4.64{\fontsize{6}{7}\selectfont(6)} & \cellcolor{assessmentgraydark}5.11{\fontsize{6}{7}\selectfont(10)} & \cellcolor{assessmentgraylight}2.98{\fontsize{6}{7}\selectfont(13)} & \cellcolor{assessmentgraydark}58.84{\fontsize{6}{7}\selectfont(8)} & \cellcolor{assessmentgraylight}3.85{\fontsize{6}{7}\selectfont(9)} & \cellcolor{assessmentgraydark}5.27{\fontsize{6}{7}\selectfont(11)} & \cellcolor{assessmentgraylight}4.58{\fontsize{6}{7}\selectfont(7)} & \cellcolor{assessmentgraydark}5.23{\fontsize{6}{7}\selectfont(13)} & \cellcolor{assessmentgraylight}5.14{\fontsize{6}{7}\selectfont(8)} & \cellcolor{assessmentgraydark}5.28{\fontsize{6}{7}\selectfont(19)} & \cellcolor{assessmentgraylight}50.95{\fontsize{6}{7}\selectfont(8)} & \cellcolor{assessmentgraydark}0.292{\fontsize{6}{7}\selectfont(17)} & \cellcolor{assessmentgraylight}2.626{\fontsize{6}{7}\selectfont(17)} & \cellcolor{assessmentgraydark}0.378{\fontsize{6}{7}\selectfont(7)} & \cellcolor{assessmentgraylight}0.464{\fontsize{6}{7}\selectfont(21)} \\
\cellcolor{assessmentgraylight}AdaInt~\cite{adaint} & \cellcolor{assessmentgraylight}57.38{\fontsize{6}{7}\selectfont(14)} & \cellcolor{assessmentgraydark}5.12{\fontsize{6}{7}\selectfont(12)} & \cellcolor{assessmentgraylight}4.63{\fontsize{6}{7}\selectfont(9)} & \cellcolor{assessmentgraydark}5.10{\fontsize{6}{7}\selectfont(12)} & \cellcolor{assessmentgraylight}2.99{\fontsize{6}{7}\selectfont(11)} & \cellcolor{assessmentgraydark}58.29{\fontsize{6}{7}\selectfont(12)} & \cellcolor{assessmentgraylight}3.84{\fontsize{6}{7}\selectfont(12)} & \cellcolor{assessmentgraydark}5.27{\fontsize{6}{7}\selectfont(12)} & \cellcolor{assessmentgraylight}4.55{\fontsize{6}{7}\selectfont(12)} & \cellcolor{assessmentgraydark}5.24{\fontsize{6}{7}\selectfont(11)} & \cellcolor{assessmentgraylight}5.10{\fontsize{6}{7}\selectfont(11)} & \cellcolor{assessmentgraydark}5.30{\fontsize{6}{7}\selectfont(10)} & \cellcolor{assessmentgraylight}50.86{\fontsize{6}{7}\selectfont(11)} & \cellcolor{assessmentgraydark}0.297{\fontsize{6}{7}\selectfont(11)} & \cellcolor{assessmentgraylight}2.792{\fontsize{6}{7}\selectfont(9)} & \cellcolor{assessmentgraydark}0.376{\fontsize{6}{7}\selectfont(11)} & \cellcolor{assessmentgraylight}0.473{\fontsize{6}{7}\selectfont(16)} \\
\cellcolor{assessmentgraylight}VeraRetouch~\cite{veraretouch} & \cellcolor{assessmentgraylight}51.25{\fontsize{6}{7}\selectfont(15)} & \cellcolor{assessmentgraydark}5.10{\fontsize{6}{7}\selectfont(13)} & \cellcolor{assessmentgraylight}4.61{\fontsize{6}{7}\selectfont(13)} & \cellcolor{assessmentgraydark}5.10{\fontsize{6}{7}\selectfont(13)} & \cellcolor{assessmentgraylight}2.98{\fontsize{6}{7}\selectfont(12)} & \cellcolor{assessmentgraydark}58.74{\fontsize{6}{7}\selectfont(9)} & \cellcolor{assessmentgraylight}3.85{\fontsize{6}{7}\selectfont(10)} & \cellcolor{assessmentgraydark}5.25{\fontsize{6}{7}\selectfont(13)} & \cellcolor{assessmentgraylight}4.54{\fontsize{6}{7}\selectfont(17)} & \cellcolor{assessmentgraydark}5.23{\fontsize{6}{7}\selectfont(12)} & \cellcolor{assessmentgraylight}5.08{\fontsize{6}{7}\selectfont(14)} & \cellcolor{assessmentgraydark}5.31{\fontsize{6}{7}\selectfont(8)} & \cellcolor{assessmentgraylight}50.72{\fontsize{6}{7}\selectfont(15)} & \cellcolor{assessmentgraydark}0.295{\fontsize{6}{7}\selectfont(15)} & \cellcolor{assessmentgraylight}2.805{\fontsize{6}{7}\selectfont(8)} & \cellcolor{assessmentgraydark}0.373{\fontsize{6}{7}\selectfont(17)} & \cellcolor{assessmentgraylight}0.467{\fontsize{6}{7}\selectfont(18)} \\
\cellcolor{assessmentgraylight}SepLUT~\cite{seplut} & \cellcolor{assessmentgraylight}48.00{\fontsize{6}{7}\selectfont(16)} & \cellcolor{assessmentgraydark}5.10{\fontsize{6}{7}\selectfont(14)} & \cellcolor{assessmentgraylight}4.62{\fontsize{6}{7}\selectfont(11)} & \cellcolor{assessmentgraydark}5.09{\fontsize{6}{7}\selectfont(14)} & \cellcolor{assessmentgraylight}2.97{\fontsize{6}{7}\selectfont(14)} & \cellcolor{assessmentgraydark}58.20{\fontsize{6}{7}\selectfont(14)} & \cellcolor{assessmentgraylight}3.85{\fontsize{6}{7}\selectfont(11)} & \cellcolor{assessmentgraydark}5.25{\fontsize{6}{7}\selectfont(14)} & \cellcolor{assessmentgraylight}4.55{\fontsize{6}{7}\selectfont(14)} & \cellcolor{assessmentgraydark}5.22{\fontsize{6}{7}\selectfont(14)} & \cellcolor{assessmentgraylight}5.09{\fontsize{6}{7}\selectfont(13)} & \cellcolor{assessmentgraydark}5.30{\fontsize{6}{7}\selectfont(11)} & \cellcolor{assessmentgraylight}50.77{\fontsize{6}{7}\selectfont(13)} & \cellcolor{assessmentgraydark}0.295{\fontsize{6}{7}\selectfont(14)} & \cellcolor{assessmentgraylight}2.729{\fontsize{6}{7}\selectfont(13)} & \cellcolor{assessmentgraydark}0.377{\fontsize{6}{7}\selectfont(9)} & \cellcolor{assessmentgraylight}0.465{\fontsize{6}{7}\selectfont(20)} \\
\cellcolor{assessmentgraylight}NamedCurves~\cite{namedcurves} & \cellcolor{assessmentgraylight}44.62{\fontsize{6}{7}\selectfont(17)} & \cellcolor{assessmentgraydark}5.09{\fontsize{6}{7}\selectfont(16)} & \cellcolor{assessmentgraylight}4.59{\fontsize{6}{7}\selectfont(19)} & \cellcolor{assessmentgraydark}5.08{\fontsize{6}{7}\selectfont(15)} & \cellcolor{assessmentgraylight}2.97{\fontsize{6}{7}\selectfont(16)} & \cellcolor{assessmentgraydark}59.62{\fontsize{6}{7}\selectfont(6)} & \cellcolor{assessmentgraylight}3.84{\fontsize{6}{7}\selectfont(13)} & \cellcolor{assessmentgraydark}5.23{\fontsize{6}{7}\selectfont(15)} & \cellcolor{assessmentgraylight}4.53{\fontsize{6}{7}\selectfont(18)} & \cellcolor{assessmentgraydark}5.22{\fontsize{6}{7}\selectfont(17)} & \cellcolor{assessmentgraylight}5.07{\fontsize{6}{7}\selectfont(15)} & \cellcolor{assessmentgraydark}5.29{\fontsize{6}{7}\selectfont(12)} & \cellcolor{assessmentgraylight}50.73{\fontsize{6}{7}\selectfont(14)} & \cellcolor{assessmentgraydark}0.294{\fontsize{6}{7}\selectfont(16)} & \cellcolor{assessmentgraylight}2.713{\fontsize{6}{7}\selectfont(14)} & \cellcolor{assessmentgraydark}0.374{\fontsize{6}{7}\selectfont(14)} & \cellcolor{assessmentgraylight}0.479{\fontsize{6}{7}\selectfont(11)} \\
\cellcolor{assessmentgraylight}Pigment~\cite{pigment} & \cellcolor{assessmentgraylight}40.62{\fontsize{6}{7}\selectfont(18)} & \cellcolor{assessmentgraydark}5.07{\fontsize{6}{7}\selectfont(19)} & \cellcolor{assessmentgraylight}4.60{\fontsize{6}{7}\selectfont(15)} & \cellcolor{assessmentgraydark}5.08{\fontsize{6}{7}\selectfont(16)} & \cellcolor{assessmentgraylight}2.97{\fontsize{6}{7}\selectfont(15)} & \cellcolor{assessmentgraydark}57.88{\fontsize{6}{7}\selectfont(19)} & \cellcolor{assessmentgraylight}3.84{\fontsize{6}{7}\selectfont(19)} & \cellcolor{assessmentgraydark}5.22{\fontsize{6}{7}\selectfont(17)} & \cellcolor{assessmentgraylight}4.53{\fontsize{6}{7}\selectfont(20)} & \cellcolor{assessmentgraydark}5.22{\fontsize{6}{7}\selectfont(18)} & \cellcolor{assessmentgraylight}5.07{\fontsize{6}{7}\selectfont(16)} & \cellcolor{assessmentgraydark}5.29{\fontsize{6}{7}\selectfont(16)} & \cellcolor{assessmentgraylight}50.68{\fontsize{6}{7}\selectfont(16)} & \cellcolor{assessmentgraydark}0.296{\fontsize{6}{7}\selectfont(13)} & \cellcolor{assessmentgraylight}2.730{\fontsize{6}{7}\selectfont(12)} & \cellcolor{assessmentgraydark}0.374{\fontsize{6}{7}\selectfont(15)} & \cellcolor{assessmentgraylight}0.470{\fontsize{6}{7}\selectfont(17)} \\
\cellcolor{assessmentgraylight}MonetGPT~\cite{monetgpt} & \cellcolor{assessmentgraylight}40.25{\fontsize{6}{7}\selectfont(19)} & \cellcolor{assessmentgraydark}5.08{\fontsize{6}{7}\selectfont(17)} & \cellcolor{assessmentgraylight}4.57{\fontsize{6}{7}\selectfont(20)} & \cellcolor{assessmentgraydark}5.08{\fontsize{6}{7}\selectfont(17)} & \cellcolor{assessmentgraylight}2.96{\fontsize{6}{7}\selectfont(17)} & \cellcolor{assessmentgraydark}57.70{\fontsize{6}{7}\selectfont(21)} & \cellcolor{assessmentgraylight}3.83{\fontsize{6}{7}\selectfont(22)} & \cellcolor{assessmentgraydark}5.21{\fontsize{6}{7}\selectfont(20)} & \cellcolor{assessmentgraylight}4.54{\fontsize{6}{7}\selectfont(16)} & \cellcolor{assessmentgraydark}5.22{\fontsize{6}{7}\selectfont(15)} & \cellcolor{assessmentgraylight}5.06{\fontsize{6}{7}\selectfont(18)} & \cellcolor{assessmentgraydark}5.29{\fontsize{6}{7}\selectfont(17)} & \cellcolor{assessmentgraylight}50.60{\fontsize{6}{7}\selectfont(17)} & \cellcolor{assessmentgraydark}0.297{\fontsize{6}{7}\selectfont(10)} & \cellcolor{assessmentgraylight}2.707{\fontsize{6}{7}\selectfont(15)} & \cellcolor{assessmentgraydark}0.375{\fontsize{6}{7}\selectfont(13)} & \cellcolor{assessmentgraylight}0.483{\fontsize{6}{7}\selectfont(9)} \\
\cellcolor{assessmentgraylight}SA-LUT~\cite{salut} & \cellcolor{assessmentgraylight}37.62{\fontsize{6}{7}\selectfont(20)} & \cellcolor{assessmentgraydark}5.08{\fontsize{6}{7}\selectfont(18)} & \cellcolor{assessmentgraylight}4.57{\fontsize{6}{7}\selectfont(21)} & \cellcolor{assessmentgraydark}4.99{\fontsize{6}{7}\selectfont(23)} & \cellcolor{assessmentgraylight}2.86{\fontsize{6}{7}\selectfont(23)} & \cellcolor{assessmentgraydark}\textbf{60.89}{\fontsize{6}{7}\selectfont(2)} & \cellcolor{assessmentgraylight}\textbf{3.89}{\fontsize{6}{7}\selectfont(4)} & \cellcolor{assessmentgraydark}5.18{\fontsize{6}{7}\selectfont(21)} & \cellcolor{assessmentgraylight}4.53{\fontsize{6}{7}\selectfont(19)} & \cellcolor{assessmentgraydark}5.19{\fontsize{6}{7}\selectfont(22)} & \cellcolor{assessmentgraylight}5.05{\fontsize{6}{7}\selectfont(19)} & \cellcolor{assessmentgraydark}5.29{\fontsize{6}{7}\selectfont(14)} & \cellcolor{assessmentgraylight}49.49{\fontsize{6}{7}\selectfont(24)} & \cellcolor{assessmentgraydark}0.265{\fontsize{6}{7}\selectfont(26)} & \cellcolor{assessmentgraylight}2.014{\fontsize{6}{7}\selectfont(25)} & \cellcolor{assessmentgraydark}0.344{\fontsize{6}{7}\selectfont(25)} & \cellcolor{assessmentgraylight}0.446{\fontsize{6}{7}\selectfont(23)} \\
\cellcolor{assessmentgreenlight}\textcolor{assessmentgreen}{Input Source Image} & \cellcolor{assessmentgreenlight}35.88{\fontsize{6}{7}\selectfont(21)} & \cellcolor{assessmentgreendark}5.06{\fontsize{6}{7}\selectfont(20)} & \cellcolor{assessmentgreenlight}4.59{\fontsize{6}{7}\selectfont(18)} & \cellcolor{assessmentgreendark}5.06{\fontsize{6}{7}\selectfont(20)} & \cellcolor{assessmentgreenlight}2.94{\fontsize{6}{7}\selectfont(19)} & \cellcolor{assessmentgreendark}57.67{\fontsize{6}{7}\selectfont(22)} & \cellcolor{assessmentgreenlight}3.82{\fontsize{6}{7}\selectfont(25)} & \cellcolor{assessmentgreendark}5.21{\fontsize{6}{7}\selectfont(19)} & \cellcolor{assessmentgreenlight}4.55{\fontsize{6}{7}\selectfont(15)} & \cellcolor{assessmentgreendark}5.22{\fontsize{6}{7}\selectfont(19)} & \cellcolor{assessmentgreenlight}5.04{\fontsize{6}{7}\selectfont(22)} & \cellcolor{assessmentgreendark}5.28{\fontsize{6}{7}\selectfont(18)} & \cellcolor{assessmentgreenlight}50.39{\fontsize{6}{7}\selectfont(19)} & \cellcolor{assessmentgreendark}0.298{\fontsize{6}{7}\selectfont(7)} & \cellcolor{assessmentgreenlight}2.669{\fontsize{6}{7}\selectfont(16)} & \cellcolor{assessmentgreendark}0.371{\fontsize{6}{7}\selectfont(18)} & \cellcolor{assessmentgreenlight}0.500{\fontsize{6}{7}\selectfont(4)} \\
\cellcolor{assessmentgraylight}ColorCtrl~\cite{colorctrl} & \cellcolor{assessmentgraylight}34.62{\fontsize{6}{7}\selectfont(22)} & \cellcolor{assessmentgraydark}5.09{\fontsize{6}{7}\selectfont(15)} & \cellcolor{assessmentgraylight}4.52{\fontsize{6}{7}\selectfont(23)} & \cellcolor{assessmentgraydark}5.06{\fontsize{6}{7}\selectfont(18)} & \cellcolor{assessmentgraylight}2.99{\fontsize{6}{7}\selectfont(10)} & \cellcolor{assessmentgraydark}56.26{\fontsize{6}{7}\selectfont(25)} & \cellcolor{assessmentgraylight}3.83{\fontsize{6}{7}\selectfont(21)} & \cellcolor{assessmentgraydark}5.23{\fontsize{6}{7}\selectfont(16)} & \cellcolor{assessmentgraylight}4.58{\fontsize{6}{7}\selectfont(8)} & \cellcolor{assessmentgraydark}5.22{\fontsize{6}{7}\selectfont(16)} & \cellcolor{assessmentgraylight}5.04{\fontsize{6}{7}\selectfont(23)} & \cellcolor{assessmentgraydark}5.29{\fontsize{6}{7}\selectfont(13)} & \cellcolor{assessmentgraylight}50.51{\fontsize{6}{7}\selectfont(18)} & \cellcolor{assessmentgraydark}0.286{\fontsize{6}{7}\selectfont(22)} & \cellcolor{assessmentgraylight}2.452{\fontsize{6}{7}\selectfont(21)} & \cellcolor{assessmentgraydark}0.353{\fontsize{6}{7}\selectfont(24)} & \cellcolor{assessmentgraylight}0.475{\fontsize{6}{7}\selectfont(14)} \\
\cellcolor{assessmentgraylight}AceTone~\cite{acetone} & \cellcolor{assessmentgraylight}32.75{\fontsize{6}{7}\selectfont(23)} & \cellcolor{assessmentgraydark}5.06{\fontsize{6}{7}\selectfont(21)} & \cellcolor{assessmentgraylight}4.60{\fontsize{6}{7}\selectfont(16)} & \cellcolor{assessmentgraydark}5.06{\fontsize{6}{7}\selectfont(21)} & \cellcolor{assessmentgraylight}2.93{\fontsize{6}{7}\selectfont(20)} & \cellcolor{assessmentgraydark}58.22{\fontsize{6}{7}\selectfont(13)} & \cellcolor{assessmentgraylight}3.83{\fontsize{6}{7}\selectfont(20)} & \cellcolor{assessmentgraydark}5.22{\fontsize{6}{7}\selectfont(18)} & \cellcolor{assessmentgraylight}4.53{\fontsize{6}{7}\selectfont(21)} & \cellcolor{assessmentgraydark}5.21{\fontsize{6}{7}\selectfont(20)} & \cellcolor{assessmentgraylight}5.05{\fontsize{6}{7}\selectfont(20)} & \cellcolor{assessmentgraydark}5.27{\fontsize{6}{7}\selectfont(21)} & \cellcolor{assessmentgraylight}50.31{\fontsize{6}{7}\selectfont(20)} & \cellcolor{assessmentgraydark}0.289{\fontsize{6}{7}\selectfont(19)} & \cellcolor{assessmentgraylight}2.581{\fontsize{6}{7}\selectfont(20)} & \cellcolor{assessmentgraydark}0.364{\fontsize{6}{7}\selectfont(22)} & \cellcolor{assessmentgraylight}0.478{\fontsize{6}{7}\selectfont(13)} \\
\cellcolor{assessmentgraylight}Exposure~\cite{exposure} & \cellcolor{assessmentgraylight}30.88{\fontsize{6}{7}\selectfont(24)} & \cellcolor{assessmentgraydark}5.04{\fontsize{6}{7}\selectfont(22)} & \cellcolor{assessmentgraylight}4.60{\fontsize{6}{7}\selectfont(17)} & \cellcolor{assessmentgraydark}5.01{\fontsize{6}{7}\selectfont(22)} & \cellcolor{assessmentgraylight}2.85{\fontsize{6}{7}\selectfont(24)} & \cellcolor{assessmentgraydark}58.94{\fontsize{6}{7}\selectfont(7)} & \cellcolor{assessmentgraylight}3.86{\fontsize{6}{7}\selectfont(7)} & \cellcolor{assessmentgraydark}5.17{\fontsize{6}{7}\selectfont(23)} & \cellcolor{assessmentgraylight}4.49{\fontsize{6}{7}\selectfont(22)} & \cellcolor{assessmentgraydark}5.17{\fontsize{6}{7}\selectfont(23)} & \cellcolor{assessmentgraylight}5.07{\fontsize{6}{7}\selectfont(17)} & \cellcolor{assessmentgraydark}5.24{\fontsize{6}{7}\selectfont(23)} & \cellcolor{assessmentgraylight}49.74{\fontsize{6}{7}\selectfont(22)} & \cellcolor{assessmentgraydark}0.283{\fontsize{6}{7}\selectfont(24)} & \cellcolor{assessmentgraylight}2.306{\fontsize{6}{7}\selectfont(23)} & \cellcolor{assessmentgraydark}0.368{\fontsize{6}{7}\selectfont(20)} & \cellcolor{assessmentgraylight}0.446{\fontsize{6}{7}\selectfont(22)} \\
\cellcolor{assessmentgraylight}DeepUPE~\cite{deepupe} & \cellcolor{assessmentgraylight}22.25{\fontsize{6}{7}\selectfont(25)} & \cellcolor{assessmentgraydark}5.02{\fontsize{6}{7}\selectfont(23)} & \cellcolor{assessmentgraylight}4.54{\fontsize{6}{7}\selectfont(22)} & \cellcolor{assessmentgraydark}5.06{\fontsize{6}{7}\selectfont(19)} & \cellcolor{assessmentgraylight}2.93{\fontsize{6}{7}\selectfont(21)} & \cellcolor{assessmentgraydark}56.48{\fontsize{6}{7}\selectfont(24)} & \cellcolor{assessmentgraylight}3.82{\fontsize{6}{7}\selectfont(23)} & \cellcolor{assessmentgraydark}5.17{\fontsize{6}{7}\selectfont(22)} & \cellcolor{assessmentgraylight}4.46{\fontsize{6}{7}\selectfont(23)} & \cellcolor{assessmentgraydark}5.20{\fontsize{6}{7}\selectfont(21)} & \cellcolor{assessmentgraylight}5.05{\fontsize{6}{7}\selectfont(21)} & \cellcolor{assessmentgraydark}5.28{\fontsize{6}{7}\selectfont(20)} & \cellcolor{assessmentgraylight}50.16{\fontsize{6}{7}\selectfont(21)} & \cellcolor{assessmentgraydark}0.286{\fontsize{6}{7}\selectfont(21)} & \cellcolor{assessmentgraylight}2.586{\fontsize{6}{7}\selectfont(19)} & \cellcolor{assessmentgraydark}0.369{\fontsize{6}{7}\selectfont(19)} & \cellcolor{assessmentgraylight}0.467{\fontsize{6}{7}\selectfont(19)} \\
\cellcolor{assessmentgraylight}MIRNet~\cite{mirnet} & \cellcolor{assessmentgraylight}14.75{\fontsize{6}{7}\selectfont(26)} & \cellcolor{assessmentgraydark}4.98{\fontsize{6}{7}\selectfont(24)} & \cellcolor{assessmentgraylight}4.46{\fontsize{6}{7}\selectfont(25)} & \cellcolor{assessmentgraydark}4.98{\fontsize{6}{7}\selectfont(24)} & \cellcolor{assessmentgraylight}2.87{\fontsize{6}{7}\selectfont(22)} & \cellcolor{assessmentgraydark}57.77{\fontsize{6}{7}\selectfont(20)} & \cellcolor{assessmentgraylight}3.79{\fontsize{6}{7}\selectfont(26)} & \cellcolor{assessmentgraydark}5.10{\fontsize{6}{7}\selectfont(24)} & \cellcolor{assessmentgraylight}4.42{\fontsize{6}{7}\selectfont(25)} & \cellcolor{assessmentgraydark}5.14{\fontsize{6}{7}\selectfont(24)} & \cellcolor{assessmentgraylight}5.01{\fontsize{6}{7}\selectfont(24)} & \cellcolor{assessmentgraydark}5.23{\fontsize{6}{7}\selectfont(24)} & \cellcolor{assessmentgraylight}49.64{\fontsize{6}{7}\selectfont(23)} & \cellcolor{assessmentgraydark}0.287{\fontsize{6}{7}\selectfont(20)} & \cellcolor{assessmentgraylight}2.407{\fontsize{6}{7}\selectfont(22)} & \cellcolor{assessmentgraydark}0.367{\fontsize{6}{7}\selectfont(21)} & \cellcolor{assessmentgraylight}0.438{\fontsize{6}{7}\selectfont(25)} \\
\cellcolor{assessmentgraylight}CSRNet~\cite{csrnet} & \cellcolor{assessmentgraylight}12.88{\fontsize{6}{7}\selectfont(27)} & \cellcolor{assessmentgraydark}4.96{\fontsize{6}{7}\selectfont(25)} & \cellcolor{assessmentgraylight}4.48{\fontsize{6}{7}\selectfont(24)} & \cellcolor{assessmentgraydark}4.92{\fontsize{6}{7}\selectfont(25)} & \cellcolor{assessmentgraylight}2.80{\fontsize{6}{7}\selectfont(25)} & \cellcolor{assessmentgraydark}57.38{\fontsize{6}{7}\selectfont(23)} & \cellcolor{assessmentgraylight}3.82{\fontsize{6}{7}\selectfont(24)} & \cellcolor{assessmentgraydark}5.09{\fontsize{6}{7}\selectfont(25)} & \cellcolor{assessmentgraylight}4.42{\fontsize{6}{7}\selectfont(24)} & \cellcolor{assessmentgraydark}5.13{\fontsize{6}{7}\selectfont(25)} & \cellcolor{assessmentgraylight}5.00{\fontsize{6}{7}\selectfont(25)} & \cellcolor{assessmentgraydark}5.22{\fontsize{6}{7}\selectfont(25)} & \cellcolor{assessmentgraylight}48.73{\fontsize{6}{7}\selectfont(25)} & \cellcolor{assessmentgraydark}0.284{\fontsize{6}{7}\selectfont(23)} & \cellcolor{assessmentgraylight}2.216{\fontsize{6}{7}\selectfont(24)} & \cellcolor{assessmentgraydark}0.364{\fontsize{6}{7}\selectfont(23)} & \cellcolor{assessmentgraylight}0.437{\fontsize{6}{7}\selectfont(26)} \\
\cellcolor{assessmentgraylight}SPL~\cite{spl} & \cellcolor{assessmentgraylight}10.75{\fontsize{6}{7}\selectfont(28)} & \cellcolor{assessmentgraydark}4.87{\fontsize{6}{7}\selectfont(26)} & \cellcolor{assessmentgraylight}4.25{\fontsize{6}{7}\selectfont(26)} & \cellcolor{assessmentgraydark}4.83{\fontsize{6}{7}\selectfont(26)} & \cellcolor{assessmentgraylight}2.73{\fontsize{6}{7}\selectfont(26)} & \cellcolor{assessmentgraydark}52.90{\fontsize{6}{7}\selectfont(26)} & \cellcolor{assessmentgraylight}3.84{\fontsize{6}{7}\selectfont(15)} & \cellcolor{assessmentgraydark}5.05{\fontsize{6}{7}\selectfont(26)} & \cellcolor{assessmentgraylight}4.26{\fontsize{6}{7}\selectfont(26)} & \cellcolor{assessmentgraydark}5.12{\fontsize{6}{7}\selectfont(26)} & \cellcolor{assessmentgraylight}4.78{\fontsize{6}{7}\selectfont(26)} & \cellcolor{assessmentgraydark}5.22{\fontsize{6}{7}\selectfont(26)} & \cellcolor{assessmentgraylight}48.11{\fontsize{6}{7}\selectfont(26)} & \cellcolor{assessmentgraydark}0.272{\fontsize{6}{7}\selectfont(25)} & \cellcolor{assessmentgraylight}1.580{\fontsize{6}{7}\selectfont(26)} & \cellcolor{assessmentgraydark}0.328{\fontsize{6}{7}\selectfont(26)} & \cellcolor{assessmentgraylight}0.486{\fontsize{6}{7}\selectfont(6)} \\
\bottomrule
\end{tabular*}
\endgroup
\end{table*}

\subsection{Optimization and Inference}

Alongside teacher supervision, Grid Smoothness Regularization smooths raw coefficients in $\mathcal B$, and Curve Smoothness Regularization smooths curve slopes. Gamut Regularization penalizes RGB values outside $[0,1]$ before clipping. The complete objective is:
\begin{equation}
\begin{aligned}
\mathcal L={}&\mathcal L_{\mathrm{col}}+\lambda_{\mathrm{dist}}\mathcal L_{\mathrm{dist}}
+\lambda_{\mathrm{grid}}\mathcal L_{\mathrm{grid}}\\
&+\lambda_{\mathrm{curve}}\mathcal L_{\mathrm{curve}}
+\lambda_{\mathrm{gamut}}\mathcal L_{\mathrm{gamut}}.
\end{aligned}
\label{eq:objective}
\end{equation}
Losses are averaged over source-teacher pairs in each batch, with weights $\lambda$ balancing the distribution and regularization terms. Training establishes the overall color mapping with global transforms before gradually introducing local corrections for regional differences. Appendices~\ref{app:optimization} and~\ref{app:student-settings} detail the regularizers, schedule, and settings.

At inference, $t$ selects the editing style. The student predicts parameters once from $I$ and $t$ and renders at full resolution. Enhancement Strength $\alpha\in[0,1]$ blends $I$ with the unclipped output as $\operatorname{clip}_{[0,1]}(I+\alpha(v_t-I))$. Thus, $\alpha=0$ reproduces $I$, while $\alpha=1$ applies the full transform.

\section{Experiments and Results}
\label{sec:experiments}

\subsection{Experimental Setup}
\label{sec:experimental-setup}

\textbfnoindent{Implementation Details.}
\studentname jointly learns four teacher conditions from 11,995 training source images. We select one checkpoint using color and distribution losses on a validation set of 1,380 source images. The 1.13M-parameter model takes the source image and a teacher condition $t$ specifying one of the four teachers' editing styles as inputs and renders at the source image's resolution. Appendix~\ref{app:student-settings} gives the training and inference settings.

\textbfnoindent{Evaluation Protocol.}
We evaluate 500 FiveK and 2,286 PPR10K test images with 12 aesthetic and 4 image-quality assessments. Main comparisons use the same jointly trained student, conditioned on the FLUX.2-dev teacher for FiveK and the FireRed-Image-Edit teacher for PPR10K. Joint Score summarizes aesthetic and image-quality performance. We compare student outputs with their corresponding teacher targets under all four conditions using PSNR, SSIM~\cite{ssim}, LPIPS~\cite{lpips}, and CIE76 color difference. Appendices~\ref{app:assessors}, \ref{app:teacher-similarity}, and~\ref{app:ranking} detail these assessments.

\textbfnoindent{Comparison Baselines.}
We compare 20 methods spanning learned color transforms, image enhancement networks, and generative editing. Four teachers and all expert targets provide reference scores. Appendix~\ref{app:baselines} specifies released weights, input preparation, reference selection, etc.

\begin{figure*}[t]
\centering
\includegraphics[width=\textwidth]{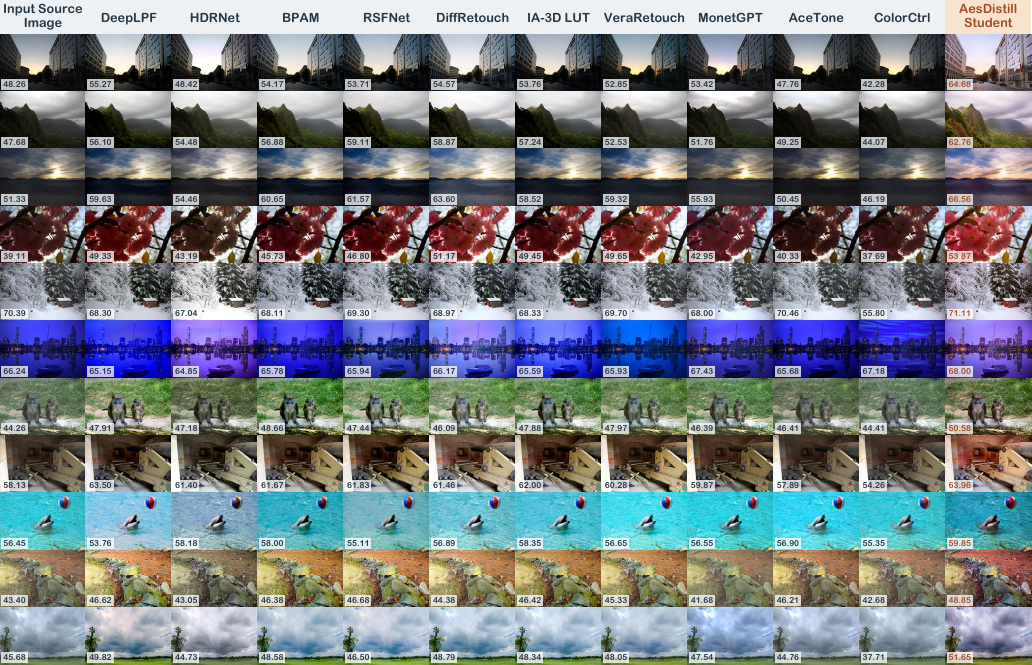}
\caption{Color enhancement comparisons with ten baselines on FiveK. \studentname achieves the highest ArtiMuse score on each shown example. Lower-left values report the scores (higher is better).}
\label{fig:baseline-qualitative}
\end{figure*}

\subsection{Aesthetic Enhancement Results}
\label{sec:enhancement-results}

\textbfnoindent{Teacher and Expert Targets.}
All four teachers rank above both expert target means in Table~\ref{tab:assessment-joint}. Within each dataset, the four-teacher mean exceeds the best expert on every aesthetic and quality assessment, supporting the value of generative targets across general photography and portraits. Appendix~\ref{app:scores} provides the complete scores.

\textbfnoindent{Per-Image Aesthetic Gains.}
Appendix~\ref{app:scores} shows per-image gains over the expert with the highest test-set mean for each dataset and assessment. Under NIMA, teachers exceed this expert on 67.1\% to 97.0\% of images. On FiveK, all four student conditions exceed the selected expert on most images under NIMA and TOPIQ-IAA. With the settings in Section~\ref{sec:experimental-setup}, the student improves NIMA over the input on 74.2\% of FiveK and 75.6\% of PPR10K images.

\textbfnoindent{Comparison with Existing Methods.}
\studentname achieves the highest Joint Score among all 21 enhancement methods, ranking directly behind the four teachers (Figure~\ref{fig:joint-rank-teaser}). Its Joint Score is 74.75, compared with 35.88 for the unedited input. Strong baselines immediately following our student include DeepLPF, HDRNet, RSFNet, and DiffRetouch. It also leads the baselines on each dataset. In the combined evaluation, it leads on TOPIQ-IAA, Q-Align, NIMA-VGG16, TANet-AVA, LAION-AES, ArtiMuse-AVA, and LIQE (Table~\ref{tab:assessment-joint}). These results show that distilling generative aesthetics enables a compact color model to lead the compared enhancement methods in Joint Score. Separate dataset results are provided in Appendix~\ref{app:dataset-comparisons}.

\begin{figure}[t]
\centering
\includegraphics[width=\columnwidth]{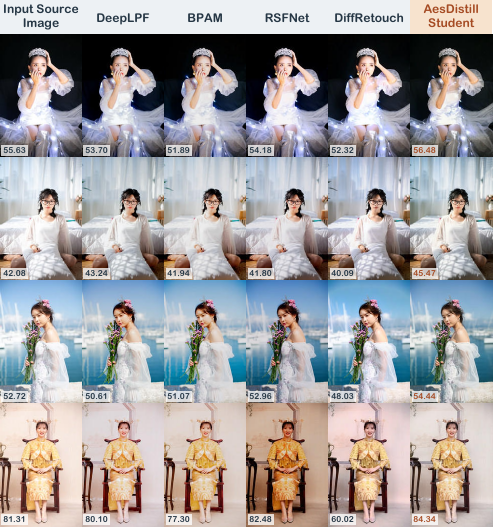}
\caption{\studentname achieves the highest ArtiMuse score among the compared methods on each PPR10K example shown. Lower-left values report the scores (higher is better).}
\label{fig:ppr-qualitative}
\end{figure}

\textbfnoindent{Qualitative Comparison.}
Figure~\ref{fig:expert-teacher-qualitative} compares expert targets with teacher and student edits on the same FiveK inputs. Figure~\ref{fig:baseline-qualitative} compares the student with ten baselines. In the street and mountain scenes, the student brightens dark regions while preserving clear differences between foreground and sky. It enhances the colors of pink blossoms, cyan water, and warm foliage while preserving their shapes. The night skyline shows a change in color balance as well as brightness. Higher ArtiMuse scores accompany these changes in the illustrated cases. Figure~\ref{fig:ppr-qualitative} shows qualitative examples on PPR10K portraits. Appendix~\ref{app:qualitative} provides qualitative evaluations at a larger display size.

\textbfnoindent{Teacher-to-Student Transfer.}
The jointly trained student reproduces warmer FLUX.2-dev colors and less saturated Step1X-Edit colors (Figure~\ref{fig:teacher-student-styles}). Table~\ref{tab:student-teacher-similarity} quantifies agreement with each teacher target, averaging 23.93\,dB PSNR, 0.8118 SSIM, 0.1662 LPIPS, and 8.40 color difference across the datasets. Figure~\ref{fig:structure-preservation} complements this target-based comparison: the student preserves the harbor and portrait backgrounds in the source images where teacher generation changes content. Appendix~\ref{app:confidence-visualization} shows six further cases across all four teachers where the student preserves source content despite added or moved people, replaced backgrounds, and removed details in teacher edits. Additional teacher-conditioned outputs appear in Appendix~\ref{app:multiteacher-styles}.

\begin{table}[t]
\centering
\caption{Similarity to teacher targets measures how well \studentname transfers teacher color edits on FiveK and PPR10K.}
\label{tab:student-teacher-similarity}
\begingroup
\definecolor{simblue}{HTML}{F0F7FC}
\definecolor{simpurple}{HTML}{F7F3FB}
\fontsize{6.8}{8.2}\selectfont
\setlength{\tabcolsep}{0.7pt}
\renewcommand{\arraystretch}{1.08}
\definecolor{simbluealt}{HTML}{E5F0F9}
\definecolor{simpurplealt}{HTML}{EEE7F6}
\definecolor{simconfig}{HTML}{FAFAFA}
\definecolor{simconfigalt}{HTML}{F0F2F4}
\begin{tabular*}{\columnwidth}{@{\extracolsep{\fill}}l|cccc|cccc@{}}
\toprule
Teacher target & \multicolumn{4}{c|}{\cellcolor{simblue}FiveK} & \multicolumn{4}{c}{\cellcolor{simpurple}PPR10K} \\
\cmidrule(lr){2-5}\cmidrule(lr){6-9}
& PSNR$^{\uparrow}$ & SSIM$^{\uparrow}$ & LPIPS$^{\downarrow}$ & $\Delta E_{ab}^{\downarrow}$ & PSNR$^{\uparrow}$ & SSIM$^{\uparrow}$ & LPIPS$^{\downarrow}$ & $\Delta E_{ab}^{\downarrow}$ \\
\midrule
\cellcolor{simconfig}\mbox{FLUX.2-dev} & \cellcolor{simblue}21.16 & \cellcolor{simblue}0.730 & \cellcolor{simblue}0.250 & \cellcolor{simblue}12.09 & \cellcolor{simpurple}22.66 & \cellcolor{simpurple}0.800 & \cellcolor{simpurple}0.210 & \cellcolor{simpurple}9.87 \\
\cellcolor{simconfigalt}\mbox{Step1X-Edit} & \cellcolor{simbluealt}26.16 & \cellcolor{simbluealt}0.839 & \cellcolor{simbluealt}0.105 & \cellcolor{simbluealt}7.52 & \cellcolor{simpurplealt}27.92 & \cellcolor{simpurplealt}0.913 & \cellcolor{simpurplealt}0.064 & \cellcolor{simpurplealt}5.91 \\
\cellcolor{simconfig}\mbox{HunyuanImage} & \cellcolor{simblue}23.12 & \cellcolor{simblue}0.794 & \cellcolor{simblue}0.188 & \cellcolor{simblue}8.81 & \cellcolor{simpurple}26.80 & \cellcolor{simpurple}0.887 & \cellcolor{simpurple}0.132 & \cellcolor{simpurple}5.78 \\
\cellcolor{simconfigalt}\mbox{FireRed-Image-Edit} & \cellcolor{simbluealt}21.02 & \cellcolor{simbluealt}0.713 & \cellcolor{simbluealt}0.220 & \cellcolor{simbluealt}9.54 & \cellcolor{simpurplealt}22.62 & \cellcolor{simpurplealt}0.818 & \cellcolor{simpurplealt}0.160 & \cellcolor{simpurplealt}7.63 \\
\bottomrule
\end{tabular*}
\endgroup
\end{table}

\begin{table}[t]
\centering
\caption{Joint and single-teacher training. Pooled averages raw scores across conditions before computing JS. Bold marks the highest score across teacher conditions in each column.}
\label{tab:single-joint-teacher}
\begingroup
\fontsize{6.8}{8.2}\selectfont
\setlength{\tabcolsep}{0.4pt}
\renewcommand{\arraystretch}{1.03}
\definecolor{ablblue}{HTML}{F0F7FC}
\definecolor{ablpurple}{HTML}{F7F3FB}
\definecolor{ablorange}{HTML}{FDF0E4}
\definecolor{ablbluealt}{HTML}{E5F0F9}
\definecolor{ablpurplealt}{HTML}{EEE7F6}
\definecolor{ablorangealt}{HTML}{F9E6D4}
\definecolor{ablconfig}{HTML}{FAFAFA}
\definecolor{ablconfigalt}{HTML}{F0F2F4}
\begin{tabular*}{\linewidth}{@{\extracolsep{\fill}}l|ccc|ccc|ccc@{}}
\toprule
& \multicolumn{3}{c|}{\cellcolor{ablblue}Single (exposure)} & \multicolumn{3}{c|}{\cellcolor{ablpurple}Single (steps)} & \multicolumn{3}{c}{\cellcolor{ablorange}Joint (ours)} \\
\cmidrule(lr){2-4}\cmidrule(lr){5-7}\cmidrule(lr){8-10}
Condition & FiveK & PPR10K & Avg. & FiveK & PPR10K & Avg. & FiveK & PPR10K & Avg. \\
\midrule
\cellcolor{ablconfig}\mbox{FLUX.2-dev} & \cellcolor{ablblue}\textbf{76.50} & \cellcolor{ablblue}60.25 & \cellcolor{ablblue}68.38 & \cellcolor{ablpurple}72.25 & \cellcolor{ablpurple}49.75 & \cellcolor{ablpurple}61.00 & \cellcolor{ablorange}\textbf{74.25} & \cellcolor{ablorange}51.50 & \cellcolor{ablorange}62.88 \\
\cellcolor{ablconfigalt}\mbox{Step1X-Edit} & \cellcolor{ablbluealt}65.25 & \cellcolor{ablbluealt}72.25 & \cellcolor{ablbluealt}68.75 & \cellcolor{ablpurplealt}64.50 & \cellcolor{ablpurplealt}68.00 & \cellcolor{ablpurplealt}66.25 & \cellcolor{ablorangealt}69.25 & \cellcolor{ablorangealt}71.75 & \cellcolor{ablorangealt}70.50 \\
\cellcolor{ablconfig}\mbox{HunyuanImage} & \cellcolor{ablblue}65.75 & \cellcolor{ablblue}70.50 & \cellcolor{ablblue}68.12 & \cellcolor{ablpurple}\textbf{72.50} & \cellcolor{ablpurple}69.75 & \cellcolor{ablpurple}71.12 & \cellcolor{ablorange}69.00 & \cellcolor{ablorange}71.50 & \cellcolor{ablorange}70.25 \\
\cellcolor{ablconfigalt}\mbox{FireRed-Image-Edit} & \cellcolor{ablbluealt}67.75 & \cellcolor{ablbluealt}\textbf{75.50} & \cellcolor{ablbluealt}\textbf{71.62} & \cellcolor{ablpurplealt}69.00 & \cellcolor{ablpurplealt}\textbf{75.50} & \cellcolor{ablpurplealt}\textbf{72.25} & \cellcolor{ablorangealt}69.00 & \cellcolor{ablorangealt}\textbf{75.25} & \cellcolor{ablorangealt}\textbf{72.12} \\
\midrule
\cellcolor{ablconfig}Pooled & \cellcolor{ablblue}70.50 & \cellcolor{ablblue}71.50 & \cellcolor{ablblue}71.00 & \cellcolor{ablpurple}71.75 & \cellcolor{ablpurple}66.50 & \cellcolor{ablpurple}69.12 & \cellcolor{ablorange}72.25 & \cellcolor{ablorange}69.75 & \cellcolor{ablorange}71.00 \\
\bottomrule
\end{tabular*}
\endgroup

\end{table}

\begin{table}[t]
\centering
\caption{Color-transform and supervision variants. Validation similarity averages all four conditions.}
\label{tab:student-components}
\begingroup
\scriptsize
\setlength{\tabcolsep}{0.6pt}
\renewcommand{\arraystretch}{1.03}
\definecolor{ablblue}{HTML}{F0F7FC}
\definecolor{ablpurple}{HTML}{F7F3FB}
\definecolor{ablorange}{HTML}{FDF0E4}
\definecolor{abldev}{HTML}{F0F6F1}
\definecolor{ablbluealt}{HTML}{E5F0F9}
\definecolor{ablpurplealt}{HTML}{EEE7F6}
\definecolor{ablorangealt}{HTML}{F9E6D4}
\definecolor{abldevalt}{HTML}{E6F0E8}
\definecolor{ablconfig}{HTML}{FAFAFA}
\definecolor{ablconfigalt}{HTML}{F0F2F4}
\begin{tabular*}{\linewidth}{@{\extracolsep{\fill}}>{\raggedright\arraybackslash}m{58pt}*{8}{c}@{}}
\toprule
& \multicolumn{2}{c}{\cellcolor{ablblue}\shortstack{Test Joint\\Score $\uparrow$}} & \multicolumn{3}{c}{\cellcolor{ablorange}\shortstack{Both test\\datasets $\uparrow$}} & \multicolumn{3}{c}{\cellcolor{abldev}\shortstack{Similarity to\\teacher targets (val)}} \\
\cmidrule(lr){2-3}\cmidrule(lr){4-6}\cmidrule(lr){7-9}
Variant & FiveK & PPR10K & Aes. & Qual. & Joint & PSNR $\uparrow$ & SSIM $\uparrow$ & MAE $\downarrow$ \\
\midrule
\cellcolor{ablconfig}w/o Local Affine Transform & \cellcolor{ablblue}52.50 & \cellcolor{ablpurple}63.75 & \cellcolor{ablorange}57.83 & \cellcolor{ablorange}59.00 & \cellcolor{ablorange}58.12 & \cellcolor{abldev}24.497 & \cellcolor{abldev}0.8677 & \cellcolor{abldev}0.05226 \\
\cellcolor{ablconfigalt}w/o Multi-Scale Color Supervision & \cellcolor{ablbluealt}63.25 & \cellcolor{ablpurplealt}66.00 & \cellcolor{ablorangealt}67.83 & \cellcolor{ablorangealt}55.00 & \cellcolor{ablorangealt}64.62 & \cellcolor{abldevalt}25.005 & \cellcolor{abldevalt}0.8732 & \cellcolor{abldevalt}0.04922 \\
\cellcolor{ablconfig}w/o Regional Color\newline Distribution Loss & \cellcolor{ablblue}66.25 & \cellcolor{ablpurple}67.50 & \cellcolor{ablorange}69.00 & \cellcolor{ablorange}60.50 & \cellcolor{ablorange}66.88 & \cellcolor{abldev}25.037 & \cellcolor{abldev}0.8744 & \cellcolor{abldev}0.04899 \\
\cellcolor{ablconfigalt}w/o Regional Partitioning & \cellcolor{ablbluealt}66.50 & \cellcolor{ablpurplealt}68.25 & \cellcolor{ablorangealt}69.83 & \cellcolor{ablorangealt}60.00 & \cellcolor{ablorangealt}67.38 & \cellcolor{abldevalt}25.026 & \cellcolor{abldevalt}0.8748 & \cellcolor{abldevalt}0.04906 \\
\cellcolor{ablconfig}w/o Confidence Weighting & \cellcolor{ablblue}62.25 & \cellcolor{ablpurple}65.50 & \cellcolor{ablorange}66.17 & \cellcolor{ablorange}57.00 & \cellcolor{ablorange}63.88 & \cellcolor{abldev}25.042 & \cellcolor{abldev}0.8738 & \cellcolor{abldev}0.04903 \\
\cellcolor{ablconfigalt}w/o Local Feature Matching & \cellcolor{ablbluealt}71.50 & \cellcolor{ablpurplealt}71.25 & \cellcolor{ablorangealt}73.50 & \cellcolor{ablorangealt}\textbf{65.00} & \cellcolor{ablorangealt}71.38 & \cellcolor{abldevalt}25.066 & \cellcolor{abldevalt}0.8766 & \cellcolor{abldevalt}0.04879 \\
\midrule
\cellcolor{ablconfig}AesDistill Student & \cellcolor{ablblue}\textbf{74.25} & \cellcolor{ablpurple}\textbf{75.25} & \cellcolor{ablorange}\textbf{78.17} & \cellcolor{ablorange}64.50 & \cellcolor{ablorange}\textbf{74.75} & \cellcolor{abldev}\textbf{25.078} & \cellcolor{abldev}\textbf{0.8772} & \cellcolor{abldev}\textbf{0.04871} \\
\bottomrule
\end{tabular*}
\endgroup

\end{table}

\subsection{Ablation Studies}
\label{sec:ablation}

\textbfnoindent{Teacher Supervision.}
We compare one student jointly trained on four teachers with separate single-teacher students in Table~\ref{tab:single-joint-teacher}. Single-teacher students use either the same sample presentations per teacher (Single (exposure)) or the same total optimizer updates (Single (steps)) as joint training. Joint training matches the pooled JS of four Single (exposure) students with one quarter of their total parameters (1.13M vs.\ 4.51M), supporting all four editing choices. Appendix~\ref{app:teacher-supervision-results} analyzes the condition-specific results.

\textbfnoindent{Student Components.}
The complete student achieves the highest Joint Score on both datasets and the best validation PSNR, SSIM, and MAE against teacher targets in Table~\ref{tab:student-components}. The largest gaps occur for variants without the Local Affine Transform or Confidence Weighting. This pattern supports combining spatially varying color transforms with supervision that emphasizes reliable teacher matches. Together, these mechanisms adapt teacher colors to the source image while limiting the influence of mismatched regions. Appendices~\ref{app:ablation-settings} and~\ref{app:component-results} provide details.

\subsection{Human Preference Study}
\label{sec:human-study}
We invited 12 participants and collected 768 binary color-preference judgments on 32 FiveK and 32 PPR10K test images. For each pair of outputs from the same source image, three participants chose the more appealing colors with method names hidden. Teachers were preferred over expert edits in 86.98\% of comparisons. \studentname was preferred over expert edits, HDRNet, and DeepLPF in 70.31\%, 72.40\%, and 72.40\% of comparisons, respectively. Every teacher and student comparison received a majority of preferences on both datasets. These results support the aesthetic value of generative targets and its transfer to a compact student. Appendices~\ref{app:human-study} and~\ref{app:human-study-results} provide the protocol and complete results, respectively.

\section{Conclusion}
\label{sec:conclusion}
\method turns generative aesthetic knowledge into reusable targets and compact image color enhancement. The \datasetname provides 64,644 edits from four large teachers whose targets surpass expert target means in the joint assessment on FiveK and PPR10K. \studentname learns their distinct color choices with 1.13M parameters while preserving source image structure. Its leading joint ranking among 21 enhancement methods shows that generative aesthetics can transfer to a small color model, extending enhancement beyond fixed expert preferences.

{\small
\bibliographystyle{ieeenat_fullname}
\bibliography{references}
}
\appendix
\maketitlesupplementary
\setcounter{page}{1}
\setcounter{dbltopnumber}{2}
\setcounter{topnumber}{1}
\setcounter{bottomnumber}{1}
\setcounter{totalnumber}{2}
\renewcommand{\textfraction}{0.16}
\renewcommand{\topfraction}{0.84}
\renewcommand{\dbltopfraction}{0.84}

\section{Method Details}
\label{app:implementation}

This section specifies the correspondence, color renderer, and supervision used by \studentname. All four teacher conditions share the same student parameters and use separate learned embeddings. Appendix~\ref{app:student-settings} provides the preprocessing, hyperparameters, and training settings.

\subsection{\correspondencemodule}
\label{app:correspondence}

\textbfnoindent{Feature Extraction.}
Source and teacher images undergo EXIF orientation correction, RGB conversion, and the resizing and normalization in Appendix~\ref{app:student-settings}. Frozen DINOv2-Small features are extracted from the last hidden state after removing the class token. Each patch feature is normalized to unit length, forming an $h\times h$ grid. Correspondence coordinates and confidence are precomputed with the frozen encoder and remain fixed during student training.

\textbfnoindent{Local Matching and Confidence.}
The search neighborhood in Eq.~\ref{eq:matching} is a disk centered at the source feature position. Let $a_p$ be the highest cosine similarity within this disk, $m_p$ the difference between the two highest similarities at distinct grid positions, and $c_p=\|\phi_t^{\leftarrow}(\phi_t(p))-p\|_2$ the cycle error in feature-grid units, where $\phi_t^{\leftarrow}$ is the reverse match. With thresholds $\tau_a\in(0,1)$, $\tau_m>0$, and $\tau_c>0$, and $[x]_0^1=\min(1,\max(0,x))$, confidence is:
\begin{equation}
\begin{aligned}
h_p&=\mathbf 1[a_p\geq\tau_a,\ m_p\geq\tau_m,\ c_p\leq\tau_c],\\
w_t(p)&=h_p\left[\frac{a_p-\tau_a}{1-\tau_a}\right]_0^1
\left[\frac{m_p}{4\tau_m}\right]_0^1 e^{-c_p/\tau_c}.
\end{aligned}
\end{equation}
For a teacher position $q$, reverse matching selects the source position with the highest cosine similarity within a disk of the same radius centered at $q$.

\textbfnoindent{Scale Conversion.}
For each color scale, the clipped student output and teacher image are resized with antialiased bilinear interpolation. Let $g_t$ denote the cached teacher coordinates in $[-1,1]^2$ on the $h\times h$ feature grid, and $g^{\mathrm{id}}_h$ the identity grid of pixel centers at that resolution. The sampling coordinates at scale $s$ are:
\begin{equation*}
g_{t,s}=g^{\mathrm{id}}_s+\operatorname{Resize}_{s}(g_t-g^{\mathrm{id}}_h).
\end{equation*}
The displacement is resized bilinearly. Teacher colors are then sampled bilinearly with border padding and pixel-center coordinates. Resizing displacement preserves an identity correspondence at every scale. Confidence is resized by nearest neighbors. Weights below a threshold $\tau_w$ and weights at coordinates outside $[-1,1]^2$ are set to zero. Rejected regions receive no direct teacher-color supervision. Appendix~\ref{app:confidence-visualization} visualizes the matching fields and confidence on actual image pairs.

\subsection{\transformmodule}
\label{app:color-transforms}

\textbfnoindent{Color Parameter Prediction.}
The source image preview preserves aspect ratio subject to a minimum short side. Its size and RGB normalization are specified in Appendix~\ref{app:student-settings}. RGB statistics are computed before normalization. They comprise three means, three population standard deviations, and three normalized 16-bin histograms, giving 54 values. Each histogram assigns intensity $x\in[0,1]$ to bin $\min(\lfloor16x\rfloor,15)$ and divides bin counts by the number of preview pixels. The context network concatenates these statistics with the 576-dimensional pooled encoder feature and a 32-dimensional Teacher Embedding. Two fully connected layers with dimensions $662\to128\to128$, each followed by SiLU, produce the context $\mathbf h$.

\textbfnoindent{Local Parameter Decoder.}
Features are extracted after layers 1, 3, 8, and 12 of the TorchVision MobileNetV3-Small \texttt{features} sequence, using zero-based indices. These four levels have 16, 24, 48, and 576 channels and feed a 64-channel top-down decoder. Each level uses a $1\times1$ lateral projection. Starting from the coarsest level, the decoder feature is bilinearly resized and added to the lateral feature at the next finer level. Each level then applies a $3\times3$ depthwise convolution, eight-group normalization, SiLU, a $1\times1$ convolution, and SiLU. Let $F$ be the resulting finest-scale feature. A linear projection of $\mathbf h$ produces channel-wise vectors $s(\mathbf h)$ and $b(\mathbf h)$, giving:
\begin{equation*}
F'=F\odot\bigl(1+0.5\tanh(s(\mathbf h))\bigr)+b(\mathbf h).
\end{equation*}
Adaptive average pooling reduces $F'$ to $16\times16$. A $1\times1$ head predicts 96 channels, reshaped to a grid of $12\times8\times16\times16$ raw coefficients, ordered by coefficient, luma, height, and width. A separate linear head on $\mathbf h$ predicts $3K$ curve logits and 12 global affine coefficients.

\textbfnoindent{Curve Evaluation.}
The $K$ increments in Eq.~\ref{eq:curves} define $K+1$ knots per channel, with $C_{c,0}=0$ and $C_{c,k}=\sum_{j=0}^{k-1}\Delta_{c,j}$. For an input value $x\in[0,1]$, set $k=\min(\lfloor Kx\rfloor,K-1)$ and $r=Kx-k$. The channel output is $(1-r)C_{c,k}+rC_{c,k+1}$. Since each segment spans $1/K$, its slope is at least $\kappa$. Channel mixing and local affine transforms then compose these monotonic curves into the final color mapping.

\textbfnoindent{Affine Parameterization.}
For raw affine coefficients $z_{ij}$, the effective $3\times4$ matrix is:
\begin{equation}
A_{ij}=\begin{cases}
2^{a\tanh(z_{ij})}, & j=i,\\
b\tanh(z_{ij}), & j\leq3,\ j\ne i,\\
c\tanh(z_{i4}), & j=4.
\end{cases}
\end{equation}
where $a,b,c>0$ bound diagonal gains by $[2^{-a},2^a]$, cross-channel coefficients by $[-b,b]$, and offsets by $[-c,c]$. 

\textbfnoindent{Grid Slicing.}
The local transform receives the unclipped global RGB values $u_t$. The local grid is first sliced at each pixel to obtain raw affine coefficients, which are then converted to bounded gains and offsets by the equation above. The guide is source luma $\ell_I=0.2126I_R+0.7152I_G+0.0722I_B$, computed directly from RGB without linearization. For zero-based pixel coordinates $(x,y)$ in an $H\times W$ image, the normalized lookup is $(2(x+0.5)/W-1,2(y+0.5)/H-1,2\ell_I-1)$. Grid lookup uses trilinear interpolation with border padding. Zero-initialized global and grid prediction heads give identity curves and affine transforms.

\subsection{\supervisionmodule}
\label{app:color-supervision}

\textbfnoindent{Color Views and Regional Sampling.}
A color view is a resized square RGB image used to compute supervision. Student and aligned teacher views at the distribution-loss resolution are split into the same fixed rectangular regions. Within each region, sample indices are obtained by rounding equally spaced points from zero to the last row-major pixel index, including both endpoints. Student and aligned teacher colors use the same positions and confidence weights. These spatial partitions preserve coarse color placement without semantic masks. Appendix~\ref{app:student-settings} gives view sizes, region counts, and sampling budgets.

\textbfnoindent{Projection Directions.}
Colors are projected onto fixed unit directions. Six are normalized versions of $(1,0,0)$, $(0,1,0)$, $(0,0,1)$, $(1,1,1)$, $(1,-1,0)$, and $(1,1,-2)$. These cover RGB channels, brightness, and opponent colors. The remaining $M$ directions use a deterministic spherical construction. For $j=0,\ldots,M-1$, let $a_j=j+0.5$, $z_j=1-2a_j/M$, and $\theta_j=\pi(3-\sqrt5)a_j$. The direction is $(\sqrt{1-z_j^2}\cos\theta_j,\sqrt{1-z_j^2}\sin\theta_j,z_j)$. These projections are fixed throughout training.

\textbfnoindent{Weighted Quantiles.}
For each direction, projected student and teacher values are sorted independently, with their confidence weights reordered accordingly. In a region with positive total confidence, sorted values with weights $w_i$ have cumulative weight centers:
\begin{equation}
u_i=\frac{\sum_{j\leq i}w_j-w_i/2}{\sum_jw_j}.
\end{equation}
Linear interpolation between these centers gives $N_q$ weighted quantiles at $\tau_q=(q+0.5)/N_q$, for $q=0,\ldots,N_q-1$. Queries outside the supported centers use the nearest endpoint. Zero-weight samples are excluded from interpolation endpoints. Only regions with positive total confidence contribute to the distribution loss. Each supervision term is averaged across images after per-image coverage normalization. The distribution term uses the aligned teacher view and confidence while relaxing individual color pairing within each region.

\subsection{Optimization and Inference}
\label{app:optimization}

\textbfnoindent{Regularization.}
Let $\langle\cdot\rangle$ denote the mean over all entries, $\delta$ a finite difference, and $[x]_+=\max(x,0)$. Let $v_{\mathrm{raw}}$ denote the rendered RGB values before clipping, including the local-strength schedule during training. The regularizers on raw grid coefficients, curve knots, and output colors are:
\begin{equation}
\begin{aligned}
\mathcal L_{\mathrm{grid}}&=\langle(\delta_\ell\mathcal B)^2\rangle
+\tfrac12\!\sum_{a\in\{x,y\}}\langle(\delta_a\mathcal B)^2\rangle,\\
\mathcal L_{\mathrm{curve}}&=\langle(\delta_k^2\mathcal C)^2\rangle,\\
\mathcal L_{\mathrm{gamut}}&=\langle[-v_{\mathrm{raw}}]_+^2+[v_{\mathrm{raw}}-1]_+^2\rangle.
\end{aligned}
\end{equation}

\textbfnoindent{Global-to-Local Schedule.}
Let $e_{\mathrm{warm}}$ and $e_{\mathrm{ramp}}$ denote the global warm-up and local ramp durations. At training epoch $e$, counted from one, local strength is $\beta_e=[(e-e_{\mathrm{warm}})/e_{\mathrm{ramp}}]_0^1$. The rendered RGB values before clipping are $u_t+\beta_e(v_t-u_t)$. Global color mapping is learned first, and local corrections are introduced gradually. Inference uses the complete local transform.

\textbfnoindent{Output Rendering.}
Color and distribution losses operate on the clipped student image, while Gamut Regularization acts on its unclipped RGB values. Enhancement Strength blends the source image with the unclipped rendered result before the final $[0,1]$ clipping, so $\alpha=0$ gives the original input.

\section{Dataset and Assessment Details}
\label{app:data-details}

\subsection{Dataset Splits}
\label{app:data-splits}

FiveK uses source images a4501 to a5000 for testing. The fixed source-ID lists in \texttt{fivek\_sources.jsonl} assign images a0001 to a4500 to 4,050 training and 450 validation images. PPR10K uses its official 2,286-image validation set for testing. Its 8,875 training images are split by capture group with seed 2026, reserving approximately 10\% of groups for validation. Table~\ref{tab:dataset} gives the split sizes. All four teacher edits of a source image share its split, and PPR10K capture groups are kept within one split.

\begin{table}[htbp]
\centering
\small
\setlength{\tabcolsep}{3pt}
\caption{Source-image splits and teacher-edit counts for AesDistill FiveK and AesDistill PPR10K. Each source has four teacher edits.}
\label{tab:dataset}
\begin{tabular}{@{}lrrrr@{}}
\toprule
Subset & Train & Val. & Test & Edits \\
\midrule
AesDistill FiveK & 4,050 & 450 & 500 & 20,000 \\
AesDistill PPR10K & 7,945 & 930 & 2,286 & 44,644 \\
Total & 11,995 & 1,380 & 2,786 & 64,644 \\
\bottomrule
\end{tabular}
\end{table}

\subsection{Teacher Generation Protocol}
\label{app:generation}

\begin{table*}[t]
\centering
\small
\caption{Teacher sizes and generation settings. Parameter counts follow Appendix~\ref{app:efficiency}. Model names link to the released weights. All teachers use bfloat16 (BF16) and seed 2026.}
\label{tab:generation}
\begin{tabular*}{\textwidth}{@{\extracolsep{\fill}}lrrl@{}}
\toprule
Teacher & Params (B) & Steps & Generation setting \\
\midrule
\href{https://huggingface.co/black-forest-labs/FLUX.2-dev/tree/26afe3a78bb242c0a8bb181dcc8937bb16e5c66c}{FLUX.2-dev}~\cite{flux2} & $\approx$56.32 & 28 & Guidance 4.0 \\
\href{https://huggingface.co/stepfun-ai/Step1X-Edit-v1p2/tree/ca85b97fd19f2235dc0d6fd3633d1319f169e149}{Step1X-Edit-v1p2}~\cite{reasonedit} & 20.82 & 50 & Classifier-free guidance 6.0, thinking and reflection \\
\href{https://huggingface.co/tencent/HunyuanImage-3.0-Instruct-Distil/tree/c8ffd07206f1b843697606968196e8f59f8ff38c}{HunyuanImage-3.0-Instruct-Distil}~\cite{hunyuan} & 83.04 & 8 & Distilled model \\
\href{https://huggingface.co/FireRedTeam/FireRed-Image-Edit-1.1}{FireRed-Image-Edit-1.1}~\cite{firered} & 28.85 & 40 & Classifier-free guidance 4.0 \\
\bottomrule
\end{tabular*}
\end{table*}

\textbfnoindent{Shared Inputs.}
All teachers use the same sRGB input. FiveK RAW images are rendered with rawpy/LibRaw using camera white balance, automatic brightness, and 8-bit sRGB output, with other settings at their defaults. Bicubic resizing gives the fixed 480p dimensions in the source manifest. PPR10K inputs use the official 360p images. After orientation correction, both datasets are converted to RGB and resized with Lanczos interpolation to a target area of $2^{20}$ pixels. For input dimensions $H,W$, the output dimensions are:
\begin{equation*}
\begin{aligned}
s&=\sqrt{2^{20}/(HW)},\\
H'&=16\max\!\left(1,\operatorname{round}(sH/16)\right),\\
W'&=16\max\!\left(1,\operatorname{round}(sW/16)\right).
\end{aligned}
\end{equation*}
Rounding uses the nearest integer, with ties to even. Images are stored as RGB PNG files.

\textbfnoindent{Generation and Prompt.}
Each teacher uses fixed weights, sampling settings, and seed 2026 to produce one output per source image. Table~\ref{tab:generation} lists the fixed model settings. The Step1X-Edit checkpoint uses the reasoning-enhanced model described in ReasonEdit, built on Step1X-Edit~\cite{reasonedit,step1x_base}. All teachers receive the following instruction:

\begin{quote}\small\itshape
Professionally retouch this photograph using only global color and tone adjustments: exposure, white balance, tonal curve, contrast, saturation, and color balance. Make the result natural, polished, and visually pleasing. Keep exactly the same composition, crop, camera viewpoint, objects, object boundaries, identities, text, texture, and fine detail. Do not add, remove, move, reshape, repaint, blur, sharpen, or locally relight anything. Preserve the aspect ratio.
\end{quote}

\textbfnoindent{Annotation Records.}
Each source--teacher pair includes the source identifier, teacher model, prompt, seed, image dimensions, generation time, and attributes describing changes in exposure, luminance contrast, saturation, and color warmth.

\textbfnoindent{Color-Edit Attributes.}
Source and teacher images are resized to aspect-preserving Lanczos thumbnails with a maximum side of 512 pixels. Pillow grayscale and HSV conversions provide grayscale intensity $Y$ and saturation $S$. Let $\mu_Y$, $\sigma_Y$, $\mu_S$, and $\mu_{R-B}$ denote the grayscale mean and population standard deviation, mean saturation, and mean red-minus-blue intensity, respectively, with pixel values scaled to $[0,1]$. For source image $I$ and teacher image $J$, the four attributes are:
\begin{equation*}
\begin{aligned}
\Delta e&=\log_2\frac{\mu_Y^J+1/255}{\mu_Y^I+1/255},\\
\Delta c&=\frac{\sigma_Y^J+10^{-8}}{\sigma_Y^I+10^{-8}}-1,\\
\Delta s&=\mu_S^J-\mu_S^I,\\
\Delta w&=\mu_{R-B}^J-\mu_{R-B}^I.
\end{aligned}
\end{equation*}
Intermediate statistics and final attributes are rounded to six decimal places.

\textbfnoindent{Annotation Cost.}
Generating all 64,644 teacher images consumed 3,364.8 H100 GPU-hours, or about 187 GPU-seconds per image. GPU-hours sum the time used across all H100 GPUs.

\subsection{Aesthetic and Image-Quality Assessment}
\label{app:assessors}

Table~\ref{tab:assessors} lists the 12 aesthetic and 4 technical image-quality assessment methods. Eleven models use PyIQA~0.1.16 with its default weights and preprocessing~\cite{pyiqa}. MLSP-AVA, TANet-AVA, ArtiMuse, ArtiMuse-AVA, and ArtQuant use their official implementations, as detailed below. Images enter each scorer at their stored resolution, followed by its own resizing and cropping. Each assessor uses the same settings for inputs, expert targets, teachers, baselines, and students.

\begin{table}[t]
\centering
\small
\caption{Assessment methods and score-training sources: 12 aesthetic (top) and 4 image-quality (bottom) models.}
\label{tab:assessors}
\begin{tabular}{@{}p{0.42\linewidth}p{0.52\linewidth}@{}}
\toprule
Assessment method & Score-training source \\
\midrule
NIMA~\cite{nima} & AVA \\
NIMA-VGG16~\cite{nima} & AVA \\
MUSIQ-AVA~\cite{musiq} & AVA \\
TOPIQ-IAA~\cite{topiq} & AVA \\
TOPIQ-Res50~\cite{topiq} & AVA \\
MLSP-AVA~\cite{mlsp} & AVA \\
TANet-AVA~\cite{tanet} & AVA \\
Q-Align~\cite{qalign} & AVA for aesthetics \\
LAION-AES~\cite{laion_aes} & AVA, SAC, LAION-Logos \\
ArtiMuse~\cite{artimuse} & ArtiMuse-10K \\
ArtiMuse-AVA~\cite{artimuse} & AVA score fine-tuning \\
ArtQuant~\cite{artquant} & APDD, description training \\
\midrule
MANIQA~\cite{maniqa} & KonIQ-10k \\
LIQE~\cite{liqe} & KonIQ-10k \\
TOPIQ-NR~\cite{topiq} & KonIQ-10k \\
CLIP-IQA~\cite{clipiqa} & No quality-score training \\
\bottomrule
\end{tabular}
\end{table}

\textbfnoindent{NIMA.}
NIMA~\cite{nima} uses Inception-ResNet-v2 weights trained on AVA~\cite{ava}. Inputs are resized to a shorter side of 299 pixels, center-cropped to $299\times299$, and normalized using the backbone's pretrained settings. The score is the expected rating over ten predicted rating bins, from 1 to 10.

\textbfnoindent{NIMA-VGG16.}
NIMA-VGG16~\cite{nima} uses the AVA-trained VGG16 checkpoint. Inputs are resized to a shorter side of 224 pixels and center-cropped to $224\times224$, with ImageNet normalization. Scoring uses the same rating expectation as NIMA.

\textbfnoindent{MUSIQ-AVA.}
MUSIQ-AVA~\cite{musiq} processes full-resolution patches together with two resized views whose longer sides are 224 and 384 pixels. Both resized views preserve the aspect ratio. The AVA-trained Transformer predicts the expected aesthetic rating.

\textbfnoindent{TOPIQ-IAA.}
TOPIQ-IAA~\cite{topiq} uses the AVA-trained Swin checkpoint with $384\times384$ inputs. The score is the expectation over ten rating bins.

\textbfnoindent{TOPIQ-Res50.}
TOPIQ-Res50~\cite{topiq} uses the AVA-trained ResNet50 checkpoint. Inputs are resized to a shorter side of 384 pixels while preserving the aspect ratio, and scores use the same rating expectation as TOPIQ-IAA.

\textbfnoindent{MLSP-AVA.}
MLSP-AVA~\cite{mlsp} uses the \href{https://github.com/subpic/ava-mlsp/tree/master/models/irnv2_mlsp_wide_orig}{official AVA-trained weights} with multi-level features and $5\times5$ spatial pooling. Inference uses TensorFlow~2.16 and Inception-ResNet-v2 preprocessing at the input image dimensions.

\textbfnoindent{TANet-AVA.}
TANet-AVA~\cite{tanet} uses the \href{https://github.com/woshidandan/TANet-image-aesthetics-and-quality-assessment/tree/main/code/AVA}{official AVA-trained weights} (\texttt{SRCC\_758\_LCC\_765}) and $224\times224$ RGB inputs. RGB values in $[0,1]$ are normalized using means $(0.485,0.456,0.406)$ and standard deviations $(0.229,0.224,0.225)$. Within each dataset and output variant, images are sorted lexicographically by source-image identifier and grouped into batches of 48. The final batch uses its actual size. A one-image batch duplicates that image for normalization and records one score. TANet's official forward pass creates a new batch-normalization layer, which uses statistics from the current batch. Scores are expectations over its ten rating bins, from 1 to 10.

\textbfnoindent{Q-Align.}
Q-Align~\cite{qalign} uses the released OneAlign weights in aesthetic mode (\texttt{task\_=aesthetic}) with FP16 inference. Each image is padded to a square using the CLIP mean RGB color, then passed through the checkpoint's image processor. The score is the probability-weighted mean of five rating levels, from \emph{bad} to \emph{excellent}, assigned values from 1 to 5.

\textbfnoindent{LAION-AES.}
LAION-AES~\cite{laion_aes} applies its released score predictor to normalized CLIP ViT-L/14 features. Inputs are resized with bicubic interpolation and center-cropped to $224\times224$, then normalized with CLIP statistics. Image embeddings are L2-normalized before scalar prediction. The predictor is trained on AVA, SAC, and LAION-Logos ratings.

\textbfnoindent{ArtiMuse.}
ArtiMuse~\cite{artimuse} uses the released \href{https://huggingface.co/Thunderbolt215215/ArtiMuse}{ArtiMuse weights}, trained with expert scores and attributes from ArtiMuse-10K. Inference uses BF16 and deterministic scoring. RGB inputs are resized to $448\times448$ with bicubic interpolation and use the same normalization as TANet-AVA. The score is the probability-weighted expectation over 101 score tokens representing values from 0 to 100.

\textbfnoindent{ArtiMuse-AVA.}
ArtiMuse-AVA~\cite{artimuse} uses the released \href{https://huggingface.co/Thunderbolt215215/ArtiMuse_AVA}{AVA-fine-tuned weights}. Preprocessing, numerical precision, and scoring follow the ArtiMuse settings above.

\textbfnoindent{ArtQuant.}
ArtQuant~\cite{artquant} uses the released \href{https://huggingface.co/Liuuuu123/ArtQuant}{APDD-trained adapter} with the \href{https://huggingface.co/MAGAer13/mplug-owl2-llama2-7b}{mPLUG-Owl2-7B} base model. Each image is centered on a square canvas padded with the image processor's mean RGB color, then passed through its preprocessing. The fixed prompt is ``How would you rate the quality of this painting?'' followed by the image token in the mPLUG-Owl2 conversation template, with the answer prefix ``The quality of the painting is''. The next-token logits for \emph{excellent}, \emph{good}, \emph{fair}, \emph{poor}, and \emph{bad} are normalized with softmax. The score is their probability-weighted average of ratings 5, 4, 3, 2, and 1.

\textbfnoindent{MANIQA.}
MANIQA~\cite{maniqa} uses the KonIQ-10k checkpoint with a ViT-B/8 backbone. RGB inputs are normalized to $[-1,1]$. Inference uses 20 uniformly placed $224\times224$ crops. The model combines patch scores across these crops using its learned weights to produce one image-quality score.

\textbfnoindent{LIQE.}
LIQE~\cite{liqe} uses the KonIQ-10k checkpoint with a CLIP ViT-B/32 backbone and quality-only text prompts. After CLIP normalization, $224\times224$ patches are extracted with stride 32. At most 15 patches are selected at fixed intervals in the patch sequence. Their image-text logits are averaged before softmax, and the score is the expectation over five quality levels from 1 to 5.

\textbfnoindent{TOPIQ-NR.}
TOPIQ-NR~\cite{topiq} uses the ResNet50 checkpoint trained on KonIQ-10k for no-reference image-quality prediction. It processes the stored image dimensions with ImageNet normalization and directly predicts a scalar quality score.

\textbfnoindent{CLIP-IQA.}
CLIP-IQA~\cite{clipiqa} uses pretrained CLIP RN50 for zero-shot quality assessment, without training on quality-score labels. Inputs retain their stored dimensions and use CLIP normalization, with positional embeddings disabled. The PyIQA implementation uses five prompt pairs covering overall quality, sharpness, edge clarity, resolution, and noise. It averages the positive-prompt probabilities from the five pairwise softmax outputs.

\textbfnoindent{Teacher and Expert Comparison.}
We compare teacher and expert edits on 500 FiveK and 2,286 PPR10K test source images, including all five FiveK and three PPR10K experts. Scores are averaged over the shared source set in each assessor's original units. For each assessment, the best expert has the highest dataset-level mean over all test source images. The teacher mean gives equal weight to the four teacher means. Comparisons use unrounded values. FiveK expert targets are converted from the official 16-bit ProPhoto RGB TIFF images to sRGB using their ICC matrices and tone response curves. PPR10K expert targets use the official 360p TIFF images. Both follow the same orientation and resizing as the corresponding source images before scoring.

\begin{figure*}[!t]
\centering
\includegraphics[width=\textwidth]{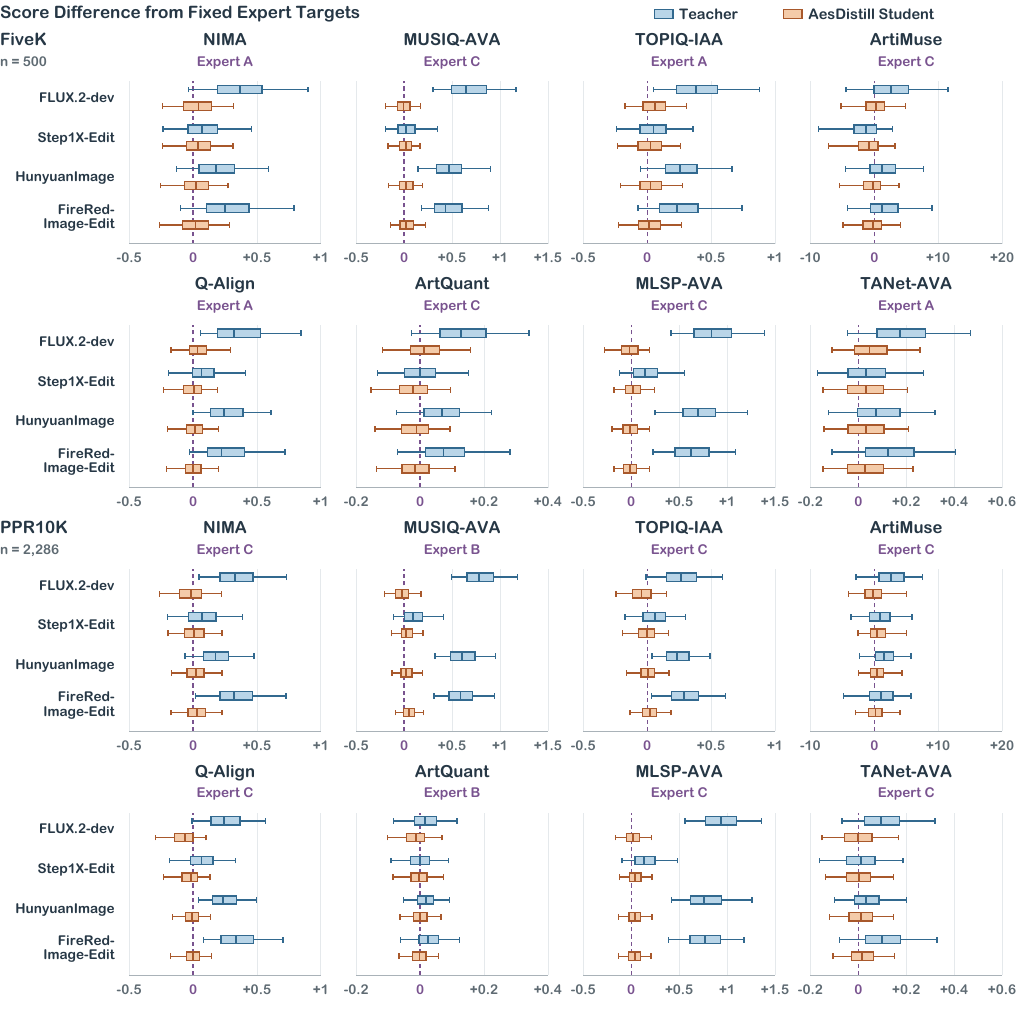}
\caption{Paired aesthetic score differences for eight assessors on all 500 FiveK and 2,286 PPR10K test images. Each panel uses the expert with the highest dataset mean for that assessor. Blue denotes teachers and orange the matching \studentname conditions. Boxes show the middle 50\%, lines mark medians, and whiskers span the 5th to 95th percentiles. Positive values favor the model.}
\label{fig:aesthetic-gain-distributions}
\end{figure*}

\textbfnoindent{Paired Aesthetic Gains.}
Outputs and the selected expert edits are paired by source image ID. Each teacher model and each student condition defines one output variant $v$. Let $s_i^v$ be the assessment score of output variant $v$ for source image $i$, and $s_i^{e^\star}$ the score of the selected expert edit. The per-image gain and positive-gain percentage are:
\begin{equation*}
\Delta_i^v=s_i^v-s_i^{e^\star},\qquad
P_+^v=\frac{100}{N}\sum_{i=1}^{N}\mathbf 1[\Delta_i^v>0].
\end{equation*}
Here $N$ is 500 for FiveK and 2,286 for PPR10K. The same expert is used for all source images within each dataset and assessment. Each variant is evaluated over the full test set. Ties have zero gain. Figure~\ref{fig:aesthetic-gain-distributions} shows the per-image gain distributions.

\subsection{Similarity to Teacher Targets}
\label{app:teacher-similarity}
We assess similarity to teacher targets by comparing each student output $\hat I_t$ directly with the corresponding teacher edit $J_t$. Test evaluation uses the approximately 1MP images. PSNR quantifies RGB reconstruction accuracy, SSIM~\cite{ssim} assesses local structural agreement, and LPIPS~\cite{lpips} quantifies learned perceptual distance. Color difference is CIE76, the mean Euclidean distance between corresponding CIELAB pixels. Evaluation uses the images at their original sampling locations, without applying the training correspondence map. Each metric is computed per image, then averaged over source images and teacher conditions. PSNR is averaged in decibels. The combined result weights FiveK and PPR10K equally.

\subsection{Joint Aesthetic and Image-Quality Ranking}
\label{app:ranking}
Assessment scales differ, so Joint Score combines percentile ranks. For dataset $d$, assessment $m$, and method $f$, let $s_{d,m,v}$ be configuration $v$'s mean score over test source images. We first compute the method score $\bar s_{d,m,f}=|\mathcal V_{d,f}|^{-1}\sum_{v\in\mathcal V_{d,f}}s_{d,m,v}$, where $\mathcal V_{d,f}$ is the evaluated configuration set. Baselines average all configurations listed in Appendix~\ref{app:baselines}. The main student uses one condition per dataset, as specified in Appendix~\ref{app:student-settings}. Each teacher has one configuration.

Each fixed reference pool contains 25 entries, comprising 20 baseline method means, 4 teacher scores, and the unedited input. Input scores are averaged over the same test source images for each assessment. Let $L_{d,m,f}$ and $E_{d,m,f}$ count reference entries with lower and equal scores, respectively, relative to $\bar s_{d,m,f}$. The percentile score is:
\begin{equation}
p_{d,m,f}=100\frac{L_{d,m,f}+\tfrac12 E_{d,m,f}}{25}.
\end{equation}
All 16 assessments use higher scores for better results. Joint Score averages these percentiles equally across assessments and datasets. The aesthetic and quality summaries use 12 and 4 assessments, respectively. Configuration averaging takes place in each assessor's original score units, before percentile conversion. Thus, each reference method has one entry regardless of its configuration count.

\textbfnoindent{External Comparison.}
Student and expert outputs are scored against the fixed reference pools without entering them. Reference members use the same rule, including their own tied score. Table~\ref{tab:assessment-joint} and Figures~\ref{fig:joint-rank-teaser}, \ref{fig:assessment-fivek}, and~\ref{fig:assessment-ppr10k} share these pools. For each expert-mean row, we first average the raw scores over all five FiveK experts or three PPR10K experts, then apply the same percentile calculation. Individual experts are also scored separately in the dataset-specific figures. In Table~\ref{tab:assessment-joint}, model and input scores average the two datasets equally, while each expert-mean row uses only its named dataset. Per-assessment ranks compare the 20 baselines, 4 teachers, student, and input. Each expert mean is ranked against these scores on its own dataset. The Joint column orders all 28 displayed rows by their reported scores. Ranks use unrounded values. Exact ties share the same rank, and the following rank skips the tied positions.

\textbfnoindent{Controlled Studies.}
Each ablation replaces the complete student as the evaluated model while keeping the 25 external reference entries fixed. Component models use FLUX.2-dev on FiveK and FireRed-Image-Edit on PPR10K. Teacher-supervision results report each condition separately. Their pooled row first averages raw scores across the four conditions, then computes JS. Adding an evaluated model leaves the reference pools unchanged.

\subsection{Evaluation Cost}
\label{app:evaluation-cost}

Scoring time is measured on a single NVIDIA H100 80GB GPU at approximately 1MP resolution. Timing covers 12 output variants per dataset, totaling 6,000 FiveK images and 27,432 PPR10K images. All 16 assessment models process every image. Table~\ref{tab:evaluation-cost} gives the time for each assessment. The H100 GPU-hours per output variant are:
\begin{equation*}
T_{\mathrm{eval}}=\frac{9{,}143+35{,}308}{12\times3{,}600}=1.029.
\end{equation*}
This comprises 0.212 GPU-hours for FiveK and 0.817 for PPR10K. Each variant requires $2{,}786\times16=44{,}576$ scores. Under the same setup, scoring $K$ output variants costs approximately $1.03\times K$ H100 GPU-hours.

\begin{table}[htbp]
\centering
\caption{Scoring time for 12 output variants at about 1MP on a single H100 GPU. Times are summed over all variants.}
\label{tab:evaluation-cost}
\small
\begin{tabular*}{\columnwidth}{@{\extracolsep{\fill}}lrr@{}}
\toprule
Assessment & FiveK (s) & PPR10K (s) \\
\midrule
NIMA~\cite{nima} & 815 & 2,466 \\
MUSIQ-AVA~\cite{musiq} & 442 & 1,183 \\
TOPIQ-IAA~\cite{topiq} & 503 & 1,466 \\
Q-Align~\cite{qalign} & 1,219 & 3,341 \\
ArtiMuse~\cite{artimuse} & 420 & 1,923 \\
ArtQuant~\cite{artquant} & 466 & 2,160 \\
NIMA-VGG16~\cite{nima} & 385 & 925 \\
MLSP-AVA~\cite{mlsp} & 610 & 2,874 \\
TANet-AVA~\cite{tanet} & 384 & 1,535 \\
TOPIQ-Res50~\cite{topiq} & 443 & 1,346 \\
LAION-AES~\cite{laion_aes} & 435 & 1,435 \\
ArtiMuse-AVA~\cite{artimuse} & 452 & 1,946 \\
MANIQA~\cite{maniqa} & 1,257 & 6,038 \\
LIQE~\cite{liqe} & 390 & 1,327 \\
TOPIQ-NR~\cite{topiq} & 467 & 3,862 \\
CLIP-IQA~\cite{clipiqa} & 455 & 1,481 \\
\midrule
Total & 9,143 & 35,308 \\
\bottomrule
\end{tabular*}
\end{table}

Runtime includes image reading, preprocessing, inference, and score writing. Model initialization is included for the 11 PyIQA assessors and excluded for the other five.

\section{Experimental Settings}
\label{app:settings}

\subsection{Student Training and Inference}
\label{app:student-settings}

\textbfnoindent{Training Data and Initialization.}
\studentname is trained jointly on all four teacher conditions using the source-level splits in Table~\ref{tab:dataset}. Each source image contributes its four teacher pairs in every epoch. Training shuffles all 47,980 source--teacher pairs without dataset-balanced resampling. Training resizes complete images to $384\times384$ without random crops or color augmentation. The encoder uses pretrained MobileNetV3-Small weights~\cite{mobilenetv3}, with fixed batch-normalization statistics and affine parameters. The remaining encoder parameters and color heads are trainable.

\textbfnoindent{Optimization.}
AdamW~\cite{adamw} runs for 30 epochs with cosine learning-rate decay. The initial learning rate is $3\times10^{-5}$ for the encoder and $3\times10^{-4}$ for the remaining parameters, with weight decay $10^{-4}$. A batch of four pairs and four-step gradient accumulation give an effective batch size of 16. Training uses gradient-norm clipping at one and seed 2026. The weights $(\lambda_{\mathrm{dist}},\lambda_{\mathrm{grid}},\lambda_{\mathrm{curve}},\lambda_{\mathrm{gamut}})$ are $(0.25,0.005,0.001,0.02)$. The schedule in Appendix~\ref{app:optimization} uses $e_{\mathrm{warm}}=2$ and $e_{\mathrm{ramp}}=4$, reaching full local strength at epoch six.

\textbfnoindent{Checkpoint Selection.}
Validation uses 5,520 pairs every two epochs with full local strength. Checkpoint selection uses only $\mathcal L_{\mathrm{col}}+0.25\mathcal L_{\mathrm{dist}}$, averaged first within each dataset--teacher group and then equally across the eight groups. We use the shared checkpoint from epoch 12, which has the lowest validation loss.

\textbfnoindent{Teacher Conditions.}
The main comparison uses the FLUX.2-dev condition on FiveK and the FireRed-Image-Edit condition on PPR10K. Each dataset uses the same condition for all its images. Both use the same validation-selected student checkpoint. Teacher-supervision, style-transfer, and target-similarity analyses report all four conditions.

\textbfnoindent{Correspondence Settings.}
For DINOv2-Small, source and teacher images are resized to $384\times384$ with Lanczos interpolation, then to $392\times392$ with antialiased bilinear interpolation. Channel means are $(0.485,0.456,0.406)$ and standard deviations are $(0.229,0.224,0.225)$. The feature grid has $h=28$. Both matching directions use a radius of four feature cells. Confidence thresholds are $(\tau_a,\tau_m,\tau_c)=(0.55,0.02,1.5)$, with cycle distance expressed in feature cells. The valid-weight threshold is $\tau_w=0.05$.

\textbfnoindent{Renderer Settings.}
Student previews use the same normalization, a 256-pixel long side, and a 32-pixel minimum short side. Curves use $K=32$ and $\kappa=0.1$. Global and local affine parameters $(a,b,c)$ are $(2,0.25,0.10)$ and $(1.5,0.20,0.12)$, respectively. Evaluation uses $\alpha=1$, preserves the source image aspect ratio, and renders at its original resolution.

\textbfnoindent{Supervision Settings.}
Color scales are $\mathcal S=\{32,64,128\}$, with $\epsilon=0.001$ and coverage floor $\eta=0.10$. The distribution view is $64\times64$, divided into four $32\times32$ regions. Each region uses 256 positions, giving 1,024 in total. There are 16 projection directions, including six predefined directions and $M=10$ spherical directions, and $N_q=128$ quantile levels.

\textbfnoindent{Training Resources.}
Training uses one NVIDIA GeForce RTX 5090. The 30-epoch training loop takes 6.15 hours, including data loading, forward and backward computation, and optimizer updates. This training-loop time excludes validation and offline correspondence construction.

\textbfnoindent{Numerical Precision.}
Neural feature extraction and parameter prediction use BF16 autocasting on CUDA. Curve construction, affine rendering, color-loss reductions, and quantile interpolation use FP32.

\subsection{Baseline Settings}
\label{app:baselines}

The 20 baselines contribute 67 output configurations, using publicly released weights and their corresponding preprocessing. All methods receive the same approximately 1MP sRGB PNG source images used by the teachers. Released weights are used for inference without further training. Each checkpoint follows its original training partition. Expert-specific weights and distinct architectures are evaluated separately. Each configuration uses one fixed checkpoint across both test datasets. Within each dataset and assessment, the main comparison first averages raw scores over the configurations of each baseline, then computes JS against the 25 method-level reference entries.

\textbfnoindent{Image-Adaptive 3D LUT~\cite{adaptive3dlut}.}
We evaluate two image-adaptive basis-LUT variants with paired and unpaired sRGB weights. Each uses its corresponding classifier architecture to combine three basis LUTs. The combined LUT is applied to RGB inputs in $[0,1]$ by trilinear interpolation at the source resolution. The FiveK weights use the training partition of \href{https://openaccess.thecvf.com/content_cvpr_2018/html/Chen_Deep_Photo_Enhancer_CVPR_2018_paper.html}{Deep Photo Enhancer (DPE)}.

\textbfnoindent{AdaInt~\cite{adaint}.}
The adaptive-interval LUT model uses two checkpoints, trained on FiveK sRGB and PPR10K expert A sRGB, respectively. AdaInt and SepLUT follow each checkpoint's MMEditing test pipeline for image-only inference and use its released tensor-to-image conversion. The FiveK training list includes 455 of our 500 FiveK test source images.

\textbfnoindent{SepLUT~\cite{seplut}.}
We evaluate five separable-LUT variants using the \href{https://github.com/ImCharlesY/SepLUT/tree/main/pretrained}{official weights}. The three FiveK variants use backbone channel multipliers and shared 1D/3D LUT dimensions of $(6,9)$, $(8,17)$, and $(8,33)$. The two PPR10K expert A variants use ResNet-18 with LUT dimensions of 9 and 17. The three FiveK variants use the same training partition as AdaInt.

\textbfnoindent{NamedCurves~\cite{namedcurves}.}
The model predicts curves from color names. We evaluate five checkpoints, comprising the \href{https://github.com/davidserra9/namedcurves#mit-adobe-fivek}{DPE and UEGAN rendering and split protocols for FiveK} and PPR10K experts A, B, and C. The DPE FiveK checkpoint uses evaluation-set scores for model selection.

\textbfnoindent{DeepLPF~\cite{deeplpf}.}
We use one released FiveK checkpoint of the local parametric filter model with the sequential FiveK training split and its corresponding preprocessing.

\textbfnoindent{HDRNet~\cite{hdrnet}.}
The bilateral affine model uses 13 \href{https://github.com/google/hdrnet#download-pretrained-models}{released retouching checkpoints}. Ten combine experts A through E with channel multipliers of 1 and 2. The remaining three use expert C. One model with channel multiplier 1 uses a split with 50 user-weighted test images. Two models with channel multipliers 1 and 2 use a split with 250 randomly selected test images. Bilateral slicing and affine transforms are evaluated in TensorFlow. The training sets associated with the 250-image and 50-image test splits contain at least 250 and 450 of our 500 FiveK test source images, respectively. Image-level training lists are unspecified for the ten expert-specific checkpoints.

\textbfnoindent{Exposure~\cite{exposure}.}
The model applies learned editing operations using its FiveK checkpoint and official PNG-input interface. This interface performs gamma-2.2 linearization and maximum-based exposure scaling before editing. Inference uses random seed 2026. Its unpaired training inputs include 190 of our FiveK test source images.

\textbfnoindent{CSRNet~\cite{csrnet}.}
We use the FiveK checkpoint of the conditional pointwise enhancement model. Its training list includes 451 of our 500 FiveK test source images.

\textbfnoindent{MIRNet~\cite{mirnet}.}
The multi-resolution feature network uses the \href{https://drive.google.com/file/d/1BsXOvhMz2z80E_V93dgD6QaEspZE0w-u/view}{official FiveK enhancement weights}.

\textbfnoindent{DeepUPE~\cite{deepupe}.}
We use the released checkpoint for color enhancement of underexposed images. Its image resizing disables antialiasing, following the supplied preprocessing, and bilateral slicing uses the same TensorFlow implementation as HDRNet.

\textbfnoindent{BPAM~\cite{bpam}.}
The pixel-adaptive MLP model uses four checkpoints, one for FiveK and three for PPR10K experts A, B, and C. The PPR10K networks take RGB and a portrait mask as four input channels. For PPR10K, the official 360p masks are resized bilinearly to the input dimensions. For FiveK, the mask channel is zero. The FiveK checkpoint follows the 3D-LUT training partition.

\textbfnoindent{Pigment~\cite{pigment}.}
We evaluate nine checkpoints of the pigment-based color model. Three FiveK checkpoints use a five-layer backbone, ResNet-18, and ResNet-34. The six PPR10K checkpoints combine ResNet-18 and ResNet-34 with experts A, B, and C. The supplied checkpoints are selected by PSNR on their evaluation partitions.

\textbfnoindent{RSFNet~\cite{rsfnet}.}
The region-aware filtering model uses 14 \href{https://huggingface.co/Vicky0522/RSFNet-models/tree/main}{released checkpoints} with their corresponding inference settings. On FiveK, the global, map, palette, and saliency variants each use two Lightroom input renderings, zeroed as shot and zeroed with expert C, giving eight checkpoints. PPR10K provides map and saliency variants for each of experts A, B, and C, giving six checkpoints. The eight FiveK training lists each include 455 of our 500 FiveK test source images.

\textbfnoindent{DiffRetouch~\cite{diffretouch}.}
We evaluate both FiveK and PPR10K checkpoints of the diffusion retouching model. All four attribute controls are zero. The official inference function uses 20 DDIM steps, one output per source image, and seed 2026.

\textbfnoindent{VeraRetouch~\cite{veraretouch}.}
The model uses its automatic retouching mode with temperature 0, BF16 inference, and full-image rendering. Its training references include FiveK and PPR10K expert edits, with image-level training membership unspecified.

\textbfnoindent{ColorCtrl~\cite{colorctrl}.}
BLIP-base generates a description of each source image using three-beam search, no sampling, and at most 24 new tokens. The source prompt is ``A photograph of [description]''. The target prompt appends ``, with professionally enhanced colors, balanced exposure, pleasing contrast, natural white balance, and rich realistic tones''. The editor uses FLUX.1-dev at $768\times768$ pixels, with 50 steps for both inversion and denoising, guidance 1, alpha 0.35, and seed 2026. Aspect-preserving Lanczos resizing and edge padding preserve the complete source image. Outputs are unpadded and resized to the evaluation dimensions.

\textbfnoindent{SPL~\cite{spl}.}
SPL uses the same text construction and image preprocessing as ColorCtrl. Its LCM Dreamshaper backbone operates at $512\times512$ pixels with 15 steps, source and target guidance of 1 and 2, a structure weight of 10,000, and 100 optimization steps. The attention and optimization schedules are both 0.8. The structure-preservation loss applies to the whole image. Post-processing is enabled and color preservation loss is disabled. The seed is 2026. Outputs are unpadded and resized to the evaluation dimensions.

The official SPL output filter selects 495 valid FiveK outputs and 2,264 PPR10K outputs. Scores are unavailable for the remaining images. TANet evaluates fixed groups, so groups containing a missing output are omitted as a whole, leaving 356 and 1,422 scores. Each reported mean uses the valid samples for that method and assessment.

\textbfnoindent{MonetGPT~\cite{monetgpt}.}
The model uses its automatic balanced editing mode with temperature 0.3 and BF16 inference. The language model receives aspect-preserving previews with a maximum side of 1,280 pixels, while editing operations act at the source resolution. Its three editing stages adjust white balance and tone, temperature and saturation, and HSL. We use the official GIMP and non-GIMP operations and convert the rendered TIFF images to RGB PNG for assessment.

\textbfnoindent{SA-LUT~\cite{salut}.}
The style-transfer model uses 64 basis LUTs and a LUT dimension of 17. Source and reference images are resized bilinearly to $512\times512$ to predict the fused LUT and context map. The context map is then resized bilinearly to the source dimensions, and quadrilinear interpolation applies the 4D LUT at full resolution. Both resizing operations use aligned corner pixels.

The style reference is selected from the training split. Each dataset uses 128 candidates sampled uniformly by source identifier, excluding groups shared with validation or test images. CLIP ViT-B/32 retrieves the nearest unedited training image by cosine similarity between L2-normalized embeddings. Retrieval is restricted to the source image's dataset. Its corresponding expert edit supplies the reference, using expert C for FiveK and expert A for PPR10K.

\textbfnoindent{AceTone~\cite{acetone}.}
We use the \href{https://huggingface.co/Vivre/AceTone-3B-PST-Preview}{AceTone-3B-PST-Preview} weights and the official LUT tokenizer with the two-image prompt. The source image appears first and the reference second. The prompt requests LUT tokens that reproduce the reference's color adjustments. Each source image receives exactly the same training reference selected for SA-LUT. The model uses BF16 inference and samples up to 128 tokens at temperature 0.1, resetting seed 2026 for each image. The first 64 LUT tokens form a $4\times4\times4$ code grid for decoding. Sequences shorter than 64 tokens are completed by repeating their first token, following the official evaluation. The decoded LUT is applied at the source image resolution.

\textbfnoindent{InstantRetouch~\cite{instantretouch}.}
InstantRetouch provides public training and inference code, but its repository does not provide download links for the trained teacher and distilled student checkpoints required for evaluation.\footnote{\href{https://github.com/OpenImagingLab/InstantRetouch}{Official InstantRetouch repository}, accessed October 28, 2026.}

\textbfnoindent{GenColor~\cite{gencolor}.}
GenColor is excluded because its trained color-generation weights have not been released.\footnote{\href{https://github.com/NTUYWANG103/GenColor}{Official GenColor repository}.} Its public Fusion network requires a generated color edit of the same source image. The missing generation stage uses Stable Diffusion 2.1 with ControlNet.

\begin{table*}[t]
\centering
\small
\caption{Aesthetic scores of teacher and expert edits on the 500 FiveK test images. Best expert is selected separately for each assessment by its mean score. Scores use each assessor's original scale.}
\label{tab:scores-fivek}
\begin{tabular}{lrrrrrr}
\toprule
Assessor $\uparrow$ & Best expert & FLUX.2-dev & Step1X-Edit & HunyuanImage & FireRed-Image-Edit & Teacher mean \\
\midrule
NIMA & 4.8844 & 5.2667 & 4.9651 & 5.0874 & 5.1709 & 5.1225 \\
MUSIQ-AVA & 4.4724 & 5.1577 & 4.5093 & 4.9590 & 4.9438 & 4.8924 \\
TOPIQ-IAA & 4.8667 & 5.2839 & 4.9168 & 5.1392 & 5.1343 & 5.1185 \\
Q-Align & 2.6093 & 2.9814 & 2.6938 & 2.8765 & 2.8841 & 2.8590 \\
ArtiMuse & 58.9402 & 61.6612 & 57.2180 & 60.3155 & 60.6237 & 59.9546 \\
ArtQuant & 3.7367 & 3.8754 & 3.7400 & 3.8083 & 3.8221 & 3.8114 \\
NIMA-VGG16 & 5.0314 & 5.3558 & 5.1340 & 5.1989 & 5.2768 & 5.2414 \\
MLSP-AVA & 4.3987 & 5.2628 & 4.5664 & 5.1169 & 5.0417 & 4.9970 \\
TANet-AVA & 5.1063 & 5.2920 & 5.1469 & 5.1910 & 5.2395 & 5.2173 \\
TOPIQ-Res50 & 4.9042 & 5.4178 & 4.9865 & 5.2607 & 5.2225 & 5.2219 \\
LAION-AES & 5.0294 & 5.3219 & 4.9925 & 5.1845 & 5.2276 & 5.1816 \\
ArtiMuse-AVA & 47.6442 & 51.5764 & 48.1989 & 50.3313 & 50.2756 & 50.0955 \\
\bottomrule
\end{tabular}
\end{table*}

\subsection{Ablation Settings}
\label{app:ablation-settings}
\textbfnoindent{Teacher Supervision.}
Each single-teacher student uses the shared student's architecture but learns only one teacher condition. Matched exposure allocates a 30-epoch budget to one teacher's training pairs, totaling 22,500 optimizer updates. Matched steps allocates the joint training budget of 89,970 updates to a single teacher. These budgets are fixed before validation-set model selection. Each model is selected by its validation color and distribution loss, using seed 2026. Both comparisons include all test source images and report the four teacher conditions separately. The pooled row averages their raw scores before computing JS.

\textbfnoindent{Color Transforms and Supervision.}
The component variants use the same training data, initialization seed, and 30-epoch budget. Test assessment fixes FLUX.2-dev on FiveK and FireRed-Image-Edit on PPR10K for every component model, using the same 25 external reference entries as the complete student. The variant without Local Affine Transform skips $A_\ell(p)$ during rendering, while keeping the global transforms and grid prediction. The variant without Multi-Scale Color Supervision sets $\mathcal S=\{28\}$ in the Multi-Scale Color Loss, using one $28\times28$ color view at the feature-grid resolution. The variant without Regional Color Distribution Loss sets $\lambda_{\mathrm{dist}}=0$, removing $\mathcal L_{\mathrm{dist}}$. The variant without Regional Partitioning uses the whole image as the single region in $\mathcal R$, while keeping distribution supervision. The variant without Confidence Weighting sets $w_t(p)=1$, while keeping local feature matching. The variant without Local Feature Matching replaces $\phi_t$ with the identity map while keeping cached confidence weights.

\textbfnoindent{Validation-Set Similarity.}
Validation-set similarity in Table~\ref{tab:student-components} compares the selected component models with their teacher targets. We evaluate $384\times384$ outputs on 450 FiveK and 930 PPR10K validation images under all four teacher conditions, giving 5,520 source--teacher pairs. PSNR, SSIM, and RGB mean absolute error (MAE) compare outputs directly with teacher targets, without alignment. MAE uses RGB values in $[0,1]$. Metrics are averaged within each dataset and teacher condition, then equally across the eight groups.

\subsection{Model Size and Inference}
\label{app:efficiency}
Inference parameter counts include all model components used to produce an output, including frozen conditioning encoders and autoencoders. For mixture-of-experts (MoE) models, we count total parameters across all experts. Optimizer states and training-only modules are excluded. Ranges in Figure~\ref{fig:joint-rank-teaser} cover the evaluated architectures within a method family. \studentname contains 1,127,660 parameters shared by all four teacher conditions, of which 1,115,548 are trainable during student learning. Four independent students require four such parameter sets, about 4.51M parameters in total. Student inference uses one parameter prediction followed by color rendering, without invoking a generative teacher or the correspondence encoder.

\begin{table*}[t]
\centering
\small
\caption{Aesthetic scores of teacher and expert edits on the 2,286 PPR10K test images. Best expert is selected separately for each assessment by its mean score. Scores use each assessor's original scale.}
\label{tab:scores-ppr10k}
\begin{tabular}{lrrrrrr}
\toprule
Assessor $\uparrow$ & Best expert & FLUX.2-dev & Step1X-Edit & HunyuanImage & FireRed-Image-Edit & Teacher mean \\
\midrule
NIMA & 5.3667 & 5.7168 & 5.4422 & 5.5533 & 5.7090 & 5.6053 \\
MUSIQ-AVA & 4.7579 & 5.5586 & 4.8682 & 5.3750 & 5.3572 & 5.2898 \\
TOPIQ-IAA & 5.3968 & 5.6691 & 5.4554 & 5.6412 & 5.6990 & 5.6162 \\
Q-Align & 3.4383 & 3.6971 & 3.5053 & 3.6869 & 3.7927 & 3.6705 \\
ArtiMuse & 58.1694 & 60.7372 & 59.1226 & 59.8198 & 59.0468 & 59.6816 \\
ArtQuant & 3.9735 & 3.9900 & 3.9726 & 3.9918 & 4.0013 & 3.9889 \\
NIMA-VGG16 & 5.5370 & 5.7059 & 5.5601 & 5.5979 & 5.7283 & 5.6481 \\
MLSP-AVA & 4.7261 & 5.6693 & 4.8801 & 5.5185 & 5.5001 & 5.3920 \\
TANet-AVA & 5.3993 & 5.5044 & 5.4093 & 5.4379 & 5.5068 & 5.4646 \\
TOPIQ-Res50 & 5.3074 & 5.7792 & 5.4135 & 5.6953 & 5.7204 & 5.6521 \\
LAION-AES & 5.7021 & 5.7677 & 5.6379 & 5.7305 & 5.8659 & 5.7505 \\
ArtiMuse-AVA & 54.6328 & 57.7837 & 55.3055 & 57.6107 & 58.2941 & 57.2485 \\
\bottomrule
\end{tabular}
\end{table*}

\begin{table*}[t]
\centering
\small
\caption{Technical image-quality scores of teacher and expert edits on FiveK and PPR10K. Best expert is selected separately for each dataset and assessment by its mean score. Scores use each assessor's original scale.}
\label{tab:scores-quality}
\begin{tabular}{llrrrrrr}
\toprule
Dataset & Assessor $\uparrow$ & Best expert & FLUX.2-dev & Step1X-Edit & HunyuanImage & FireRed-Image-Edit & Teacher mean \\
\midrule
FiveK & MANIQA & 0.3036 & 0.4239 & 0.3149 & 0.3820 & 0.3901 & 0.3777 \\
FiveK & LIQE & 2.9262 & 4.4660 & 3.0162 & 4.3860 & 4.3952 & 4.0659 \\
FiveK & TOPIQ-NR & 0.3907 & 0.6436 & 0.4136 & 0.5581 & 0.5761 & 0.5479 \\
FiveK & CLIP-IQA & 0.5239 & 0.6805 & 0.4434 & 0.6186 & 0.5424 & 0.5713 \\
\midrule
PPR10K & MANIQA & 0.3054 & 0.4483 & 0.3118 & 0.3908 & 0.4031 & 0.3885 \\
PPR10K & LIQE & 2.9793 & 4.7462 & 3.1944 & 4.5596 & 4.6253 & 4.2813 \\
PPR10K & TOPIQ-NR & 0.3696 & 0.6818 & 0.4033 & 0.5962 & 0.6050 & 0.5716 \\
PPR10K & CLIP-IQA & 0.5023 & 0.6844 & 0.4396 & 0.6090 & 0.5173 & 0.5626 \\
\bottomrule
\end{tabular}
\end{table*}

\section{Additional Assessment Results}
\label{app:additional-assessment}

\subsection{Dataset-Specific Comparisons}
\label{app:dataset-comparisons}

\begin{figure*}[t]
\centering
\includegraphics[width=\textwidth]{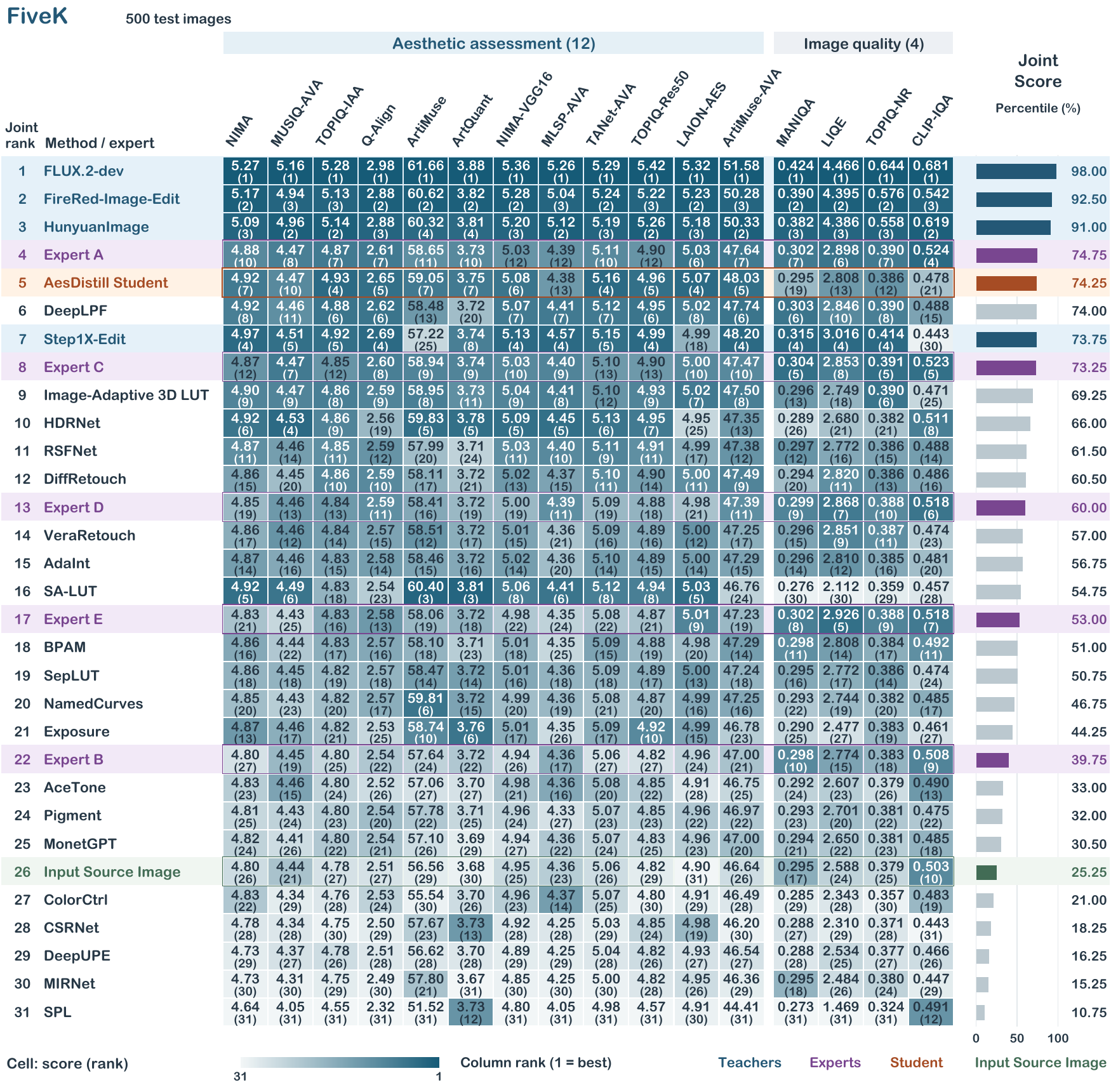}
\caption{Scores and ranks on FiveK for 4 teachers, 20 baselines, \studentname, the input, and all experts. Each cell shows a mean score and its rank in parentheses. Darker cells indicate better ranks. Bars show the Joint Score over all 16 assessments. Higher scores are better.}
\label{fig:assessment-fivek}
\end{figure*}

\begin{figure*}[t]
\centering
\includegraphics[width=\textwidth]{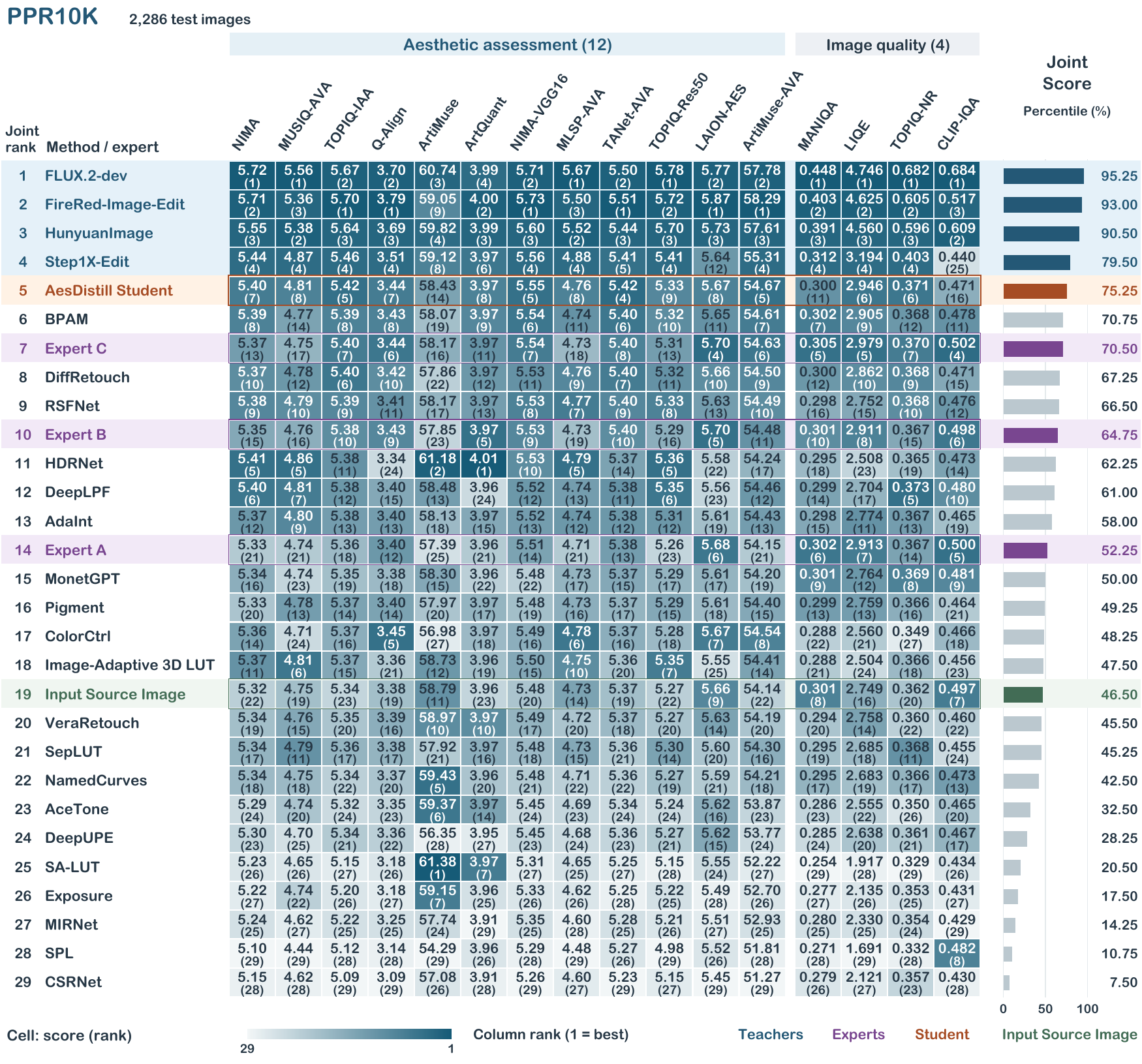}
\caption{Scores and ranks on PPR10K for 4 teachers, 20 baselines, \studentname, the input, and all experts. Each cell shows a mean score and its rank in parentheses. Darker cells indicate better ranks. Bars show the Joint Score over all 16 assessments. Higher scores are better.}
\label{fig:assessment-ppr10k}
\end{figure*}

Figures~\ref{fig:assessment-fivek} and~\ref{fig:assessment-ppr10k} report the separate FiveK and PPR10K results underlying Table~\ref{tab:assessment-joint}, with every expert shown individually. Joint Scores use the fixed pools in Appendix~\ref{app:ranking}. Inference settings and SPL coverage follow Appendix~\ref{app:baselines}.

\subsection{Comparison of Score Aggregation Methods}
\label{app:score-aggregation}

We compare Joint Score with normalized mean scores and mean ranks using the outputs and settings of Table~\ref{tab:assessment-joint}. All three summaries give equal weight to the 16 assessments and to the two datasets.

For method $f$, let $\bar s_{d,m,f}$ be its mean test score under assessment $m$ on dataset $d$. The two alternative summaries are:
\begin{equation*}
\begin{aligned}
S_{\mathrm{norm}}(f)&=\frac{100}{2\cdot16}\sum_{d=1}^{2}\sum_{m=1}^{16}
\frac{\bar s_{d,m,f}}{U_m},\\
R_{\mathrm{mean}}(f)&=\frac{1}{2\cdot16}\sum_{d=1}^{2}\sum_{m=1}^{16}
r_{d,m,f}.
\end{aligned}
\end{equation*}
Here $U_m$ is the nominal upper score: 100 for ArtiMuse and ArtiMuse-AVA, 5 for Q-Align, ArtQuant, and LIQE, 1 for MANIQA, TOPIQ-NR, and CLIP-IQA, and 10 for the remaining assessments. Ranks $r_{d,m,f}$ are computed jointly over the 4 teachers, 20 baselines, student, and input within each dataset and assessment, with tied scores assigned average ranks.

Table~\ref{tab:aggregation-comparison} shows that \studentname leads all 20 baselines and ranks directly behind the four teachers under all three aggregation rules.

\begin{table}[htbp]
\centering
\caption{Aggregation results on FiveK and PPR10K. Parentheses show overall ranks among the 26 entries. Bold marks the best enhancement model.}
\label{tab:aggregation-comparison}
\begingroup
\footnotesize
\definecolor{aggteacher}{HTML}{F0F7FC}
\definecolor{agginput}{HTML}{F1F4F2}
\definecolor{aggstudent}{HTML}{FDF0E4}
\setlength{\tabcolsep}{2pt}
\renewcommand{\arraystretch}{1.05}
\begin{tabular*}{\columnwidth}{@{\extracolsep{\fill}}l|rrr@{}}
\toprule
Method & \shortstack{Joint\\Score $\uparrow$} & \shortstack{Normalized\\mean (\%) $\uparrow$} & \shortstack{Mean\\rank $\downarrow$} \\
\midrule
\rowcolor{aggteacher}
FLUX.2-dev~\cite{flux2} & 96.62\,(1) & 60.6386\,(1) & 1.3438\,(1) \\
\rowcolor{aggteacher}
FireRed-Image-Edit~\cite{firered} & 92.75\,(2) & 58.2966\,(3) & 2.3125\,(2) \\
\rowcolor{aggteacher}
HunyuanImage~\cite{hunyuan} & 90.75\,(3) & 58.3312\,(2) & 2.8125\,(3) \\
\rowcolor{aggteacher}
Step1X-Edit~\cite{reasonedit} & 76.62\,(4) & 52.3061\,(4) & 6.6250\,(4) \\
\midrule
\rowcolor{aggstudent}
\textbf{\studentname} & \textbf{74.75}\,(5) & \textbf{51.7096}\,(5) & \textbf{7.3125}\,(5) \\
\midrule
DeepLPF~\cite{deeplpf} & 67.50\,(6) & 51.4654\,(6) & 9.2812\,(6) \\
HDRNet~\cite{hdrnet} & 64.12\,(7) & 51.3480\,(9) & 10.0312\,(7) \\
RSFNet~\cite{rsfnet} & 64.00\,(8) & 51.3021\,(10) & 10.3125\,(8) \\
DiffRetouch~\cite{diffretouch} & 63.88\,(9) & 51.3694\,(8) & 10.4375\,(9) \\
BPAM~\cite{bpam} & 60.88\,(10) & 51.3999\,(7) & 11.1562\,(10) \\
Image-Adaptive 3D LUT~\cite{adaptive3dlut} & 58.38\,(11) & 51.0109\,(14) & 11.6875\,(11) \\
AdaInt~\cite{adaint} & 57.38\,(12) & 51.2148\,(11) & 12.0625\,(12) \\
VeraRetouch~\cite{veraretouch} & 51.25\,(13) & 51.1107\,(12) & 13.5625\,(13) \\
SepLUT~\cite{seplut} & 48.00\,(14) & 50.9877\,(15) & 14.4688\,(14) \\
NamedCurves~\cite{namedcurves} & 44.62\,(15) & 51.0457\,(13) & 15.2188\,(15) \\
Pigment~\cite{pigment} & 40.62\,(16) & 50.8852\,(16) & 16.3438\,(16) \\
MonetGPT~\cite{monetgpt} & 40.25\,(17) & 50.8849\,(17) & 16.3438\,(16) \\
SA-LUT~\cite{salut} & 37.62\,(18) & 49.3689\,(24) & 16.8750\,(18) \\
\rowcolor{agginput}
Input Source Image & 35.88\,(19) & 50.8619\,(18) & 17.3750\,(19) \\
ColorCtrl~\cite{colorctrl} & 34.62\,(20) & 50.2490\,(20) & 17.7188\,(20) \\
AceTone~\cite{acetone} & 32.75\,(21) & 50.5289\,(19) & 18.2500\,(21) \\
Exposure~\cite{exposure} & 30.88\,(22) & 49.7860\,(22) & 18.7188\,(22) \\
DeepUPE~\cite{deepupe} & 22.25\,(23) & 50.2130\,(21) & 20.9375\,(23) \\
MIRNet~\cite{mirnet} & 14.75\,(24) & 49.4571\,(23) & 22.7812\,(24) \\
CSRNet~\cite{csrnet} & 12.88\,(25) & 48.9657\,(25) & 23.2812\,(25) \\
SPL~\cite{spl} & 10.75\,(26) & 47.2775\,(26) & 23.7500\,(26) \\
\bottomrule
\end{tabular*}
\endgroup
\end{table}

\begin{figure*}[t]
\centering
\includegraphics[width=\textwidth]{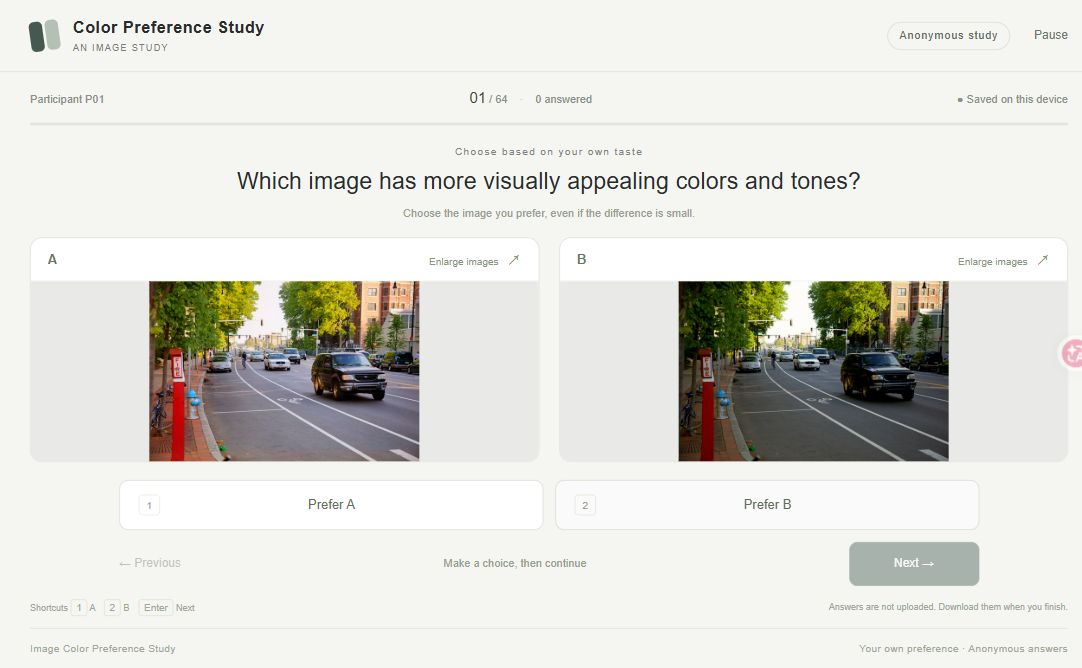}
\caption{Binary color-preference interface. Viewers compare two anonymous outputs and choose Prefer A or Prefer B based on colors and tones.}
\label{fig:human-study-interface}
\end{figure*}

\subsection{Teacher and Expert Scores}
\label{app:scores}

Tables~\ref{tab:scores-fivek} and \ref{tab:scores-ppr10k} report the aesthetic mean scores, and Table~\ref{tab:scores-quality} reports the technical-quality scores. Figure~\ref{fig:aesthetic-gain-distributions} shows paired aesthetic gains in the original score units. Expert selection and score aggregation follow Appendix~\ref{app:assessors}.

On both datasets, the four-teacher mean exceeds the best expert on all 12 aesthetic and 4 image-quality assessments. FLUX.2-dev, HunyuanImage, and FireRed-Image-Edit each exceed the best expert on all 12 aesthetic assessments. Step1X-Edit exceeds the best expert on 10 of 12, with lower scores on ArtiMuse and LAION-AES in FiveK, and on LAION-AES and ArtQuant in PPR10K.

\textbfnoindent{Gains over Input Images.}
Using the main evaluation settings in Section~\ref{sec:experimental-setup}, \studentname increases NIMA for 371 of 500 FiveK images (74.2\%) and 1,729 of 2,286 PPR10K images (75.6\%). Input and output scores are paired by source image ID, and gains count strictly positive differences. Mean NIMA increases from 4.802 to 4.920 on FiveK and from 5.322 to 5.397 on PPR10K.

\subsection{Additional Ablation Results}
\label{app:teacher-supervision-results}
\label{app:component-results}

\textbfnoindent{Teacher Supervision.}
Table~\ref{tab:single-joint-teacher} reports each teacher condition and the pooled score across all four. Joint training and matched exposure reach the same pooled JS of 71.00, compared with 69.12 for matched steps. Among the jointly trained conditions, FireRed-Image-Edit gives the highest combined JS of 72.12.

\textbfnoindent{Student Components.}
Table~\ref{tab:student-components} shows lower aesthetic scores for variants without Multi-Scale Color Supervision or regional supervision, supporting color learning across scales and image regions. The variant without Local Feature Matching scores 65.00 in quality and 73.50 in aesthetics, compared with 64.50 and 78.17 for the complete student. The complete student achieves the highest joint result and the closest agreement to teacher targets on validation data.

\subsection{Correspondence and Source Content}
\label{app:confidence-visualization}

\begin{figure*}[p]
\centering
\includegraphics[width=0.84\textwidth]{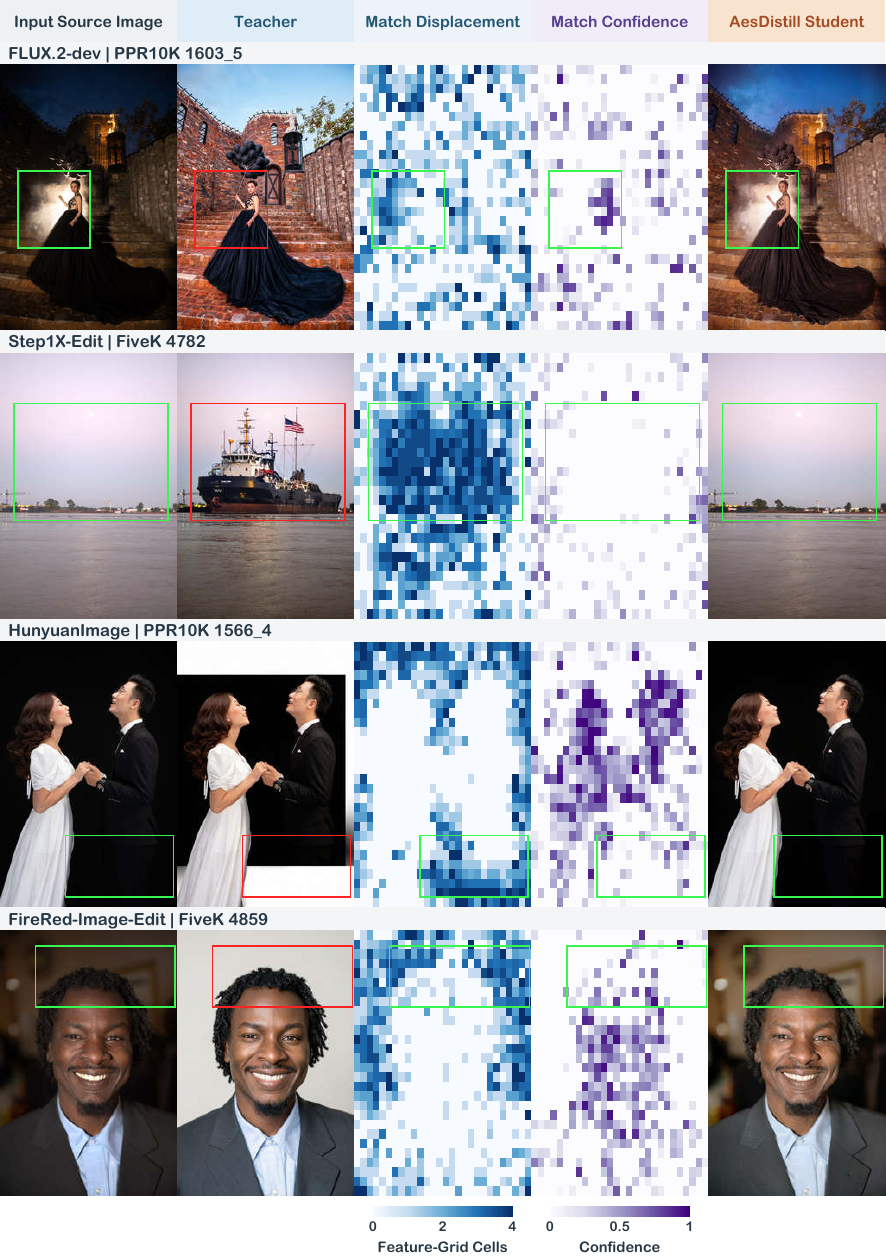}
\caption{Correspondence diagnostics for teacher edits that remove smoke, add a boat or frame, or replace the background. Red boxes mark teacher changes. Green boxes mark matching input, map, and student regions. Displacement and confidence maps use shared scales.}
\label{fig:appendix-correspondence}
\end{figure*}

\textbfnoindent{Visualization Settings.}
Figure~\ref{fig:appendix-correspondence} inspects the supervision mechanism on the four image pairs in Figure~\ref{fig:structure-preservation}. The visualizations use the frozen DINOv2-Small encoder, preprocessing, and matching rule in Appendix~\ref{app:correspondence}. Each map is defined at source image locations on a $28\times28$ feature grid and displayed at the source image aspect ratio. Displacement is the distance to the matched teacher location in feature-grid cells. Confidence is shown before applying the supervision threshold $\tau_w$, using the same $[0,1]$ scale in all cases.

\textbfnoindent{Source Content.}
Regions containing the added boat, white frame, and replaced background receive little matching weight. In the FLUX.2-dev example, the removed smoke has low confidence while the nearby person has stronger matches. Figure~\ref{fig:structure-preservation} shows that the student keeps the smoke, harbor layout, and portrait backgrounds in the source images while changing their colors.

\begin{figure*}[p]
\centering
\includegraphics[width=\textwidth]{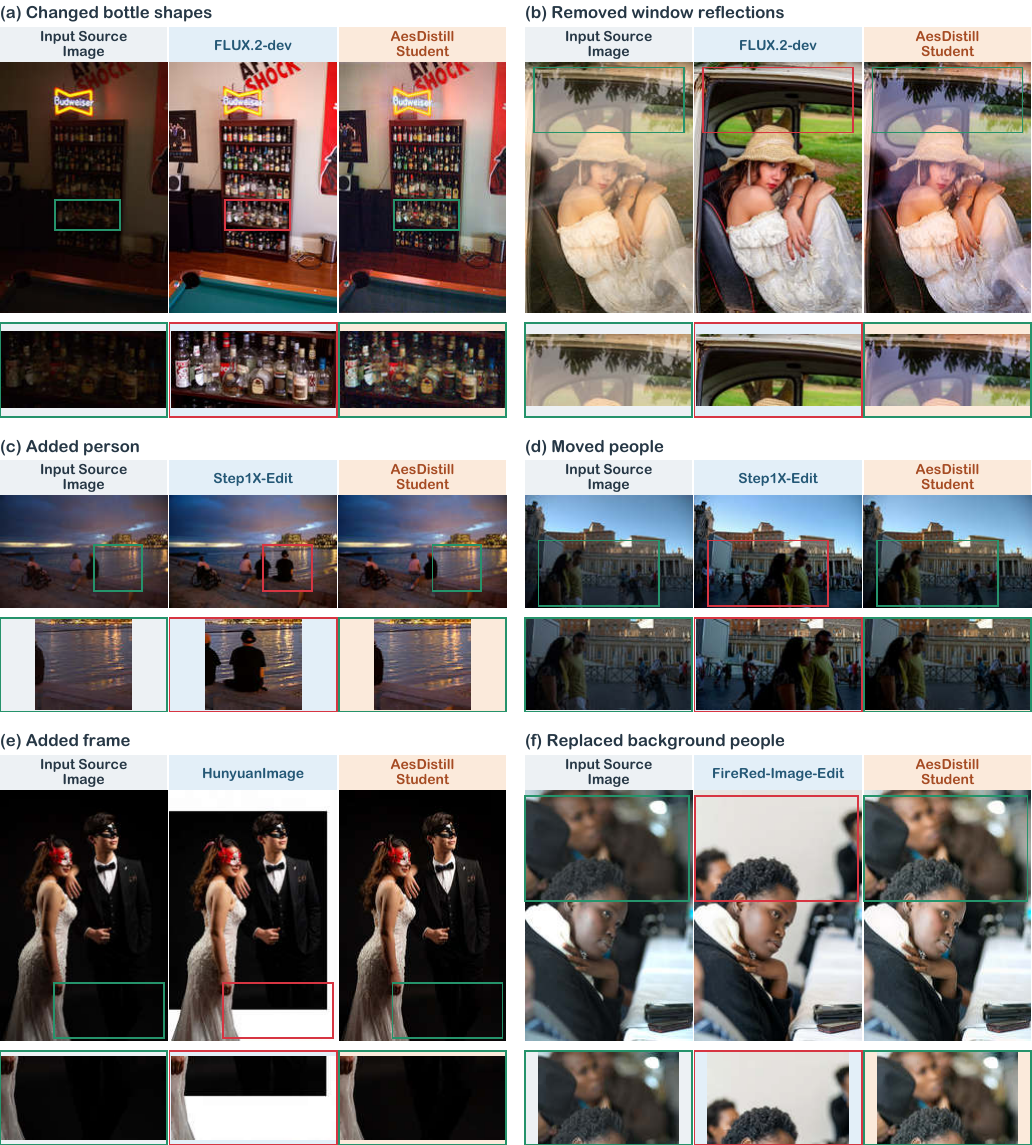}
\caption{Source image structure under teacher hallucinations on FiveK and PPR10K. Within each group, columns show the input, teacher edit, and \studentname output. Red boxes highlight teacher content changes, and green boxes mark corresponding input and student regions. Lower strips enlarge the boxed regions.}
\label{fig:appendix-structure-preservation}
\end{figure*}

\textbfnoindent{Structure under Teacher Hallucinations.}
Figure~\ref{fig:appendix-structure-preservation} shows six additional FiveK and PPR10K examples. Here, teacher hallucinations refer to unintended additions, removals, or replacements of scene content. The teachers change bottle shapes, remove window reflections, add a person or a frame, move people, and replace background figures. Each student output uses the condition of its paired teacher. \studentname preserves the source image's objects, person locations, and background layout while enhancing colors. Its color transforms act at the original pixel positions (Section~\ref{sec:color-transforms}), allowing aesthetic color transfer while preserving semantic structure.

\subsection{Teacher Color Transfer}
\label{app:multiteacher-styles}

\begin{figure*}[p]
\centering
\includegraphics[width=\textwidth]{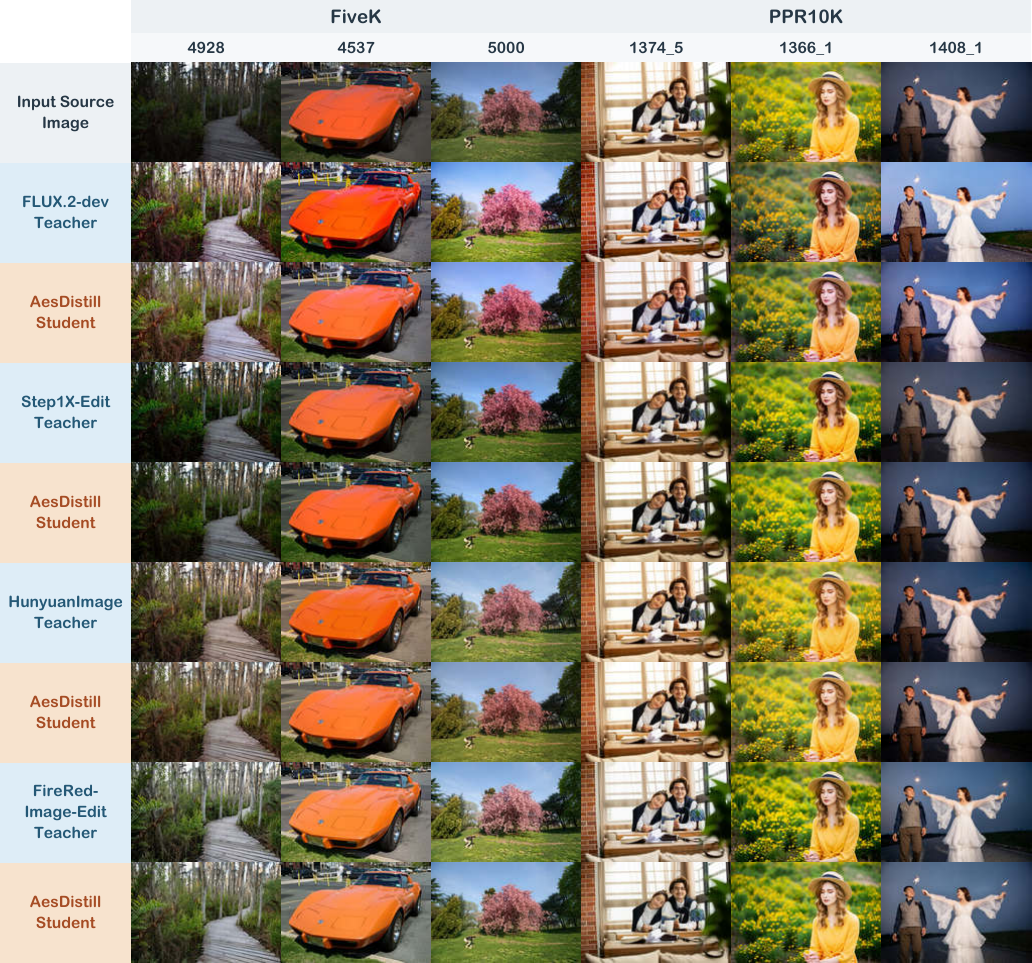}
\caption{Teacher color styles and their transfer to one jointly trained \studentname on FiveK and PPR10K. Each column shows one source image. Below the input row, each teacher edit is paired with the student output under the same teacher condition.}
\label{fig:appendix-styles}
\end{figure*}

\textbfnoindent{Teacher-Conditioned Styles.}
Figure~\ref{fig:appendix-styles} extends Figure~\ref{fig:teacher-student-styles} to six additional test images. One shared student checkpoint is used across both datasets and all four teacher conditions. In the forest scene, the FLUX.2-dev condition raises shadow brightness and strengthens warm colors, while the Step1X-Edit condition preserves deeper shadows. In the dusk portrait, the shared student follows differences in background brightness and color temperature across teachers.

\textbfnoindent{Similarity to Teacher Targets.}
Table~\ref{tab:student-teacher-similarity} quantifies agreement with the full teacher image under the protocol in Appendix~\ref{app:teacher-similarity}. Step1X-Edit gives the highest PSNR and SSIM and the lowest LPIPS on both datasets, while HunyuanImage gives the lowest PPR10K color difference.

\subsection{Additional Qualitative Comparisons}
\label{app:qualitative}

\begin{figure*}[p]
\centering
\includegraphics[width=\textwidth]{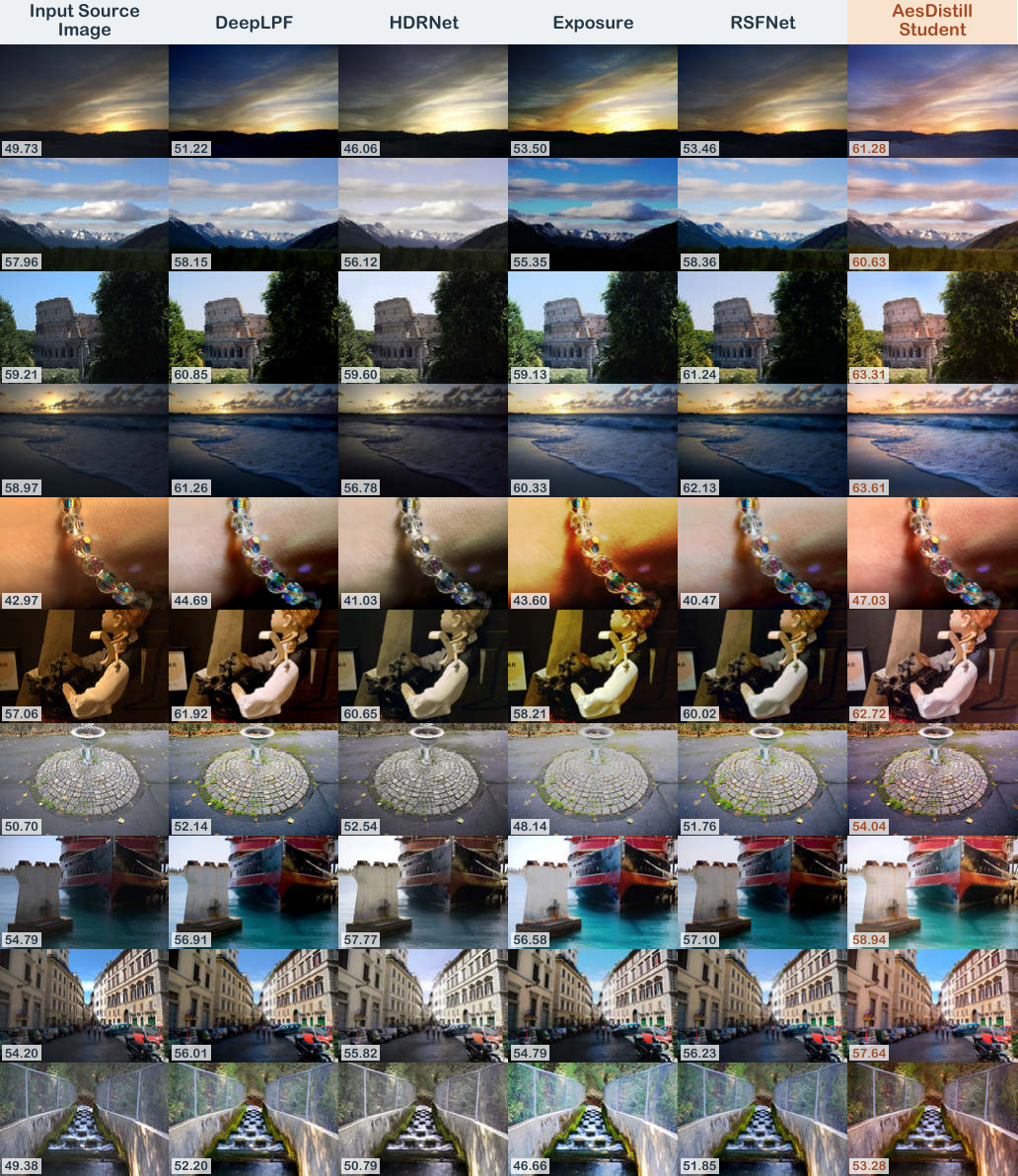}
\caption{Additional FiveK comparisons of \studentname with DeepLPF, HDRNet, Exposure, and RSFNet on ten distinct inputs. Lower-left values are ArtiMuse scores (higher is better).}
\label{fig:app-fivek-a}
\end{figure*}

\begin{figure*}[p]
\centering
\includegraphics[width=\textwidth]{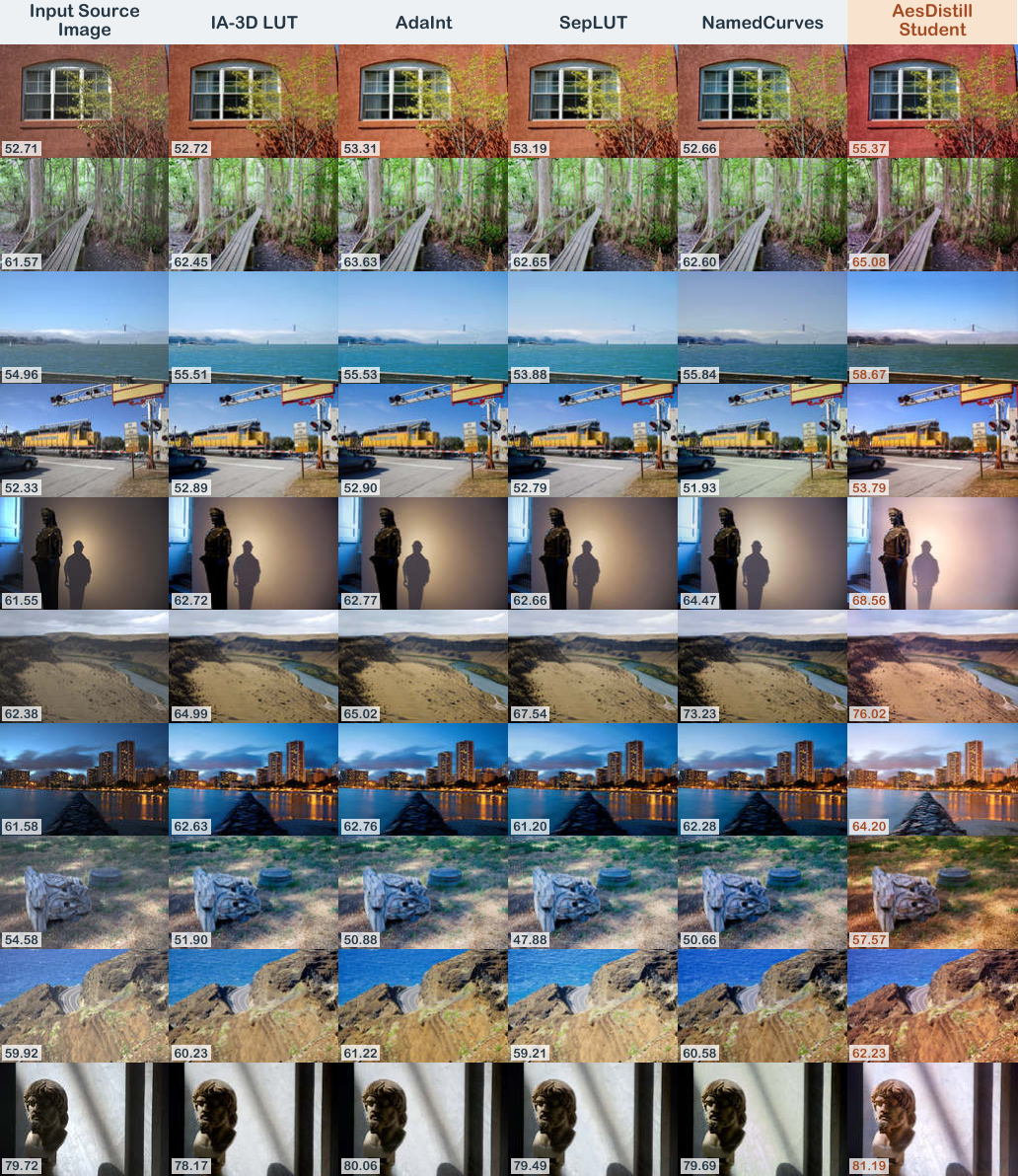}
\caption{Additional FiveK comparisons of \studentname with Image-Adaptive 3D LUT, AdaInt, SepLUT, and NamedCurves on ten distinct inputs. Lower-left values are ArtiMuse scores (higher is better).}
\label{fig:app-fivek-b}
\end{figure*}

\begin{figure*}[p]
\centering
\includegraphics[width=\textwidth]{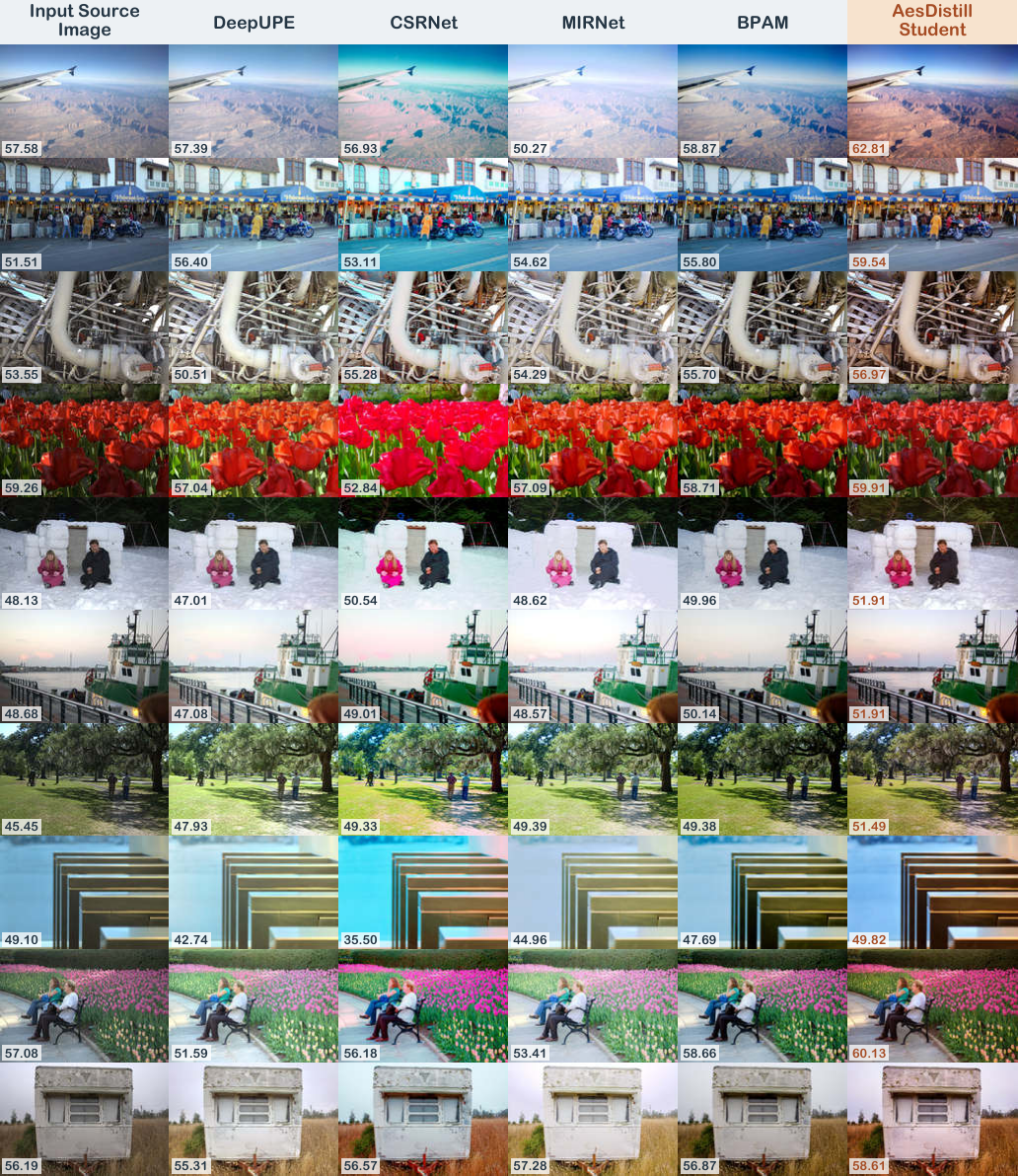}
\caption{Additional FiveK comparisons of \studentname with DeepUPE, CSRNet, MIRNet, and BPAM on ten distinct inputs. Lower-left values are ArtiMuse scores (higher is better).}
\label{fig:app-fivek-c}
\end{figure*}

\begin{figure*}[p]
\centering
\includegraphics[width=\textwidth]{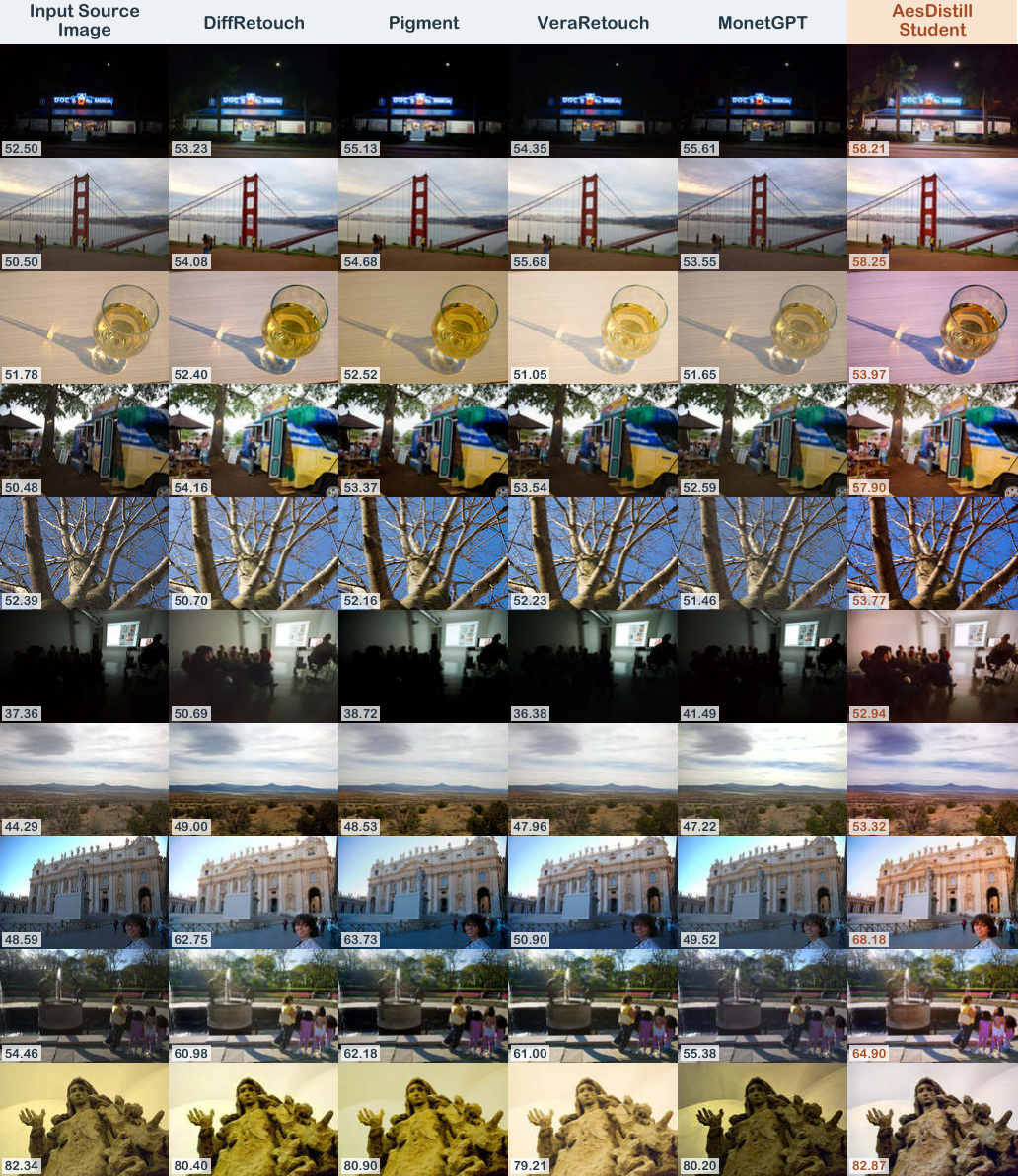}
\caption{Additional FiveK comparisons of \studentname with DiffRetouch, Pigment, VeraRetouch, and MonetGPT on ten distinct inputs. Lower-left values are ArtiMuse scores (higher is better).}
\label{fig:app-fivek-d}
\end{figure*}

\begin{figure*}[p]
\centering
\includegraphics[width=\textwidth]{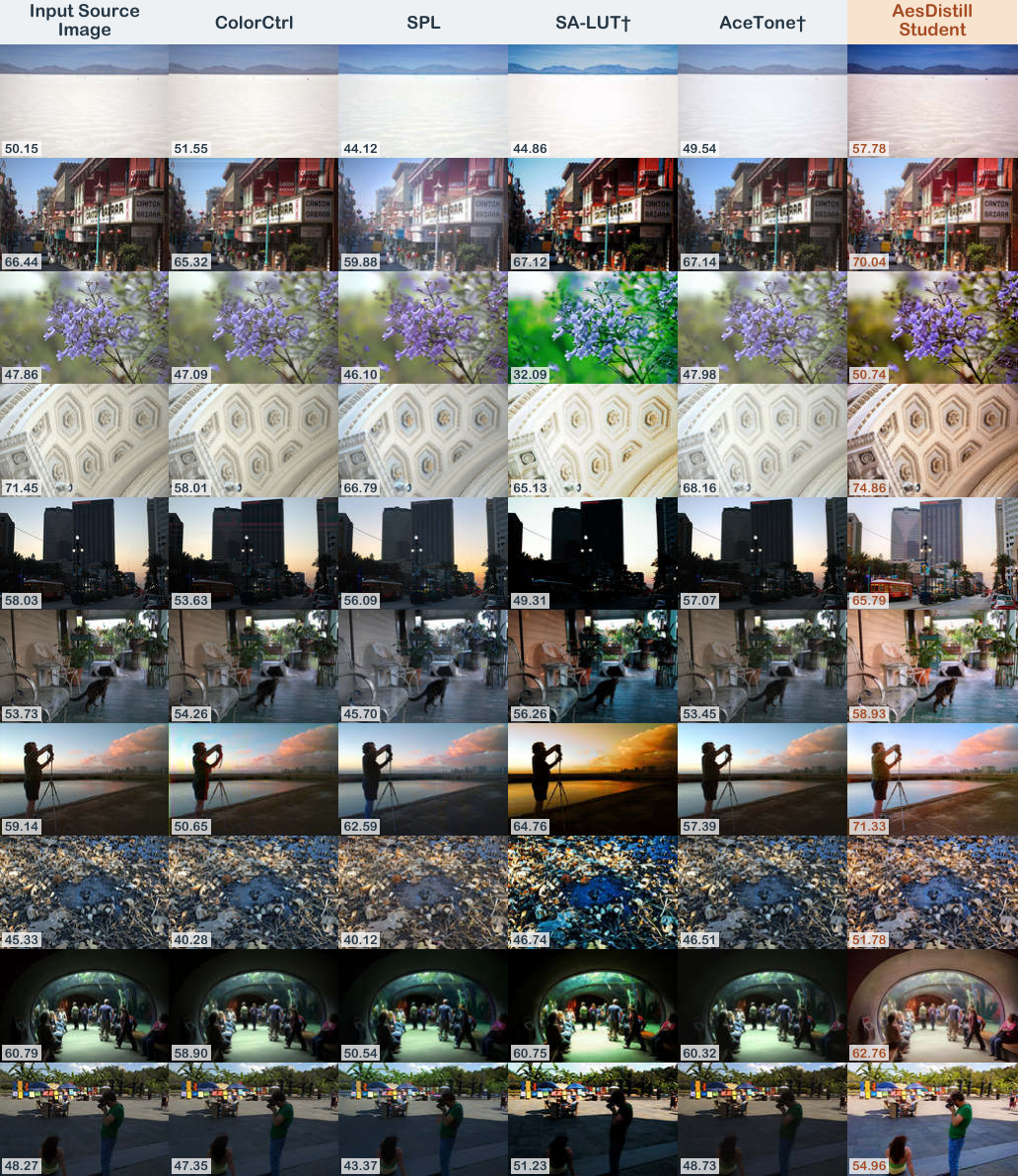}
\caption{Additional FiveK comparisons with ColorCtrl, SPL, SA-LUT, and AceTone. Lower-left values are ArtiMuse scores (higher is better). $\dagger$~SA-LUT and AceTone use a retrieved training-set expert reference (Appendix~\ref{app:baselines}).}
\label{fig:app-fivek-e}
\end{figure*}

Figures~\ref{fig:app-fivek-a}--\ref{fig:app-fivek-e} present five groups of ten distinct additional FiveK test inputs, covering all 20 baselines. Each group compares \studentname with four baselines. The cases span landscapes, architecture, vegetation, transport, interiors, human activities, and close-up objects. Model settings follow Appendix~\ref{app:baselines}.

\section{Human Preference Study}
\label{app:human-preference-study}

\begin{table*}[t]
\centering
\small
\caption{Human preference for image color enhancement. Cells report wins/total judgments and preference rates for the first-listed output. Expert references are FiveK expert A and PPR10K expert C. Combined rates weight the datasets equally.}
\label{tab:human-preference}
\begin{tabular}{@{}lrrr@{}}
\toprule
Comparison & FiveK & PPR10K & Combined \\
\midrule
FLUX.2-dev vs.\ expert & 23/24 (95.83\%) & 24/24 (100.00\%) & 47/48 (97.92\%) \\
Step1X-Edit vs.\ expert & 17/24 (70.83\%) & 18/24 (75.00\%) & 35/48 (72.92\%) \\
HunyuanImage vs.\ expert & 22/24 (91.67\%) & 19/24 (79.17\%) & 41/48 (85.42\%) \\
FireRed-Image-Edit vs.\ expert & 21/24 (87.50\%) & 23/24 (95.83\%) & 44/48 (91.67\%) \\
All teachers vs.\ expert & 83/96 (86.46\%) & 84/96 (87.50\%) & 167/192 (86.98\%) \\
\midrule
\studentname vs.\ expert & 75/96 (78.13\%) & 60/96 (62.50\%) & 135/192 (70.31\%) \\
\studentname vs.\ HDRNet & 78/96 (81.25\%) & 61/96 (63.54\%) & 139/192 (72.40\%) \\
\studentname vs.\ DeepLPF & 75/96 (78.13\%) & 64/96 (66.67\%) & 139/192 (72.40\%) \\
\bottomrule
\end{tabular}
\end{table*}

\subsection{Study Protocol}
\label{app:human-study}
The study compares the color appeal of generative teacher targets and student outputs with expert edits and baseline outputs.

\textbfnoindent{Participants and Images.}
Twelve participants evaluated 64 test source images, with 32 from FiveK and 32 from PPR10K. Expert references were FiveK expert A and PPR10K expert C, selected by dataset-level Joint Score in Appendix~\ref{app:ranking}.

\textbfnoindent{Comparisons.}
The four comparisons were teacher versus expert, \studentname versus expert, \studentname versus HDRNet~\cite{hdrnet}, and \studentname versus DeepLPF~\cite{deeplpf}. For teacher-versus-expert comparisons, each source image was assigned one teacher, with eight source images per teacher in each dataset. Teacher results are reported separately.

\textbfnoindent{Trial Allocation.}
The 64 source images provided 256 image pairs across the four comparisons. Each pair was evaluated by three different participants, giving 768 judgments. Each participant completed 64 trials, seeing each source image once, with 32 trials per dataset and 16 per comparison. Thus, each teacher received 24 judgments per dataset against the expert, while each student comparison received 96 judgments per dataset. Method identities were hidden, and trial order and left/right positions were randomized with balanced presentation.

\textbfnoindent{Viewing Procedure.}
Figure~\ref{fig:human-study-interface} illustrates the binary interface. Participants answered ``Which image has more visually appealing colors and tones?'' by choosing ``Prefer A'' or ``Prefer B.'' The paired outputs were shown side by side without method names or assessment scores. The task required a choice between the two outputs and did not include a tie option.

\textbfnoindent{Analysis.}
The preference rate is $P=W/N$, where $W$ is the number of judgments favoring the first-listed output and $N$ is the total number of judgments for that comparison. We report counts and rates separately for each dataset and average the two dataset rates equally. The pooled teacher rate gives equal weight to all four teachers.

\subsection{Results}
\label{app:human-study-results}
Table~\ref{tab:human-preference} reports the binary preferences. The four teachers received 167 of 192 preferences over expert edits (86.98\%). Each teacher was preferred over its expert reference on both datasets. \studentname received 135 of 192 preferences over experts (70.31\%), 139 of 192 over HDRNet (72.40\%), and 139 of 192 over DeepLPF (72.40\%). Student preference rates were higher on FiveK, while remaining above 50\% in all three PPR10K comparisons. These judgments support the color appeal of teacher targets and the student's learned color transforms.

\end{document}